\pdfoutput=1
\documentclass[11pt]{article}

\usepackage[final]{acl}

\usepackage{times}
\usepackage{latexsym}

\usepackage[T1]{fontenc}

\usepackage[utf8]{inputenc}

\usepackage{microtype}

\usepackage{inconsolata}

\usepackage{graphicx}

\usepackage{amsmath,amsfonts,bm}

\def\eqref#1{equation~\ref{#1}}

\def\1{\bm{1}}

\DeclareMathAlphabet{\mathsfit}{\encodingdefault}{\sfdefault}{m}{sl}
\SetMathAlphabet{\mathsfit}{bold}{\encodingdefault}{\sfdefault}{bx}{n}

\usepackage{subcaption}
\usepackage{hyperref}
\usepackage{url}
\usepackage{multirow}
\usepackage{lipsum}
\usepackage{microtype}
\usepackage{amsmath}
\usepackage{amssymb}
\usepackage{enumitem}

\usepackage{wrapfig}
\usepackage{inconsolata}
\usepackage{booktabs}
\usepackage{siunitx}
\usepackage[table]{xcolor}
\usepackage[colorinlistoftodos]{todonotes}
\usepackage{placeins}
\usepackage{makecell} %
\newcommand{\phnum}[1]{\textcolor{gray!70}}

\usepackage{graphicx}
\usepackage[table]{xcolor}
\newcommand{\phmark}{\(\triangleright\)} %

\usepackage{multicol}

\title{Beyond the Final Layer: Intermediate Representations for Better Multilingual Calibration in Large Language Models}

\author{\textbf{Ej Zhou}\thanks{\,Equal contribution.}\quad
  \textbf{Caiqi Zhang}\footnotemark[1]\quad
  \textbf{Tiancheng Hu}\quad
  \textbf{Chengzu Li}\\
  \textbf{Nigel Collier}\quad
  \textbf{Ivan Vuli\'c}\quad
  \textbf{Anna Korhonen}\\
  Language Technology Lab, University of Cambridge\\
  \texttt{\{yz926,cz391,th656,cl917,nhc30,iv250,alk23\}@cam.ac.uk}}

\begin{document}

\maketitle

\begin{abstract}
Confidence calibration, the alignment of a model's predicted confidence with its actual accuracy, is critical for the reliable deployment of Large Language Models (LLMs), yet remains under-explored in multilingual contexts. We conduct the first large-scale systematic study of multilingual calibration across eight model families and over 100 languages, finding that non-English languages are worse-calibrated than English (e.g., LLaMA3's average Expected Calibration Error is 23.1\% on non-English against 4.6\% on English). To diagnose this, we investigate the model's internal representations and find that the final layer, shaped by English-centric training, provides a poor signal for multilingual confidence, whereas late-intermediate layers offer a better-calibrated signal. Building on this, we introduce a suite of training-free methods, including Language-Aware Confidence Ensemble (LACE), which adaptively selects an ensemble of layers for each language. Both the multilingual calibration gap and the late-intermediate peak persist beyond MCQA, in open-ended QA, chain-of-thought reasoning, and alternative MCQA scoring protocols. Our study highlights the calibration costs associated with English-centric alignment and offers a path toward more globally equitable LLMs.

\end{abstract}

\section{Introduction}

Calibration in machine learning denotes the alignment between a model’s predicted confidence and the empirical probability that its predictions are correct \citep{guo2017calibration, geng-etal-2024-survey, zhang-etal-2025-roads, zhang-etal-2026-lovec}.\footnote{In this paper, we distinguish between calibration as a \emph{property} and as a \emph{process}. We use the term \emph{calibration} to refer to the property of a model's confidence being well-aligned with its accuracy. In contrast, the methods used to achieve this alignment are referred to as \emph{calibration methods}.} A model is perfectly calibrated if predictions assigned 80\% confidence are correct approximately 80\% of the time. Calibration is particularly critical for large language models (LLMs) in high-stakes applications such as medical diagnosis, legal advice, and decision support, where miscalibration can amplify harm \citep{zhang-etal-2024-luq, yang-etal-2025-logu, yang-etal-2025-uncle, hu-etal-2026-dial}. Well-calibrated LLMs make their reliability explicit and interpretable, improving downstream trust and safety.

Despite its importance, most calibration work on LLMs has focused on English \citep{xue2024mlingconf}. This reflects what \citet{ruder-etal-2022-square} termed \emph{Square One Bias}: research often advances along a single axis (e.g., English alignment or multilingual coverage), while their intersection (multilingual calibration) remains under-explored. Prior studies also have practical limitations: they cover only a few languages, often rely on machine-translated data, and evaluate a limited set of models \citep{xue2024mlingconf, yang2023calibration}.

To address this gap, we conduct a large-scale systematic study of multilingual calibration. Our analysis spans eight major model families and over 100 languages, using human-curated benchmarks (MMMLU and Belebele) with $\sim10^5$ instances. Non-English languages consistently exhibit lower accuracy and worse calibration: LLaMA3's average Expected Calibration Error (ECE) is 23.1\% on non-English against 4.6\% on English, a five-fold gap. The miscalibration pattern itself depends on training priorities. Prior work has largely framed LLM miscalibration as overconfidence \citep{zhang-etal-2024-calibrating, chhikara2025mindconfidencegapoverconfidence}; we observe two regimes. English-aligned models like LLaMA3 maintain calibration only for English; multilingual-first models like Aya are systematically over-confident across all languages, including English. Current alignment strategies thus fail to generalise to non-English contexts.

To identify the architectural source of this miscalibration, we examine the common practice of deriving confidence scores only from the final layer. Following work showing intermediate layers encode more language-agnostic representations \citep{bandarkar-etal-2025-layer, wendler-etal-2024-llamas}, we hypothesise that the final layer, shaped by English-dominated training, provides a sub-optimal signal for multilingual confidence. Our layer-wise analysis reveals a dichotomy: English calibration improves \textit{monotonically} with model depth and peaks at the final layer, whereas for nearly all non-English languages, \textit{late-intermediate layers} provide better-calibrated confidence estimates than the final layer.

This insight motivates our methodological contribution: training-free calibration methods that leverage these intermediate representations. We compare three confidence elicitation strategies: the \textit{best layer} method selects the single most calibrated intermediate layer; the \textit{good-layers ensemble} aggregates signals from multiple layers; and \textit{LACE} (Language-Aware Confidence Ensemble) adaptively tailors layer selection to each target language. The three methods yield consistent multilingual calibration improvements across all models, and combining \textit{LACE} with post-hoc methods such as Temperature Scaling \citep{guo2017calibration} yields further gains. Finally, we verify that both the multilingual gap and the late-intermediate peak hold in free-form generation (open-ended QA, chain-of-thought) and under alternative MCQA scoring protocols (\S\ref{sec:generative}).

Our contributions are fourfold:
(1) An empirical analysis of multilingual LLM calibration across eight model families and over 100 languages, revealing systematic disparities between English and other languages.
(2) The first layer-wise investigation of multilingual calibration, showing that intermediate layers offer a more reliable calibration signal than the final layer.
(3) Training-free calibration methods that leverage intermediate representations and close part of the cross-lingual calibration gap.
(4) Evidence that both findings hold beyond MCQA, in open-ended QA, chain-of-thought reasoning, and alternative MCQA scoring protocols.

\section{Related Work}
 
\paragraph{Multilingual Calibration}  
Recent work has highlighted that modern LLMs, despite their strong performance, frequently struggle with calibration in their predictions \citep{xiong2023can, zhang-etal-2025-atomic}. 
Parallel studies document language-specific biases in LLMs \citep{zhang-etal-2024-need, qin2025survey, zhou-lu-2025-bias, zhou2026lowerresourcehigherscoreslanguage} and a broader linguistic hierarchy in the global AI ecosystem \citep{occhini2026linguistic}.
Yet calibration in multilingual settings remains underexplored. 
\citet{ahuja-etal-2022-calibration} first established that multilingual models like mBERT and XLM-R are poorly calibrated, especially for low-resource languages like Swahili. 
\citet{xue2024mlingconf} conducted a confidence estimation study across various models, covering both language-agnostic and language-specific tasks, but datasets in their study included only 5 languages and were machine-translated which can potentially import bias~\citep{vanmassenhove-etal-2021-machine, choenni2024evaluation}. We extend prior research in two key dimensions. First, we provide more comprehensive evaluation using human-curated datasets across 100+ languages and eight LLM families. Second, while existing studies focus exclusively on final-layer outputs, we investigate the internal architectural origins of calibration through a novel layer-wise analysis.

\paragraph{Layer-wise Representations}
Recent work studies how layers specialise in multilingual transformers. Intermediate layers often encode cross-lingual semantics in a largely language-agnostic shared space \citep{bandarkar-etal-2025-layer, zhou-etal-2026-crosslingual}. By contrast, later layers become more language-specific, adapting these representations to surface features such as syntax and word order. 
For predominantly English-trained LLMs (e.g., LLaMA), studies further suggest that multilingual inputs are mapped in middle layers to shared internal representations that often resemble English-like structures \citep{wendler-etal-2024-llamas, kojima-etal-2024-multilingual, alabi-etal-2024-hidden}. This view helps explain why prompting the model to ``think in English'' can improve cross-lingual generation: it matches the model's internal pathway before decoding back into the target language \citep{shi2022language, zhang-etal-2024-plug}. Building on these findings, we examine how layer-wise specialisation, especially the language-neutrality of intermediate layers, affects calibration across languages.

\begin{table*}[t]
    \centering
    \rowcolors{2}{gray!10}{white}
    \resizebox{.75\textwidth}{!}{
    \begin{tabular}{
        l
        S[table-format=2.2]
        S[table-format=2.2]
        S[table-format=2.2]
        S[table-format=2.2]
        | S[table-format=2.2]
        S[table-format=2.2]
        S[table-format=2.2]
        S[table-format=2.2]
    }
        \toprule
        \textbf{Language}
        & \multicolumn{4}{c}{\textbf{LLaMA3}}
        & \multicolumn{4}{c}{\textbf{Aya}} \\
        \cmidrule(lr){2-5}
        \cmidrule(lr){6-9}
        & \textbf{ECE} $\downarrow$ & \textbf{BRIER} $\downarrow$ & \textbf{AUROC} $\uparrow$ & \textbf{Accuracy} 
        & \textbf{ECE} $\downarrow$ & \textbf{BRIER} $\downarrow$ & \textbf{AUROC} $\uparrow$ & \textbf{Accuracy}  \\
        \midrule
        Arabic       & 33.06 & 24.37 & 61.00 & 38.20 & 28.41 & 33.79 & 71.49 & 45.20 \\
        Bengali      & 24.93 & 23.39 & 58.44 & 35.20 & 29.01 & 31.48 & 60.01 & 31.30 \\
        German       & 25.81 & 24.92 & 65.36 & 44.40 & 26.54 & 33.51 & 69.70 & 53.00 \\
        Spanish      & 18.21 & 21.89 & 71.65 & 52.00 & 28.17 & 31.86 & 71.12 & 51.10 \\
        French       & 13.87 & 22.75 & 71.39 & 51.30 & 23.80 & 32.72 & 70.69 & 53.40 \\
        Hindi        & 28.31 & 24.28 & 62.07 & 39.90 & 30.21 & 34.98 & 70.08 & 42.30 \\
        Indonesian   & 19.67 & 23.76 & 66.25 & 45.00 & 27.88 & 31.54 & 70.85 & 51.20 \\
        Italian      & 21.19 & 22.74 & 71.57 & 51.80 & 26.65 & 30.33 & 71.76 & 52.70 \\
        Japanese     & 28.36 & 27.27 & 61.73 & 43.00 & 16.30 & 26.26 & 69.92 & 46.70 \\
        Korean       & 30.86 & 25.06 & 62.59 & 42.50 & 32.07 & 37.09 & 72.06 & 45.00 \\
        Portuguese   & 10.51 & 21.76 & 71.37 & 50.40 & 27.33 & 31.42 & 70.71 & 53.50 \\
        Swahili      & 23.84 & 21.45 & 61.10 & 32.20 & 32.01 & 36.72 & 58.23 & 31.30 \\
        Yoruba       & 8.18  & 19.43 & 58.00 & 27.40 & 30.11 & 28.56 & 60.73 & 26.40 \\
        Chinese      & 41.94 & 19.56 & 50.63 & 23.10 & 17.12 & 28.75 & 67.35 & 52.20 \\
        \midrule
        English      & \bfseries 4.61 & \bfseries 17.63 & \bfseries 80.36 & \bfseries 61.20 & \bfseries 20.66 &  \bfseries 25.30 & \bfseries 74.65 & \bfseries 57.40 \\
        \textit{Avg. Non-English} &  23.12  &  22.95 & 64.06 &   41.47 &  26.77 &  31.97 &   68.26 &  45.49  \\
        \midrule
        \textit{Avg. Low-Resource}       & 23.00 & 22.78 & 61.14 & 36.32 & 29.60 & 32.84 & 65.23 & 37.95 \\
        \textit{Avg. High-Resource}      & 21.71 & 22.62 & 67.41 & 46.63 & 24.29 & 30.80 & 70.88 & 51.67 \\
        \textit{Avg. Non-Latin-Script}   & 27.44 & 23.10 & 59.44 & 35.19 & 26.90 & 32.20 & 66.23 & 40.05 \\
        \textit{Avg. Latin-Script}       & 16.27 & 22.21 & 71.14 & 50.87 & 25.86 & 30.95 & 71.35 & 53.19 \\
        \midrule
        \textit{Average (All)} &
        22.22 & 22.68 &  64.90 &  42.51 
        & 26.42 &  31.62 & 68.62 &  46.18 \\
        \bottomrule
    \end{tabular}
    }
    \caption{Multilingual performance of \textbf{LLaMA3} (left) and \textbf{Aya} (right) on the MMMLU dataset. Metrics include ECE, Brier Score, AUROC, and Accuracy. All numbers are in percentages. }
    \label{tab:llama3_vs_cohere_mmmlu}
\end{table*}

\section{Benchmarking Multilingual Calibration}
\label{sec:benchmarking}

\subsection{Experimental Setup}

\paragraph{Task and Confidence Elicitation.}
We focus on multilingual Multiple-Choice Question Answering (MCQA) for two reasons. First, MCQA provides \emph{unambiguous correctness labels}: in open-ended generation, correctness is approximated via metrics such as ROUGE, BLEU, or LLM-as-a-judge, which often disagree and can flip conclusions depending on the metric \citep{ielanskyi2025addressing}. Second, MCQA admits a \emph{clear definition of confidence} as the probability assigned to the selected option, whereas open-ended confidence estimation is method-dependent and unstable. Concretely, for each instance $x$ with $C$ choices we take the model's prediction $\hat y = \arg\max_i p_i$ at the final layer and elicit confidence as $\mathrm{Conf}(x) = \max_{i \in \{1,\dots,C\}} p_i$. We later validate that our findings generalise beyond MCQA on generative tasks (\S\ref{sec:generative}) and across alternative MCQA scoring protocols (Appendix~\ref{sec:protocol_robustness}).

\paragraph{Datasets and Models.}
We use two high-quality, human-translated benchmarks: (1)~\textbf{MMMLU} \citep{hendrycks2020mmmlu} (15 languages) and (2)~\textbf{Belebele} \citep{Bandarkar_2024} (122 languages). Together they cover a much larger range of languages and instances ($\sim 10^5$) than prior multilingual calibration studies. All experiments use eight-shot prompting in the respective language. We evaluate eight model families spanning 7--14B parameters: \textbf{LLaMA3}, \textbf{Qwen2.5}, \textbf{Qwen3}, \textbf{Mistral}, \textbf{Aya}, \textbf{DeepSeek}, \textbf{Phi}, and \textbf{Olmo3}. Full model identifiers and additional setup details are in Appendix~\ref{sec:appendix_eva}.

\paragraph{Metrics}
We evaluate calibration using expected calibration error (ECE; \citealp{guo2017calibration}) and the Brier score \citep{brier1950verification}. To measure model's ability to discriminate between correct and incorrect predictions, we also report AUROC \citep{FAWCETT2006861}. 
We report ECE with number of bins $K = 10$; for bin size robustness, see Appendix \ref{sec:ECE binning}.

\subsection{Results}

Table~\ref{tab:llama3_vs_cohere_mmmlu} reports per-language MMMLU results for \textbf{LLaMA3} and \textbf{Aya}; Table~\ref{tab:cross_family} extends the comparison to the English-vs.-Non-English gap across all eight model families.

\begin{table}[t]
    \centering
    \footnotesize
    \setlength{\tabcolsep}{4pt}
    \renewcommand{\arraystretch}{1.1}
    \rowcolors{2}{gray!10}{white}
    \resizebox{\columnwidth}{!}{%
    \begin{tabular}{l
        S[table-format=2.2]
        S[table-format=2.2]
        c
        S[table-format=2.2]
        S[table-format=2.2]
        c
    }
        \toprule
        \multirow{2}{*}{\textbf{Model}}
        & \multicolumn{2}{c}{\textbf{ECE}$\downarrow$}
        & \multirow{2}{*}{\textbf{$\Delta$}}
        & \multicolumn{2}{c}{\textbf{AUROC}$\uparrow$}
        & \multirow{2}{*}{\textbf{$\Delta$}} \\
        \cmidrule(lr){2-3}\cmidrule(lr){5-6}
        & \textbf{EN} & \textbf{Non-EN} &
        & \textbf{EN} & \textbf{Non-EN} & \\
        \midrule
        LLaMA3       &  4.61 & 23.12 & {-18.51} & 80.36 & 64.06 & {+16.30} \\
        Aya          & 20.66 & 26.77 &  {-6.11} & 74.65 & 68.26 &  {+6.39} \\
        Mistral      & 23.92 & 34.24 & {-10.32} & 73.75 & 63.67 & {+10.08} \\
        Qwen2.5      & 15.77 & 21.22 &  {-5.45} & 78.23 & 70.18 &  {+8.05} \\
        Phi          & 20.48 & 27.05 &  {-6.57} & 71.13 & 57.41 & {+13.72} \\
        DeepSeek     &  9.10 & 26.21 & {-17.11} & 66.21 & 57.14 &  {+9.07} \\
        Qwen3        & 12.09 & 13.65 &  {-1.56} & 82.26 & 73.41 &  {+8.85} \\
        Olmo3        & 28.65 & 27.24 &  {+1.41} & 64.70 & 56.37 &  {+8.33} \\
        \bottomrule
    \end{tabular}%
    }
    \caption{Per-model MMMLU calibration. $\Delta = \textrm{EN}-\textrm{Non-EN}$ for both metrics. Per-language results are in Table~\ref{tab:llama3_vs_cohere_mmmlu} and Tables~\ref{tab:mistral_mmmlu_multilingual_metrics}--\ref{tab:deepseekq_mmmlu_multilingual_metrics}.}
    \label{tab:cross_family}
\end{table}

\begin{figure*}[!h]
    \centering
    \includegraphics
    [width= \linewidth]
    {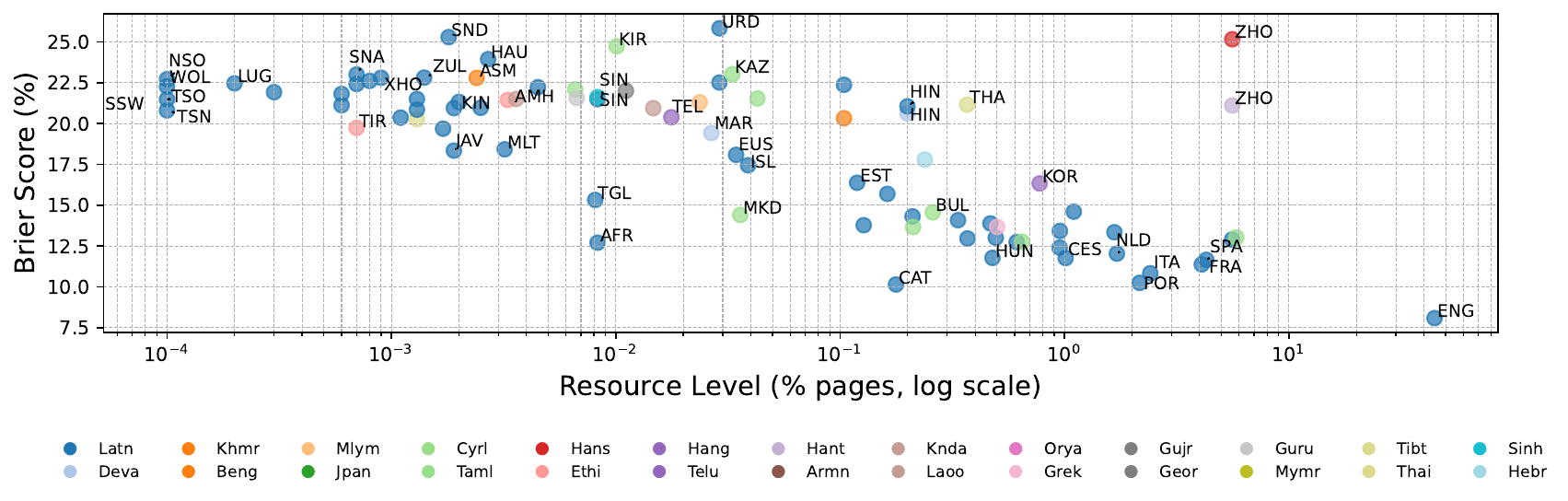}
    \caption{Brier score vs.\ resource level for LLaMA3 on Belebele. Each point is one language; colour indicates writing system. Correlations: Spearman $\rho=-0.59$, $p<10^{-8}$; Kendall $\tau=-0.43$, $p<10^{-8}$; Pearson $r=-0.39$, $p<0.001$.}
    \label{fig:belebele_llama3_ece_resource}
\end{figure*}

\paragraph{Not only are LLMs more accurate but also more calibrated in English.}
Table~\ref{tab:llama3_vs_cohere_mmmlu} shows that non-English languages underperform English in both accuracy and calibration: LLaMA3's average ECE is 23.1\% on non-English against 4.6\% on English. The same pattern appears in Brier Score and AUROC and holds across all eight model families: Table~\ref{tab:cross_family} reports a positive AUROC $\Delta$ for every model, and a negative ECE $\Delta$ for every model except Olmo3, which is poorly calibrated in English as well. The size of the ECE gap tracks multilingual pre-training coverage: Qwen3, pre-trained on 36T tokens spanning 119 languages \citep{qwen3-2025}, shows the smallest English-favouring gap (1.56), while Qwen2.5, whose 18T-token corpus covers 29 languages \citep{yang2024qwen2technicalreport}, shows 5.45; this within-family comparison partially controls for general model quality. English-first LLaMA3 shows the largest gap (18.51). Bootstrap confidence intervals for Table~\ref{tab:cross_family} are in Appendix~\ref{sec:bootstrap_ci}.

\paragraph{Calibration correlates with language resource availability.}
The calibration gap also tracks resource availability: in Table~\ref{tab:llama3_vs_cohere_mmmlu}, Hindi and Swahili are worse calibrated than the high-resource group. Plotting resource level\footnote{LLaMA3 does not release its exact training data; we use Common Crawl (\texttt{CC-MAIN-2025-30}; \citealp{commoncrawl}) per-language page percentages as a proxy.} against calibration on Belebele (Figure~\ref{fig:belebele_llama3_ece_resource}; per-language tables in Appendix~\ref{sec:belebele_results}) yields strong negative correlations with Brier (Spearman $\rho{=}{-}0.59$, $p{<}10^{-9}$) and positive correlations with AUROC ($\rho{=}0.66$, $p{<}10^{-12}$), suggesting that calibration is associated with the representation of a language in the pre-training corpus.

\begin{figure}[h]
    \centering
    \begin{subfigure}[b]{0.47\textwidth}
        \centering
        \includegraphics[width=\textwidth]{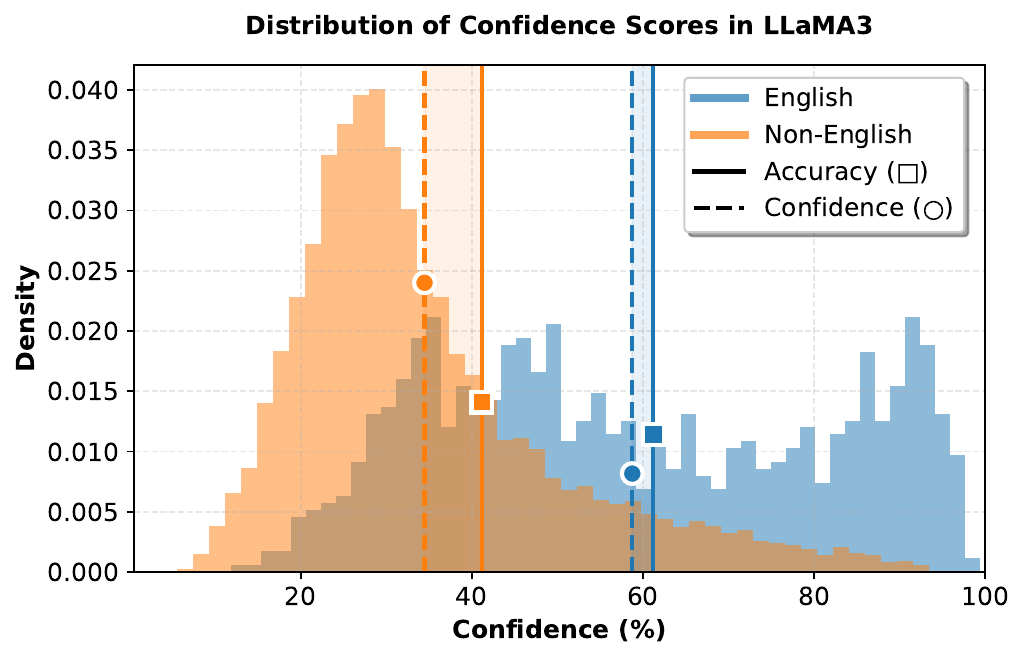}
        \caption{LLaMA3 confidence distribution}
        \label{fig:llama}
    \end{subfigure}
    \hfill
    \begin{subfigure}[b]{0.47\textwidth}
        \centering
        \includegraphics[width=\textwidth]{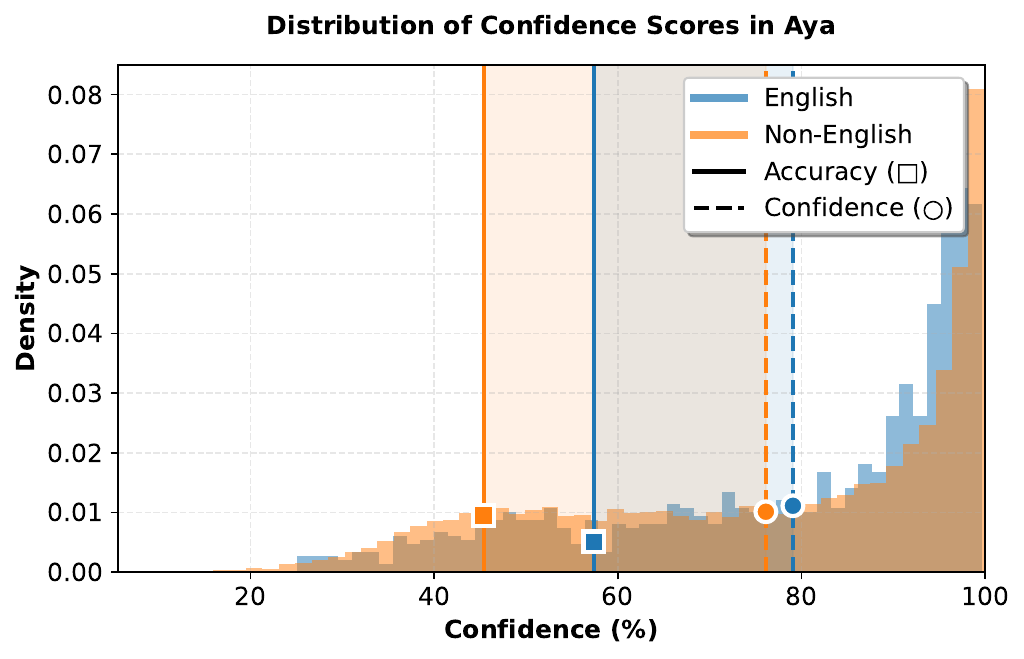}
        \caption{Aya confidence distribution}
        \label{fig:aya}
    \end{subfigure}
    \caption{Density of model confidence scores for \textcolor{blue}{English} and \textcolor{orange}{Non-English} samples. Dashed lines mark mean confidence; solid lines mark accuracy.}
    \label{fig:combined}
\end{figure}

\begin{figure*}[h]
    \centering
    \includegraphics[width=0.9\linewidth]{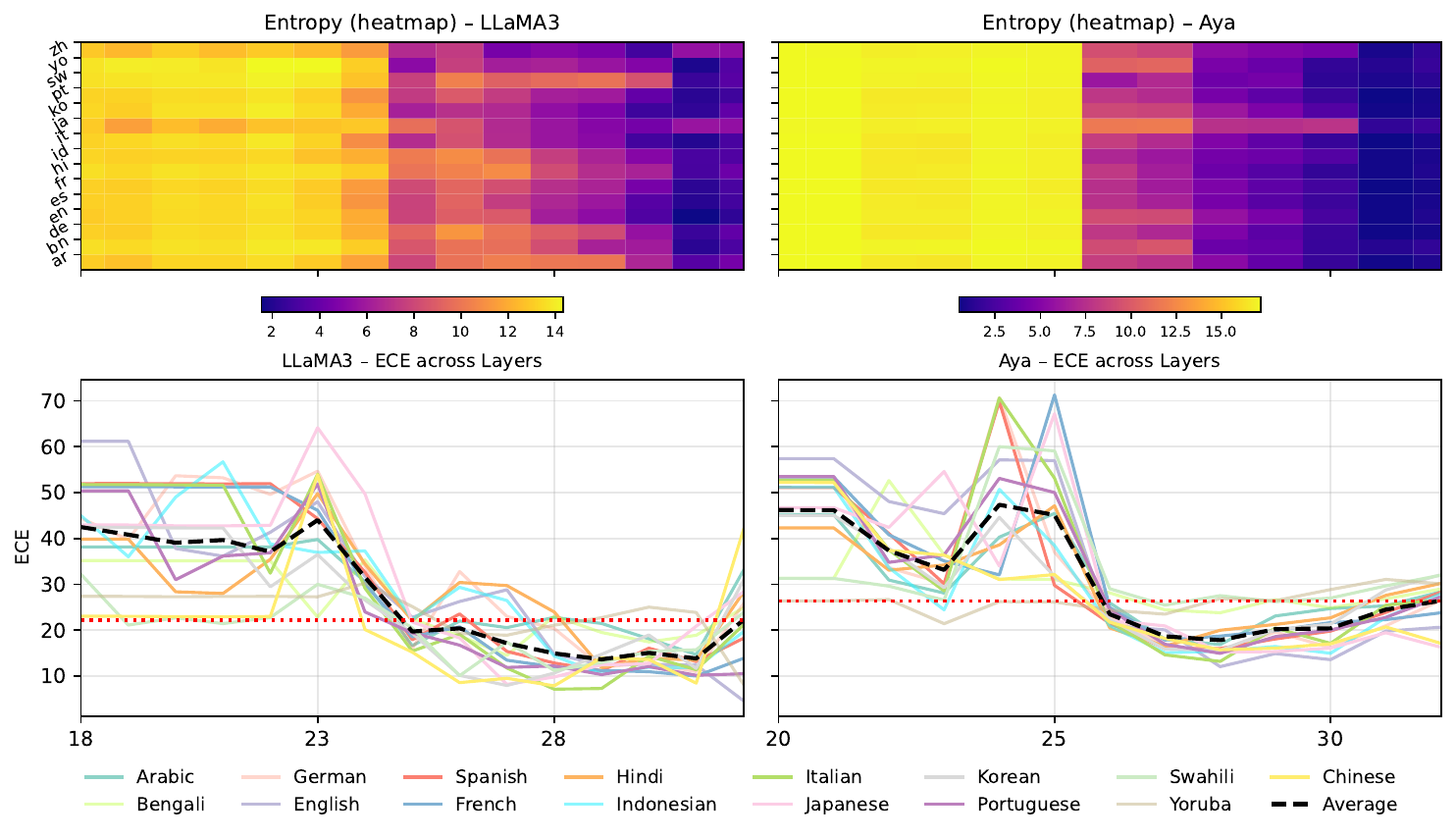}
    \caption{ECE vs.\ entropy across layers on the MMMLU subset for LLaMA3 and Aya. See Figure~\ref{fig:english_llama3_calibration_vs_entropy_ece} for the English-only curve.}
    \label{fig:mmmlu_llama3_calibration_vs_entropy_ece}
\end{figure*}

\paragraph{Miscalibration patterns reflect training and alignment priorities.}
Confidence distributions reveal two regimes. For LLaMA3 (Figure~\ref{fig:llama}), the English setting is well-calibrated (ECE = 4.6) while the non-English setting shows underconfidence. Aya's distributions in English and non-English are similar in shape (Figure~\ref{fig:aya}) but right-skewed (overconfident) in both. These patterns track training priorities. LLaMA3's alignment pipeline (supervised fine-tuning, preference ranking, safety) \citep{grattafiori2024llama} is primarily English-focused. Aya prioritises multilingual coverage \citep{ustun-etal-2024-aya, dang2024ayaexpansecombiningresearch} with less attention to calibration. The two regimes echo the \textit{Square One Bias} \citep{ruder-etal-2022-square}: progress along either axis (English alignment or multilingual coverage) leaves the orthogonal dimension under-served. Confidence behaviour of other models is in Appendix~\ref{sec:behaviour}.

\section{Mid-Layers Reveal Better Calibration}
\label{sec:layerwise}

Inspired by recent insights in layer-wise multilingual representations~\citep{bandarkar-etal-2025-layer, wendler-etal-2024-llamas}, we examine how confidence evolves throughout the model's depth to understand the source of the poor calibration observed in \S \ref{sec:benchmarking}. 
Specifically, we ask: \textit{is it possible to identify intermediate representations that elicit better calibrated confidence scores w.r.t. final-layer accuracy?}

\subsection{Early-Decoded Confidence Estimation}
\label{sec: methodology}
We adopt a layer-wise probing technique inspired by the early exiting paradigm \citep{elbayad2020depthadaptivetransformer} to offer a new way of confidence estimation. Instead of applying the modelling head only to the final hidden state, we attach it to each intermediate transformer layer. This allows us to extract logits and compute prediction confidence from every layer, providing a granular view of the model's decision-making process. 

Formally, let $\mathbf{h}_\ell \in \mathbb{R}^d$ denote the hidden representation at layer $\ell$, where $\ell = 1, \dots, L$, and $d$ is the dimensionality of the hidden state. We apply the original language modelling head, with weight matrix $W \in \mathbb{R}^{V \times d}$, to compute the logits at each layer:
\[
\mathbf{z}_\ell = W \mathbf{h}_\ell
\]
where $\mathbf{z}_\ell \in \mathbb{R}^V$ are the unnormalised token logits over the vocabulary of size $V$. These logits are then converted into probabilities using the softmax function:
\[
\mathbf{p}_\ell = \mathrm{softmax}(\mathbf{z}_\ell), \quad
\sum_{v=1}^{V} [\mathbf{p}_\ell]_v = 1 .
\]
With the $\mathbf{p}_\ell$, we \emph{trace} the token ultimately predicted at the final layer, $\hat{y}_L = \arg\max_{v} [\mathbf{p}_L]_v$, back through the intermediate layers. At each layer $\ell$, we then define the confidence score as the probability mass that this layer assigns to the final prediction $\text{Conf}_\ell(x) = [\mathbf{p}_\ell]_{\hat{y}_L}$. Calibration is then evaluated by comparing these $\text{Conf}_\ell(x)$ with the prediction accuracy determined at the final layer $\hat{y}_L$:
\[
\text{ECE}_\ell 
= \text{ECE}\Big(\{ (\text{Conf}_\ell(x), \; \mathbf{1}\{\hat{y}_L = y\}) \}\Big).
\]
where $\mathbf{1}\{\hat{y}_L = y\}$ is the indicator function of whether the final-layer prediction is correct.

\subsection{Layer-wise Multilingual Calibration Results}

\paragraph{Multilingual settings reveal best calibration in late intermediate layers.}
The conventional understanding \citep{tenney2019bertrediscoversclassicalnlp} holds that intermediate representations become progressively more refined and task-specific, so calibration should improve with depth and peak at the final layer; we observe this pattern in the English-only setting (Appendix~\ref{appendix:english_gets_better}). However, the multilingual picture differs: as Figure~\ref{fig:mmmlu_llama3_calibration_vs_entropy_ece} shows, many languages achieve their best ECE in the late-intermediate layers (from layer 24 for LLaMA3, and from layer 26 for Aya, both 32-layer models), with calibration worsening again towards the final output layer. The turning point aligns with the entropy trend: as entropy drops sharply in these intermediate layers, ECE also decreases. Reliability diagrams are in Appendix~\ref{sec:reliablility}.

\paragraph{Why are late-intermediate layers better calibrated?}
Three observations point to a consistent interpretation. First, we see English calibration improves monotonically with depth while non-English calibration peaks before the final layer; Second, the region of late-intermediate layers is where prior work locates the most language-agnostic representations \citep{bandarkar-etal-2025-layer, wendler-etal-2024-llamas}, and where models develop alignable geometry \citep{zhou-etal-2026-crosslingual}; a shared representation space supports a confidence signal that transfers across languages. Third, alignment updates act most directly on the output distribution, and the models with the most English-centric alignment show the largest gaps (\S\ref{sec:benchmarking}). Together these are consistent with the final layer re-specialising towards English output distributions at the cost of multilingual confidence.

\paragraph{The mid-layer calibration peak is robust across model sizes and families.}
The same pattern holds across all eight model families we analyse layer-wise (Appendix~\ref{sec:multicalib}) and on LLaMA-3.1-70B (Appendix~\ref{sec:70b}): for most non-English languages, a late-intermediate layer yields lower ECE than the decoding layer. This questions the common practice of using final-layer probabilities as confidence, and motivates the layer-aware calibration strategies developed next.

\section{Improving Multilingual Calibration}
\label{Improving Multilingual Calibration}
\label{sec:method}

\subsection{Multilingual Calibration Methods}

Our observations in Section~\ref{sec:layerwise} motivate confidence estimators that exploit intermediate representations. We explore to extract confidence in three different ways:

\begin{itemize}[leftmargin=1em, labelsep=0.4em, itemsep=1pt, topsep=-1pt, parsep=0pt]

    \item \textbf{Final layer (baseline).}  
    We follow the standard practice in prior work by using the model’s final layer to derive probabilities.

    \item \textbf{Best layer.}  
    We identify the \emph{best layer} as the one that achieves the lowest average calibration error across languages. From this layer, we extract probability estimates following the procedure in Section~\ref{sec: methodology}. The best layer $\ell^*$ is selected using a separate validation set and defined as: 
    
    {\small
    \[
        \ell^* \;=\; \arg\min_{\ell \in \{1,\dots,L\}} \mathrm{ECE}^{\text{avg}}_{\ell}.
    \]
    }
    \item \textbf{Good layers ensemble.}  
    We consider the set of layers (\textit{good layers}) whose multilingual calibration outperforms the final layer. To obtain confidence estimates, we average the predictive distributions across these layers $\mathcal{G}$, this reduces layer-specific noise while pooling calibration-aware signals:
    
    {\small
    \[
        \mathcal{G} \;=\; \bigl\{\ell : \mathrm{ECE}^{\text{avg}}_{\ell} < \mathrm{ECE}^{\text{avg}}_{L}\bigr\},   \
        \mathbf{p}_{\text{ensemble}} \;=\; \frac{1}{|\mathcal{G}|}\sum_{\ell \in \mathcal{G}} \mathbf{p}_{\ell}.
    \]
    }
\end{itemize}

\paragraph{Combining with Classical Post-hoc Calibration.}
Since confidence elicitation and calibration are orthogonal approaches, we further test whether the proposed elicitation methods can be enhanced by standard post-hoc calibration techniques such as Temperature Scaling~\citep{guo2017calibration} and Isotonic Regression~\citep{zadrozny2002transforming}, which operate
independently of how probabilities were obtained~\citep{kadavath2022language,minderer2021revisiting}. We adopt a two-stage pipeline:

{\small
\[
    \mathbf{p}_{\text{final}} \;=\; \text{Calibrate}\!\left(\mathbf{p}_{\text{intermediate}}\right),
\]
}
where $\mathbf{p}_{\text{intermediate}}$ comes from the confidence scores discussed above. See Appendix~\ref{sec:baseline} for details.

\paragraph{Language-Aware Confidence Ensemble.}
The methods above optimise jointly across languages. \emph{LACE} (Language-Aware Confidence Ensemble) instead tailors layer selection to each language. For each language $k$, we select the layers that are better calibrated than the final layer,

{\small
\[
\begin{aligned}
&\mathcal{G}^{(k)} \;=\; \{\ell \;:\; \mathrm{ECE}^{(k)}_{\ell} < \mathrm{ECE}^{(k)}_{L}\},\\[2pt]
&\mathbf{p}^{(k)}_{\text{ensemble}} \;=\; \tfrac{1}{|\mathcal{G}^{(k)}|}\sum_{\ell \in \mathcal{G}^{(k)}} \mathbf{p}^{(k)}_{\ell},
\end{aligned}
\]
}

and apply a language-specific post-hoc calibrator:

{\small
\[
\mathbf{p}^{(k)}_{\text{final}} \;=\; \text{Calibrate}^{(k)}\!\bigl(\mathbf{p}^{(k)}_{\text{ensemble}}\bigr).
\]
}

Per-language layer selection avoids negative transfer from layers that are miscalibrated for the target language, and ensembling reduces variance while preserving language-relevant signals. LACE is training-free and adds negligible latency: it reuses intermediate logits from the single forward pass already required for generation, with no retraining or weight modification.

\paragraph{Experiment Setup} We use the data from Section~\ref{sec:layerwise} as a held-out validation set and evaluate on a separate MMMLU test split. Both splits are balanced across languages, with a total of 30K examples. For Belebele, we construct a comparable evaluation set with 24K examples overall with a similar validation/test split. We report macro-averaged results across languages.

\subsection{LACE Results}

\begin{table*}[t]
    \centering
    \renewcommand{\arraystretch}{1.4}
    \resizebox{0.8\textwidth}{!}{
    \begin{tabular}{lcccccc|cccccc}
        \toprule
         & \multicolumn{6}{c}{\textbf{MMMLU}} & \multicolumn{6}{c}{\textbf{Belebele}} \\
         & \multicolumn{3}{c}{\textbf{LLaMA3} (Acc. = 43.2\%)} & \multicolumn{3}{c|}{\textbf{Aya} (Acc. = 48.8\%)}
         & \multicolumn{3}{c}{\textbf{LLaMA3} (Acc. = 68.6\%)} & \multicolumn{3}{c}{\textbf{Aya} (Acc. = 67.8\%)} \\
        \cmidrule(lr){2-4}\cmidrule(lr){5-7}\cmidrule(lr){8-10}\cmidrule(lr){11-13}

         \textbf{Method} & \textbf{ECE} & \textbf{Brier} & \textbf{AUROC} 
                         & \textbf{ECE} & \textbf{Brier} & \textbf{AUROC} 
                         & \textbf{ECE} & \textbf{Brier} & \textbf{AUROC} 
                         & \textbf{ECE} & \textbf{Brier} & \textbf{AUROC} \\
        \midrule
        \rowcolor{gray!15}
        \textsc{Final Layer Baseline}$^\dagger$ 
            & 22.44 & 23.03 & 64.05 
            & 24.39 & 30.45 & 68.31 
            & 17.68 & 20.38 & 69.01 
            & 15.73 & 19.08 & \textbf{72.92} \\
        \phmark \quad \footnotesize \textit{(+Temperature Scaling)} 
            & 23.35 & 22.59 & 64.05 
            & 15.40 & 23.76 & 68.31 
            & 17.63 & 20.39 & 69.01 
            & 10.66 & 17.19 & 72.92 \\
        \phmark \quad \footnotesize \textit{(+Isotonic Regression)} 
            & 20.23 & 22.55 & 63.47 
            & 9.15  & 22.02 & 68.07 
            & 11.09 & 19.07 & 68.85 
            & 8.33  & 16.44 & 72.78 \\
        \midrule
        \multicolumn{13}{c}{\textit{Intermediate Layer Calibration (Global Selection)}} \\
        \midrule
        \rowcolor{gray!15}
        \textsc{Best Layer}$^\dagger$ 
            & 14.28 & 22.78 & 71.44 
            & 17.57 & 27.08 & 66.68 
            & 13.67 & 18.28 & 71.33 
            & 15.40 & 20.06 & 66.79 \\
        \phmark\quad \footnotesize \textit{(+Temperature Scaling)} 
            & 13.71 & 20.60 & 71.44  
            & 9.34  & 22.52 & 66.68 
            & 13.12 & 17.52 & 71.34 
            & 14.99 & 19.16 & 66.79 \\
        \phmark\quad  \footnotesize \textit{(+Isotonic Regression)} 
            & 13.12 & 20.80 & 71.28 
            & 10.66 & 22.46 & 66.26 
            & 12.40 & 17.39 & 71.48 
            & 14.02 & 19.00 & 66.80 \\
        \midrule
        \rowcolor{gray!15}
        \textsc{Good Layers Ensemble}$^\dagger$ 
            & 11.84 & 20.23 & \textbf{73.91} 
            & 13.10 & 23.78 & \textbf{68.62 }
            & 10.78 & 15.59 & \textbf{76.26 }
            & 11.47 & 18.00 & 70.24 \\
        \phmark\quad \footnotesize \textit{(+Temperature Scaling)} 
            & 11.30 & 20.01 & 73.91 
            & 10.23 & 22.21 & 68.62 
            & 10.49 & 15.62 & 76.26 
            & 11.81 & 17.57 & 70.24 \\
        \phmark\quad \footnotesize \textit{(+Isotonic Regression)} 
            & 9.60 & 19.90 & 73.49 
            & 7.71  & 21.82 & 68.25
            & 10.16 & 15.54 & 76.07 
            & 10.42 & 17.48 & 70.31 \\
        \midrule
        \multicolumn{13}{c}{\textit{Intermediate Layer Calibration (LACE)}} \\
        \midrule
        \rowcolor{gray!15}
        \textsc{Language-Aware Ensemble}$^\ddagger$  
            & 5.96 & 20.51 & 72.94 
            & 11.42 & 22.70 & 68.38 
            & 7.05  & \textbf{14.35} & 75.61 
            & 10.22 & 17.77 & 69.79 \\
        \phmark\quad \footnotesize \textit{(+Temperature Scaling)}   
            & 4.34 & \textbf{19.73} & 73.40 
            & 4.88 & 21.87 & 68.49 
            & 6.05  & 14.47 & 74.98 
            & 5.46  & \textbf{16.36} & 70.45 \\
        \phmark\quad \footnotesize \textit{(+Isotonic Regression)} 
            & \textbf{3.09} & 20.51 & 69.13 
            & \textbf{3.45} & \textbf{21.46} & 67.10 
            & \textbf{5.79}  & 14.53 & 73.70 
            & \textbf{4.80 } & 16.73 & 68.40 \\           
        \bottomrule

    \end{tabular}
    }
    \caption{Calibration results for \textbf{LLaMA3} and \textbf{Aya} on MMMLU (left) and the Belebele subset (right). 
    $^\dagger$ denote methods that do not assume access to language identity. 
    $^\ddagger$ denote methods with given language identity. 
Indented italic rows correspond to post-hoc calibration adjustments.}
    \label{tab:llama3_cohere_rescheduled}
\end{table*}

\begin{figure}
    \centering
    \includegraphics[width=0.95\columnwidth]{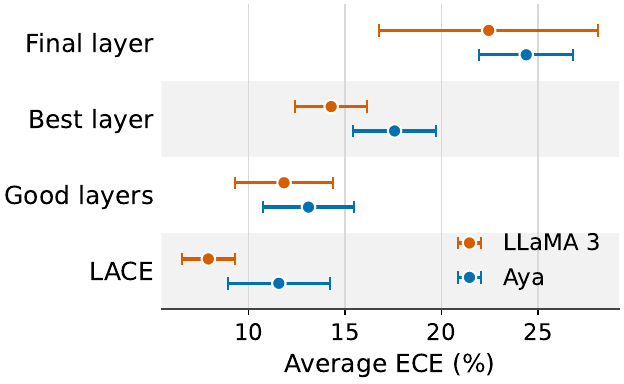}
    \caption{Forest plot of average ECE in MMMLU, with means and 95\% CIs.}
    \label{fig:ece_forestplot}
\end{figure}

\paragraph{Intermediate-layer confidence shows better calibration than final-layer confidence.}
Figure~\ref{fig:ece_forestplot} shows that moving from Best Layer to Good-Layers Ensemble to \textsc{LACE} yields progressive improvements; full numbers for MMMLU and Belebele are in Table~\ref{tab:llama3_cohere_rescheduled}. Occasional drops in AUROC arise because discrimination and calibration are not necessarily correlated \citep{gao2022comparativestudyconfidencecalibration, carriero-etal-2025-harms} and our layer selection optimises ECE. By tailoring layer selection and calibration to each language, \textsc{LACE} achieves the best overall results, with ECE as low as 5.96\% on LLaMA3.

\paragraph{Classical post-hoc methods offer improvements but face challenges in multilingual settings.}
Temperature scaling and isotonic regression noticeably reduce calibration error on Aya, which is systematically overconfident (avg.\ confidence $76.9\%$ on MMMLU, $82.5\%$ on Belebele; Figure~\ref{fig:combined}). However, they offer marginal benefit or even degradation for LLaMA3 (avg.\ confidence $36.0\%$ / $58.3\%$): a globally fit temperature gives limited per-language gains under distributional heterogeneity (cf.\ \citealp{simpson1951interpretation}).

\paragraph{LACE is complementary to post-hoc approaches and delivers the best calibration performance.}
Importantly, our method \textit{reshapes}, rather than \textit{rescales}, the calibration signal.
Therefore, our method does not compete with post-hoc calibrators but provides orthogonal improvements. As shown in Table~\ref{tab:llama3_cohere_rescheduled}, all methods combined with temperature scaling or isotonic regression yields further incremental gains, and notably \textsc{LACE} is further boosted to consistently deliver the lowest calibration error across benchmarks: ECE to $3.09\%$ on LLaMA3 and $3.45\%$ on Aya. 

\subsection{Analysis}

We confirm that \textsc{LACE} scales to larger models, and analyse robustness.

\begin{table}[t]
    \centering
\footnotesize
\resizebox{0.7\columnwidth}{!}{%
\begin{tabular}{lccc}
\toprule
\textbf{Method} & \textbf{ECE} & \textbf{Brier} & \textbf{AUROC} \\
\midrule
\rowcolor{gray!15}
\textsc{Orig. Prob.}
    & 23.51 & 22.10 & 65.19 \\
\phantom{a}\footnotesize\textit{(+Temp. Scal.)}
    & 22.56 & 21.97 & 65.19 \\
\phantom{a}\footnotesize\textit{(+Isotonic)}
    & 19.01 & 21.34 & 65.07 \\
\midrule
\rowcolor{gray!15}
\textsc{LACE}
    & 6.14 & 15.76 & 79.42 \\
\phantom{a}\footnotesize\textit{(+Temp. Scal.)}
    & 4.93 & 15.60 & 79.92 \\
\phantom{a}\footnotesize\textit{(+Isotonic)}
    & 2.81 & 15.38 & 78.79 \\
\bottomrule
\end{tabular}%
}
\caption{Calibration for LLaMA-3.1 70B on MMMLU.}
\label{tab:70b_calibration_main}
\end{table}

\paragraph{LACE scales effectively to larger models.}
Consistent with our findings at the 7--14B scales, our layer-based methods continue to outperform final-layer confidence on LLaMA-3.1-70B (Table~\ref{tab:70b_calibration_main}; full results in Appendix~\ref{sec:70b}). \textsc{LACE} provides the largest overall improvement at this scale, reducing ECE from 23.51\% (baseline) to as low as 2.81\% when combined with Isotonic Regression.

\begin{figure}
    \centering
    \includegraphics[width=0.66\columnwidth]{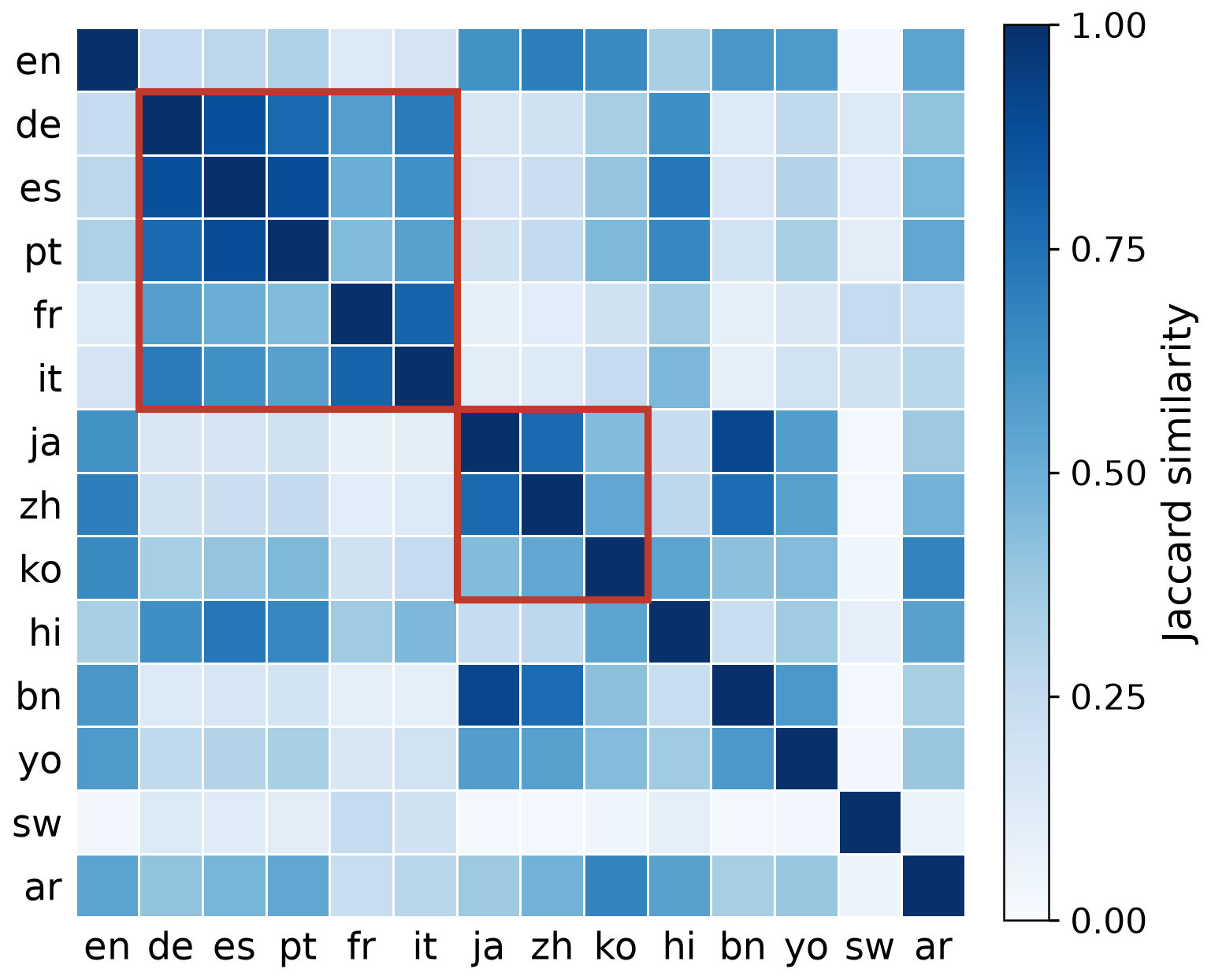}
    \caption{Jaccard Similarity between different language's layer-sets in LLaMA-3.1 70B.}
    \label{fig:similarity_70b_main}
\end{figure}

\paragraph{Layer-set overlap and robustness across languages and datasets.} To assess whether LACE's layer sets generalise beyond specific samples or datasets, we examine their consistency across languages and benchmarks. Similar languages share overlapping layer subsets: Figure~\ref{fig:similarity_70b_main} shows clusters of high overlap between Western European languages and CJK (Chinese, Japanese, Korean) languages (red boxes). Layer sets derived from Belebele also retain strong calibration when transferred to MMMLU. The overlapping structure makes LACE robust to language-ID noise: in a controlled corruption experiment that randomly replaces the language tag of test examples, LACE degrades only minimally across all metrics. Full discussion is in Appendix~\ref{sec:lace_details}.

\section{Generalisation Beyond MCQA}
\label{sec:generative}

Our main analysis focuses on MCQA because it offers unambiguous correctness labels and a well-defined notion of confidence (\S\ref{sec:benchmarking}). To probe whether our two core findings (the multilingual calibration gap and the late-intermediate sweet spot) generalise to free-form generation, we extend the analysis to two generative benchmarks.

\paragraph{Setup.}
We use \textbf{MKQA} \citep{longpre-etal-2021-mkqa} (open-ended multilingual QA) and \textbf{MGSM} \citep{shi2022language} (multilingual chain-of-thought math) on \textbf{LLaMA3}. Correctness is scored against gold answers: for MKQA we apply Unicode-aware normalisation and require gold-answer containment; for MGSM we extract the final numeric token and require an exact match. Following \citet{lewislim2025analysing}, we evaluate four standard generative-confidence elicitation methods: \emph{Perplexity} (geometric mean of next-token probabilities), \emph{Margin} (top-1 vs.\ top-2 softmax gap, averaged across generated tokens), \emph{P(True)} \citep{kadavath2022language} (re-prompted true/false probability), and \emph{Verbalised} \citep{tian-etal-2023-just} (expected confidence on a 1--9 scale, mapped to $[0,1]$). Further details are in Appendix~\ref{sec:generative_setup}.

\begin{table}[t]
    \centering
    \footnotesize
    \setlength{\tabcolsep}{4pt}
    \renewcommand{\arraystretch}{1.1}
    \resizebox{\columnwidth}{!}{%
    \begin{tabular}{l l
        S[table-format=2.1]
        S[table-format=2.1]
        S[table-format=2.1]
        S[table-format=2.1]
    }
        \toprule
        \textbf{Dataset} & \textbf{Group} & {\textbf{Perp.}} & {\textbf{Margin}} & {\textbf{P(True)}} & {\textbf{Verb.}} \\
        \midrule
        \rowcolor{gray!10}
        \multirow{3}{*}{MKQA}
            & English          & \bfseries 13.5 & \bfseries 10.6 & \bfseries 19.2 & \bfseries 26.9 \\
            & Avg.\ Non-Eng.   & 33.9 & 34.8 & 31.7 & 41.1 \\
        \rowcolor{gray!10}
            & $\Delta$         & {-20.4} & {-24.2} & {-12.5} & {-14.2} \\
        \midrule
        \rowcolor{gray!10}
        \multirow{3}{*}{MGSM}
            & English          & \bfseries 18.9 & \bfseries 15.9 & 11.0 & \bfseries 12.8 \\
            & Avg.\ Non-Eng.   & 42.9 & 41.1 & \bfseries 12.6 & 23.8 \\
        \rowcolor{gray!10}
            & $\Delta$         & {-24.0} & {-25.2} &  {+1.6} & {-11.0} \\
        \bottomrule
    \end{tabular}%
    }
    \caption{Generative-task calibration on \textbf{LLaMA3}: ECE (\%, $\downarrow$) for four confidence elicitation methods on MKQA and MGSM. $\Delta = \textrm{EN} - \textrm{Non-EN}$; negative values mean English is better-calibrated. Per-language numbers are in Appendix~\ref{sec:generative_full}.}
    \label{tab:generative_main}
\end{table}

\paragraph{The multilingual calibration gap persists in generation.}
Table~\ref{tab:generative_main} reports group-averaged ECE for each confidence method on the two datasets. English is the best-calibrated language for most cells. Averaging across the four methods, English ECE is 17.6\% on MKQA and 14.7\% on MGSM, while non-English ECE is 35.4\% and 30.1\%, roughly $2\times$ worse on both datasets. The gap is widest on the lowest-resource languages (Japanese on MKQA, avg.\ ECE 47.8\%; Bengali on MGSM, avg.\ ECE 46.7\%), mirroring the resource correlation we report on MCQA (\S\ref{sec:benchmarking}).

\begin{table}[t]
    \centering
    \footnotesize
    \setlength{\tabcolsep}{4pt}
    \renewcommand{\arraystretch}{1.1}
    \resizebox{\columnwidth}{!}{%
    \begin{tabular}{l l
        S[table-format=2.1]
        S[table-format=2.1]
        c
        S[table-format=+2.1]
    }
        \toprule
        \textbf{Method} & \textbf{Lang} & {\textbf{Acc.}} & {\textbf{Last}} & {\textbf{Best ($\ell^\star$)}} & {\textbf{$\Delta$ECE}} \\
        \midrule
        \rowcolor{gray!10}
            & en & 64.0 & 18.0 & 11.9 ($\ell{=}29$)  & -6.1  \\
            & de & 55.0 & 26.2 & 9.9 ($\ell{=}29$)   & -16.3 \\
        \rowcolor{gray!10}
            & es & 55.0 & 26.7 & 6.0 ($\ell{=}29$)   & -20.7 \\
            & fr & 41.0 & 42.5 & 5.1 ($\ell{=}27$)   & -37.4 \\
        \rowcolor{gray!10}
            & ru & 53.0 & 30.3 & 8.2 ($\ell{=}30$)   & -22.1 \\
            & ja & 45.0 & 36.9 & 6.2 ($\ell{=}28$)   & -30.7 \\
        \rowcolor{gray!10}
            & th & 41.0 & 42.7 & 9.0 ($\ell{=}27$)   & -33.7 \\
        \multirow{-8}{*}{Margin}
            & bn & 18.0 & 66.0 & 1.6 ($\ell{=}23$)   & -64.4 \\
        \midrule
        \rowcolor{gray!10}
            & en & 64.0 & 15.4 & 15.4 ($\ell{=}32$)  & 0.0   \\
            & de & 55.0 & 29.1 & 10.7 ($\ell{=}31$)  & -18.4 \\
        \rowcolor{gray!10}
            & es & 55.0 & 28.7 & 11.3 ($\ell{=}31$)  & -17.5 \\
            & fr & 41.0 & 43.2 & 23.1 ($\ell{=}31$)  & -20.1 \\
        \rowcolor{gray!10}
            & ru & 53.0 & 29.5 & 10.2 ($\ell{=}31$)  & -19.3 \\
            & ja & 45.0 & 39.4 & 15.7 ($\ell{=}31$)  & -23.6 \\
        \rowcolor{gray!10}
            & th & 41.0 & 42.9 & 33.2 ($\ell{=}31$)  & -9.6  \\
        \multirow{-8}{*}{Perplexity}
            & bn & 18.0 & 69.1 & 8.6 ($\ell{=}30$)   & -60.5 \\
        \bottomrule
    \end{tabular}%
    }
    \caption{Layer-wise calibration on MGSM with \textbf{LLaMA3} (32 layers). ECE in percent. Last~=~final-layer ECE; Best~=~minimum ECE over all layers (best-layer index in parentheses); $\Delta$ECE~=~Best~$-$~Last.}
    \label{tab:generative_layerwise}
\end{table}

\begin{figure}[t]
    \centering
    \includegraphics[width=\columnwidth]{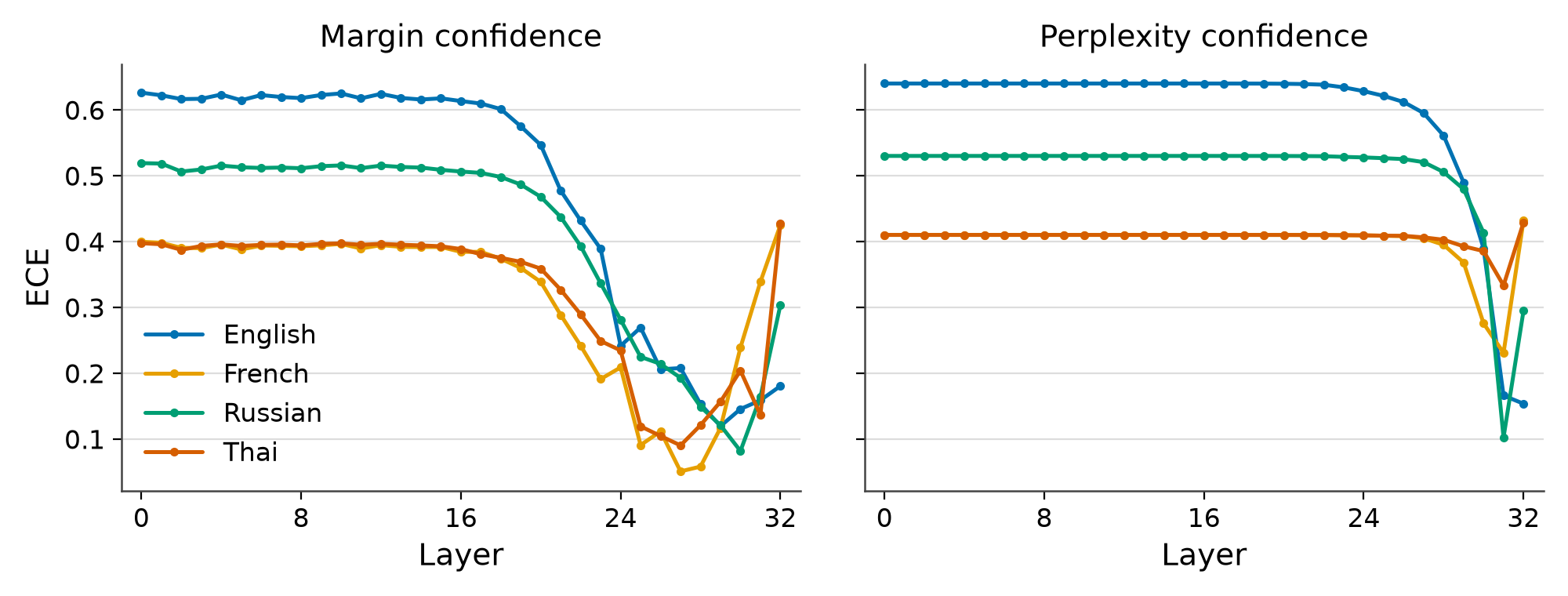}
    \caption{Layer-wise ECE on MGSM (LLaMA3) for en, fr, ru, and th. Left: margin confidence. Right: perplexity confidence. $x$-axis is layer index (0--32); $y$-axis is ECE.}
    \label{fig:layerwise_generative}
\end{figure}

\paragraph{Late-intermediate layers remain better-calibrated.}
We apply the early-decoded confidence-estimation procedure of \S\ref{sec: methodology} to the two probability-space methods that admit a per-layer formulation (margin and perplexity; P(True) and Verbalised require re-prompting and therefore have no natural layer analogue). Table~\ref{tab:generative_layerwise} reports the result for the eight MGSM languages: for every non-English (language $\times$ method) pair, a late-intermediate layer is better calibrated than the final layer, with $\Delta$ECE from $-9.6$ to $-64.4$ percentage points and the largest gains on Bengali. English follows its MCQA pattern (Appendix~\ref{appendix:english_gets_better}): the margin gain is small ($-6.1$) and the final layer is already best for perplexity. For non-English languages the best layers concentrate in $\ell{=}23$--$31$ on a 32-layer model, the same region we identified for MCQA. Figure~\ref{fig:layerwise_generative} shows the per-layer ECE curves.

\begin{table}[t]
    \centering
    \footnotesize
    \setlength{\tabcolsep}{4pt}
    \renewcommand{\arraystretch}{1.1}
    \resizebox{\columnwidth}{!}{%
    \begin{tabular}{l
        S[table-format=2.2]
        S[table-format=2.2]
        S[table-format=2.2]
        S[table-format=2.2]
        S[table-format=2.2]
        S[table-format=2.2]
    }
        \toprule
        \multirow{2}{*}{\textbf{Method}}
        & \multicolumn{3}{c}{\textbf{MGSM}}
        & \multicolumn{3}{c}{\textbf{MKQA}} \\
        \cmidrule(lr){2-4}\cmidrule(lr){5-7}
        & {\textbf{EN}} & {\textbf{Non-EN}} & {\textbf{Avg}}
        & {\textbf{EN}} & {\textbf{Non-EN}} & {\textbf{Avg}} \\
        \midrule
        Final layer$^\dagger$      & 14.75 & 38.41 & 35.45 & 15.52 & 25.72 & 24.44 \\
        \rowcolor{gray!10}
        Global layer$^\dagger$     & 17.57 & 25.13 & 24.18 & 25.02 & 24.03 & 24.15 \\
        Learned selector$^\dagger$ & 20.07 & 19.18 & 19.29 & 22.04 & 23.73 & 23.51 \\
        \rowcolor{gray!10}
        \textsc{LACE}$^\ddagger$   & 18.09 & 19.28 & 19.13 & 18.24 & 21.14 & 20.78 \\
        \bottomrule
    \end{tabular}%
    }
    \caption{Test ECE (\%) on MGSM and MKQA with LLaMA3, averaged over 20 random validation/test splits. Global layer = a single layer tuned on validation data; Learned selector = a logistic regression over per-layer confidences (Appendix~\ref{sec:selector_details}). $^\dagger$:~without language identity; $^\ddagger$:~given language identity.}
    \label{tab:selector}
\end{table}

\paragraph{Layer selection without language identity.}
\textsc{LACE} requires a language-ID signal at inference time. To test whether the layer choice can instead be inferred from the input alone, we train a logistic-regression selector that routes each example to a layer using only its per-layer confidence signature, with no access to language identity (Appendix~\ref{sec:selector_details}). Table~\ref{tab:selector} compares it with the final layer, a single language-agnostic layer tuned on validation data, and \textsc{LACE}, averaged over 20 random validation/test splits. On MGSM the selector matches \textsc{LACE} (average ECE 19.29 vs.\ 19.13) and halves the non-English ECE of the final layer (38.41 $\rightarrow$ 19.18); on MKQA it improves over both language-agnostic baselines, while \textsc{LACE} remains ahead (20.78 vs.\ 23.51).

Both findings also hold under two alternative MCQA scoring protocols (full-option-likelihood and text-generation; Appendix~\ref{sec:protocol_robustness}).

\section{Conclusion}
This paper provides a full-scale evaluation of multilingual calibration in LLMs over 100 languages, uncovering disparities between English and other languages in both accuracy and reliability. Our layer-wise analysis identifies a mid-layer calibration peak: for non-English languages, late-intermediate layers are better-calibrated than the final layer. Leveraging this insight, we introduce \textsc{LACE}, a training-free, layer-aware ensemble that complements post-hoc techniques such as Temperature Scaling. Because these methods reuse intermediate logits without retraining, they can apply directly to deployed models to extract more reliable answering. We further show that the multilingual gap and the mid-layer effect generalise beyond MCQA to generative settings. We hope these results encourage further work on multilingual calibration to support more equitable language technologies.

\section*{Limitations}

Our primary analysis is on MCQA. \S\ref{sec:generative} extends the findings to MKQA and MGSM, but a larger study across more models, languages, and generative confidence-aggregation schemes remains future work. \textsc{LACE} is post-hoc and requires a small per-language validation set (Appendix~\ref{app:validation-size-sensitivity}). A natural next step is to address the gap at training time, e.g.\ via per-language ECE penalties in RLHF/DPO, or via mechanistic interventions \citep{templeton2024scaling} that localise the final-layer English bias. \textsc{LACE} also selects layers at the language level; a per-instance scoring mechanism could further improve robustness for code-mixed inputs. Finally, our evaluation is statistical (ECE, Brier, AUROC); we do not measure whether improved calibration translates into better human decision-making under deployment, which would require a targeted user study.

\section*{Ethics Statement}
Our research adheres to strict ethical guidelines. We verified the licenses of all software and datasets used in this study to ensure full compliance with their terms. No privacy concerns have been identified. We have conducted a thorough assessment of the project and do not anticipate any further risks. We only use LLMs for grammar checking during the paper writing.

\section*{Acknowledgements}
This work was supported by the UKRI Frontier Research Grant EP/Y031350/1 (EQUATE). T.H. was supported by the Gates Cambridge Trust (grant OPP1144 from the Bill \& Melinda Gates Foundation). 

\bibliography{custom}

\clearpage
\newpage
\appendix

\section{Benchmarking Multilingual Calibration}
\label{sec:appendix_eva}
In this section, we present the detailed multilingual evaluation results for the models and benchmarks discussed in the main text.

\subsection{Experimental Details}
\subsubsection{Task Motivation and Confidence Elicitation}

In this study, we focus on MCQA for two key reasons: first, MCQA provides unambiguous correctness labels for calibration. In open-ended generation, correctness is typically approximated using metrics such as ROUGE, BLEU, or LLM-as-a-judge, which often disagree and can even lead to flipped conclusions depending on the choice of metric \citep{ielanskyi2025addressing}. 
Second, MCQA enables a clear definition of confidence as the probability assigned to the selected option. In contrast, confidence estimation for open-ended generation remains unstable and method-dependent (e.g., verbalised confidence, perplexity-based methods, or sampling-based consistency). 
By using MCQA, we can isolate the effect of layer-wise signals on calibration without additional noise from generation-based confidence heuristics.

\paragraph{Correctness Calculation and Confidence Elicitation}
We take the output of the final layer as the model answer when evaluating correctness. 
We do not compute calibration using each layer's own prediction, because our goal is to provide a reliability signal for the deployed model in a standard inference setting, where the user only observes the final-layer answer. 
Therefore, the confidence score should approximate \textit{the probability that this final-layer answer is correct}. 
We follow the standard MCQA practice and use the output probability of the selected answer choice:
$\text{Conf}(x) = \max_{i \in \{1, \dots, C\}} p_i, \quad \text{where } \sum_{i=1}^{C} p_i = 1,$
where $C$ is the number of choices.

\subsubsection{Models}
\label{appendix:models}
Our experiments evaluate eight model families, plus a larger LLaMA-3.1-70B variant for scale analysis:
\begin{itemize}[nosep, leftmargin=*]
    \item \textbf{LLaMA3}~\citep{grattafiori2024llama} (if \textbf{not} specified: \texttt{Llama-3.1-8B\-Instruct})
    \item \textbf{Qwen2.5}~\citep{yang2024qwen2technicalreport} (\texttt{Qwen2.5-7B\-Instruct})
    \item \textbf{Qwen3}~\citep{qwen3-2025} (\texttt{Qwen3-8B})
    \item \textbf{Mistral}~\citep{jiang2023mistral7b} (\texttt{Mistral-7B\-Instruct-v0.3})
    \item \textbf{Aya}~\citep{dang2024ayaexpansecombiningresearch} (\texttt{aya-expanse-8b})
    \item \textbf{DeepSeek}~\citep{deepseekai2025deepseekr1incentivizingreasoningcapability} (\texttt{DeepSeek-R1\-Distill\-Qwen-7B})
    \item \textbf{Phi}~\citep{abdin2024phi4technicalreport} (\texttt{phi-4})
    \item \textbf{Olmo3}~\citep{olmo3-2025} (\texttt{Olmo-3-7B})
    \item \textbf{LLaMA-3.1 70B}~\citep{grattafiori2024llama} (\texttt{Llama-3.1-70B\-Instruct})
\end{itemize}

\subsubsection{Language Group Definitions}
\label{sec:group_def}
We randomly sampled 1,000 examples per language for MMMLU. We group languages in the MMMLU dataset according to resource availability and script type, as summarised in Table~\ref{tab:language_groups}.

\begin{table}[h]
\centering
\footnotesize
\setlength{\tabcolsep}{4pt}
\renewcommand{\arraystretch}{1.15}
\begin{tabular}{@{}p{2.4cm} p{4.8cm}@{}}
\toprule
\textbf{Group} & \textbf{Languages} \\
\midrule
Low-Resource & Arabic, Bengali, Swahili, Yoruba, Hindi, Indonesian \\
\addlinespace[2pt]
High-Resource & German, French, English, Spanish, Chinese, Italian, Japanese, Korean, Portuguese \\
\midrule
Latin-Script & German, English, Spanish, French, Indonesian, Italian, Portuguese, Swahili, Yoruba \\
\addlinespace[2pt]
Non-Latin-Script & Arabic, Bengali, Hindi, Japanese, Korean, Chinese \\
\bottomrule
\end{tabular}
\caption{MMMLU language group definitions by resource availability and script type.}
\label{tab:language_groups}
\end{table}

\paragraph{Language Codes}
The FLORES-200 language codes (3-letter form) for all evaluated languages are provided in the Belebele dataset documentation \citep{Bandarkar_2024}.

\subsection{MMMLU Results}

\begin{table}[h]
    \centering
    \small
    \rowcolors{2}{gray!10}{white}
    \resizebox{\linewidth}{!}{\begin{tabular}{
        l
        | S[table-format=2.2]
        S[table-format=2.2]
        S[table-format=2.2]
        S[table-format=2.2]
    }
        \toprule
        \textbf{Language} & \textbf{AUROC} & \textbf{ECE} & \textbf{BRIER} & \textbf{Accuracy} \\
        \midrule
        Arabic       & 64.91 & 41.18 & 11.87 & 4.50 \\
        Bengali      & 64.56 & 49.70 & 11.72 & 0.10 \\
        German       & 70.84 & 24.14 & 29.32 & 43.00 \\
        English      &  73.75 & 23.92 & 27.95 &  54.00 \\
        Spanish      & 71.33 & 21.64 & 26.79 & 42.90 \\
        French       & 71.25 & 22.20 & 28.36 & 46.40 \\
        Hindi        &  75.08 & 39.77 &  6.23 & 1.60 \\
        Indonesian   & 69.48 & 26.98 & 29.69 & 38.80 \\
        Italian      & 74.08 & 25.24 & 28.25 & 44.50 \\
        Japanese     & 56.09 & 44.15 & 15.48 & 6.50 \\
        Korean       & 39.78 & 46.62 & 16.25 & 5.50 \\
        Portuguese   & 71.11 & 29.25 & 27.59 & 47.10 \\
        Swahili      & 56.02 & 30.81 & 27.34 & 26.30 \\
        Yoruba       & 44.79 & 44.18 & 21.99 & 16.10 \\
        Chinese      & 62.12 & 33.55 & 24.58 & 16.70 \\
        \midrule
        \textit{Avg. Low-Resource}       & 62.47 & 38.77 & 18.14 & 14.57 \\
        \textit{Avg. High-Resource}      & 65.59 & 30.08 & 24.95 & 34.07 \\
        \textit{Avg. Latin-Script}       &  71.69 &  24.77 &  28.28 &  45.24 \\
        \textit{Avg. Non-Latin-Script}   & 57.92 & 41.24 & 16.93 & 9.66 \\
        \midrule
        \textit{\textbf{Average (All Languages)}} & \bfseries 64.35 &\bfseries 33.56 & \bfseries 22.23 & \bfseries 26.27 \\
        \bottomrule
    \end{tabular}}
    \caption{Performance comparison across languages for AUROC, ECE, BRIER score, and Accuracy in \textbf{Mistral}, evaluated on the MMMLU dataset.}
    \label{tab:mistral_mmmlu_multilingual_metrics}
\end{table}

\begin{table}[h]
    \centering
    \small
    \rowcolors{2}{gray!10}{white}
    \resizebox{\linewidth}{!}{\begin{tabular}{
        l
        | S[table-format=2.2]
        S[table-format=2.2]
        S[table-format=2.2]
        S[table-format=2.2]
    }
        \toprule
        \textbf{Language} & \textbf{AUROC} & \textbf{ECE} & \textbf{BRIER} & \textbf{Accuracy} \\
        \midrule
        Arabic       & 67.15 & 14.30 & 26.67 & 54.90 \\
        Bengali      & 64.10 & 26.68 & 31.98 & 33.20 \\
        German       & 76.94 & 21.59 & 25.08 & 55.60 \\
        English      & 78.23 & 15.77 & 19.25 & 65.60 \\
        Spanish      & 76.95 & 19.26 & 23.98 & 61.10 \\
        French       & 75.65 & 16.92 & 22.88 & 62.20 \\
        Hindi        & 72.01 & 28.73 & 28.86 & 33.90 \\
        Indonesian   & 75.69 & 15.83 & 23.53 & 54.30 \\
        Italian      & 75.32 & 21.07 & 24.46 & 58.70 \\
        Japanese     & 80.03 & 6.71  & 17.10 & 33.10 \\
        Korean       & 74.15 & 17.60 & 25.75 & 52.20 \\
        Portuguese   & 75.85 & 18.86 & 23.61 & 58.40 \\
        Swahili      & 59.93 & 30.12 & 33.09 & 32.30 \\
        Yoruba       & 23.49 & 46.99 & 36.11 & 2.00 \\
        Chinese      & 85.31 & 12.47 & 17.42 & 47.00 \\
        \midrule
        \textit{Avg. Low-Resource}       & 60.40 & 27.11 & 30.04 & 35.10 \\
        \textit{Avg. High-Resource}      & 77.60 & 16.69 & 22.17 & 54.88 \\
        \textit{Avg. Latin-Script}       & 76.38 & 18.47 & 23.26 & 59.41 \\
        \textit{Avg. Non-Latin-Script}   & 65.77 & 22.95 & 27.12 & 36.08 \\
        \midrule
        \textit{\textbf{Average (All Languages)}} & \textbf{70.72} & \textbf{20.86} & \textbf{25.32} & \textbf{46.97} \\
        \bottomrule
    \end{tabular}
    }
    \caption{Performance comparison across languages for AUROC, ECE, BRIER score, and Accuracy in \textbf{Qwen 2.5}, evaluated on the MMMLU dataset.}
    \label{tab:qwen_mmmlu_multilingual_metrics}
\end{table}

\begin{table}[h]
    \centering
    \small
    \rowcolors{2}{gray!10}{white}
    \resizebox{\linewidth}{!}{\begin{tabular}{
        l
        | S[table-format=2.2]
        S[table-format=2.2]
        S[table-format=2.2]
        S[table-format=2.2]
    }
        \toprule
        \textbf{Language} & \textbf{AUROC} & \textbf{ECE} & \textbf{BRIER} & \textbf{Accuracy} \\
        \midrule
        Arabic       & 52.66 & 30.21 & 25.35 & 36.50 \\
        Bengali      & 52.62 & 34.13 & 24.73 & 27.20 \\
        German       & 63.47 & 22.86 & 22.86 & 65.60 \\
        English      & 71.13 & 20.48 & 17.92 & 73.10 \\
        Spanish      & 61.29 & 27.15 & 25.32 & 56.40 \\
        French       & 71.57 & 17.07 & 20.21 & 68.90 \\
        Hindi        & 37.74 & 46.43 & 26.16 & 15.70 \\
        Indonesian   & 42.89 & 32.36 & 30.63 & 30.70 \\
        Italian      & 72.25 & 10.51 & 19.13 & 67.50 \\
        Japanese     & 30.62 & 46.69 & 17.59 & 8.30 \\
        Korean       & 66.95 & 29.00 & 24.50 & 50.00 \\
        Portuguese   & 73.79 & 13.24 & 18.77 & 66.60 \\
        Swahili      & 64.42 & 16.18 & 23.61 & 40.50 \\
        Yoruba       & 53.76 & 20.83 & 21.01 & 27.60 \\
        Chinese      & 59.73 & 31.98 & 26.17 & 44.60 \\
        \midrule
        \textit{Avg. Low-Resource}       & 50.68 & 30.02 & 25.25 & 29.70 \\
        \textit{Avg. High-Resource}      & 63.42 & 24.33 & 21.39 & 55.67 \\
        \textit{Avg. Latin-Script}       & 65.20 & 20.52 & 22.12 & 61.26 \\
        \textit{Avg. Non-Latin-Script}   & 52.31 & 31.93 & 23.64 & 31.30 \\
        \midrule
        \textit{\textbf{Average (All Languages)}} & \textbf{58.33} & \textbf{26.61} & \textbf{22.93} & \textbf{45.28} \\
        \bottomrule
    \end{tabular}
    }
    \caption{Performance comparison across languages for AUROC, ECE, BRIER score, and Accuracy in \textbf{Phi}, evaluated on the MMMLU dataset.}
    \label{tab:phi_mmmlu_multilingual_metrics}
\end{table}

\begin{table}[h]
    \centering
    \small
    \rowcolors{2}{gray!10}{white}
    \resizebox{\linewidth}{!}{\begin{tabular}{
        l
        | S[table-format=2.2]
        S[table-format=2.2]
        S[table-format=2.2]
        S[table-format=2.2]
    }
        \toprule
        \textbf{Language} & \textbf{AUROC} & \textbf{ECE} & \textbf{BRIER} & \textbf{Accuracy} \\
        \midrule
        Arabic       & 55.33 & 32.74 & 21.54 & 26.40 \\
        Bengali      & 58.50 & 40.80 & 14.41 & 13.70 \\
        German       & 60.28 & 18.50 & 23.91 & 39.80 \\
        English      & 66.21 & 9.10  & 22.92 & 47.10 \\
        Spanish      & 62.24 & 12.47 & 23.51 & 40.80 \\
        French       & 62.93 & 10.84 & 23.12 & 41.40 \\
        Hindi        & 56.08 & 30.42 & 20.62 & 26.40 \\
        Indonesian   & 61.00 & 31.61 & 21.11 & 27.30 \\
        Italian      & 63.14 & 5.65  & 22.85 & 40.40 \\
        Japanese     & 55.56 & 18.05 & 23.14 & 32.10 \\
        Korean       & 21.56 & 49.09 & 18.66 & 1.10 \\
        Portuguese   & 62.37 & 16.78 & 23.26 & 39.10 \\
        Swahili      & 51.67 & 45.76 & 12.45 & 12.00 \\
        Yoruba       & 60.35 & 38.16 & 4.94  & 2.80 \\
        Chinese      & 69.00 & 16.13 & 23.97 & 43.10 \\
        \midrule
        \textit{Avg. Low-Resource}       & 57.16 & 36.58 & 15.84 & 18.10 \\
        \textit{Avg. High-Resource}      & 58.14 & 17.40 & 22.82 & 36.10 \\
        \textit{Avg. Latin-Script}       & 62.60 & 14.99 & 22.95 & 39.41 \\
        \textit{Avg. Non-Latin-Script}   & 53.51 & 33.89 & 17.47 & 19.70 \\
        \midrule
        \textit{\textbf{Average (All Languages)}} & \textbf{57.75} & \textbf{25.07} & \textbf{20.03} & \textbf{28.90} \\
        \bottomrule
    \end{tabular}}
    \caption{Performance comparison across languages for AUROC, ECE, BRIER score, and Accuracy in \textbf{DeepSeek}, evaluated on the MMMLU dataset.}
    \label{tab:deepseekq_mmmlu_multilingual_metrics}
\end{table}

Results for each model are reported in the following tables: Mistral (Table~\ref{tab:mistral_mmmlu_multilingual_metrics}), Qwen2.5 (Table~\ref{tab:qwen_mmmlu_multilingual_metrics}), Phi (Table~\ref{tab:phi_mmmlu_multilingual_metrics}), and DeepSeek (Table~\ref{tab:deepseekq_mmmlu_multilingual_metrics}).

\subsection{Model Confidence Behaviours}
\label{sec:behaviour}
\begin{figure}[t]
    \centering
    \begin{subfigure}[b]{0.49\linewidth}
        \centering
        \includegraphics[width=\linewidth]{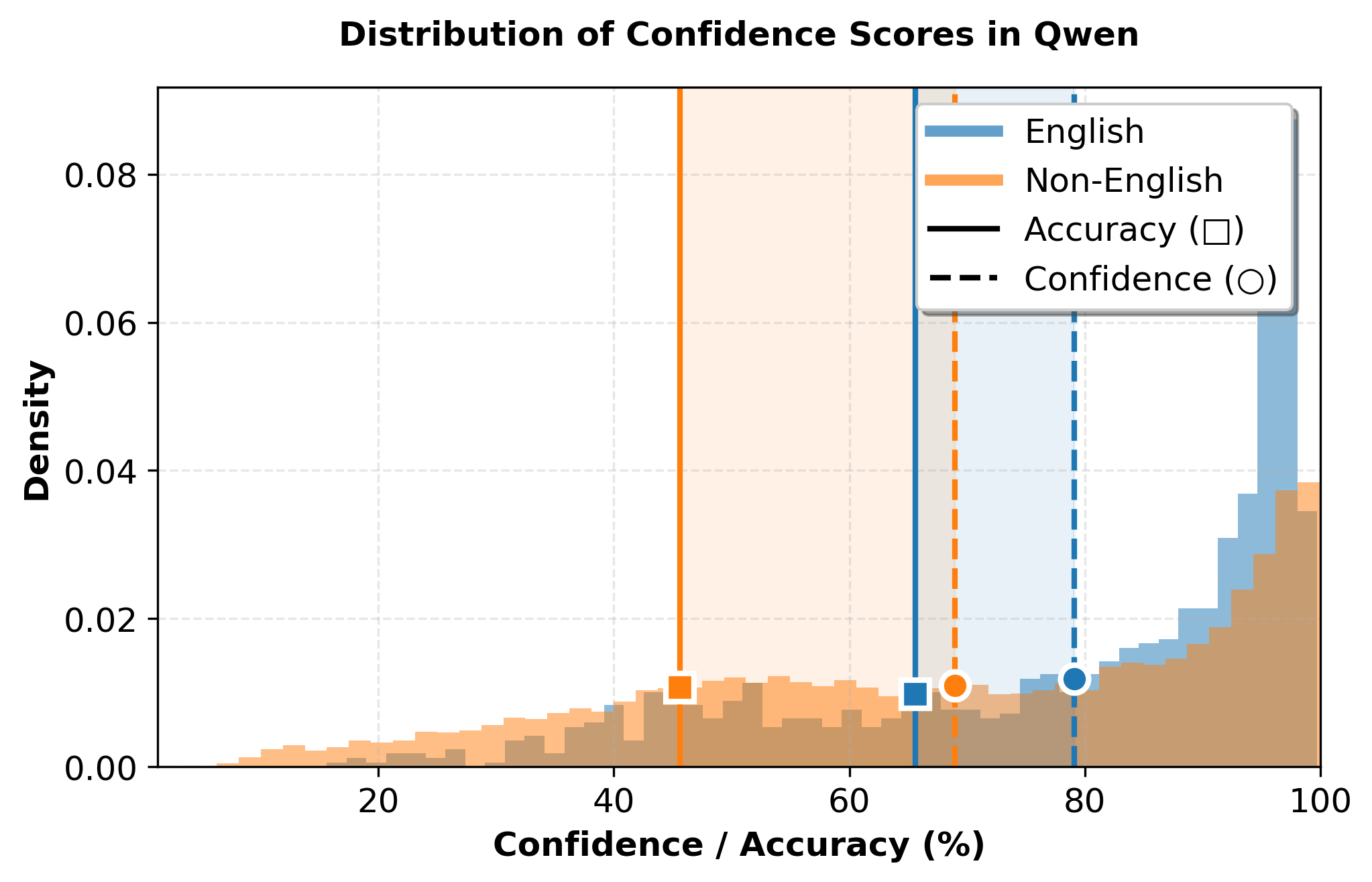}
        \caption{Qwen 2.5}
        \label{fig:qwen}
    \end{subfigure}
    \hfill
    \begin{subfigure}[b]{0.49\linewidth}
        \centering
        \includegraphics[width=\linewidth]{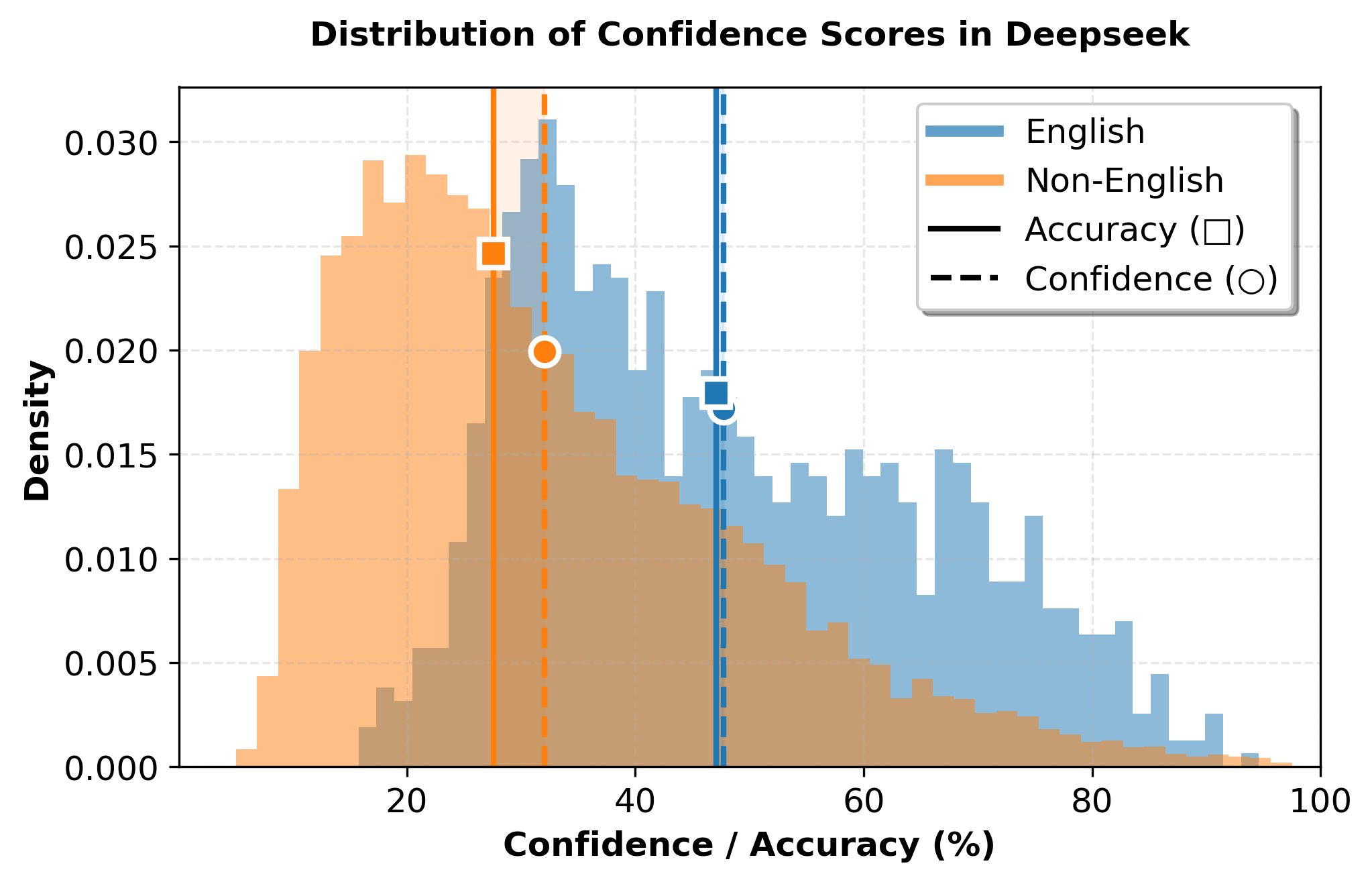}
        \caption{DeepSeek}
        \label{fig:deepseek}
    \end{subfigure}
    
    \medskip
    
    \begin{subfigure}[b]{0.49\linewidth}
        \centering
        \includegraphics[width=\linewidth]{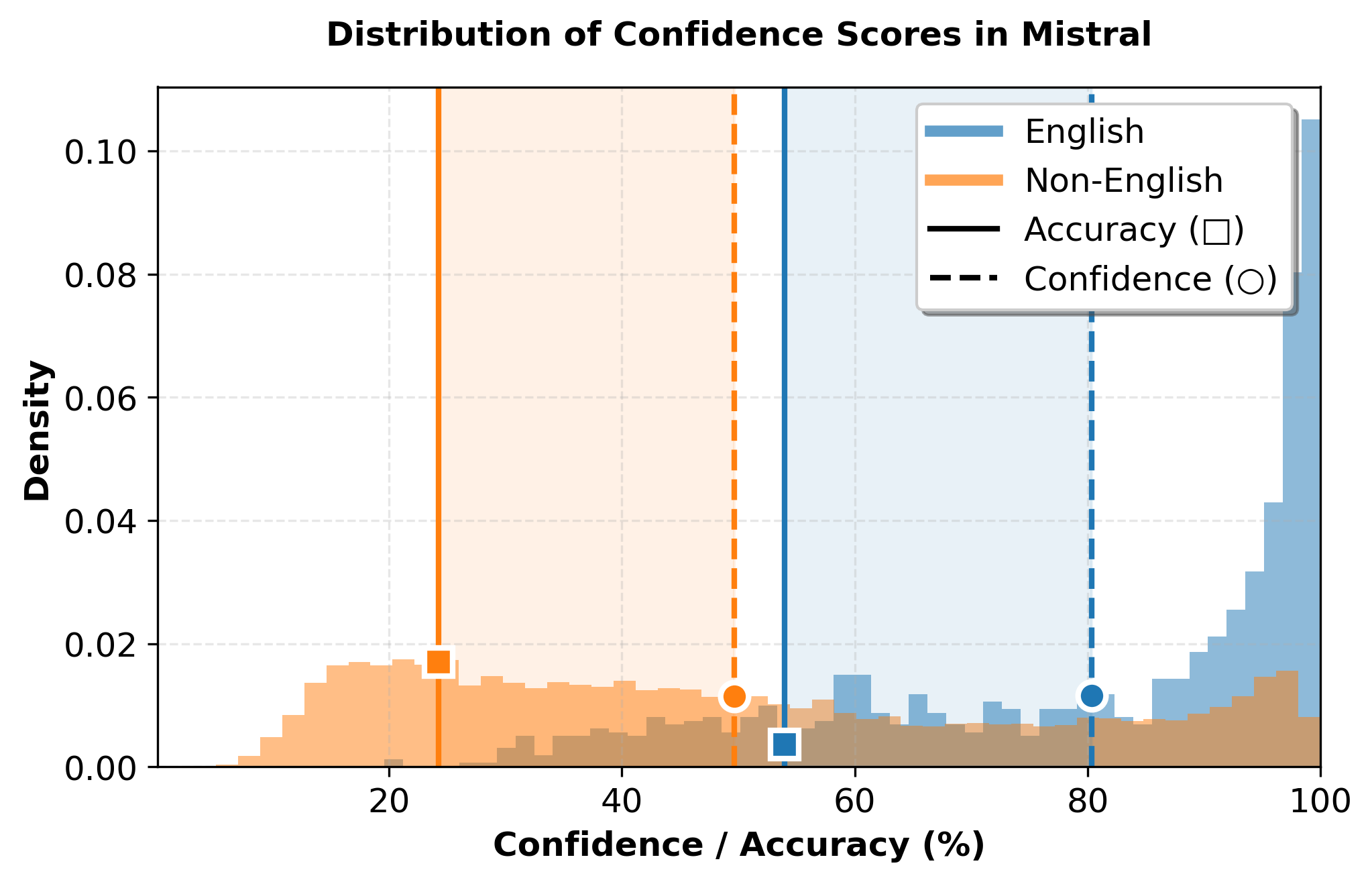}
        \caption{Mistral}
        \label{fig:mistral}
    \end{subfigure}
    \hfill
    \begin{subfigure}[b]{0.49\linewidth}
        \centering
        \includegraphics[width=\linewidth]{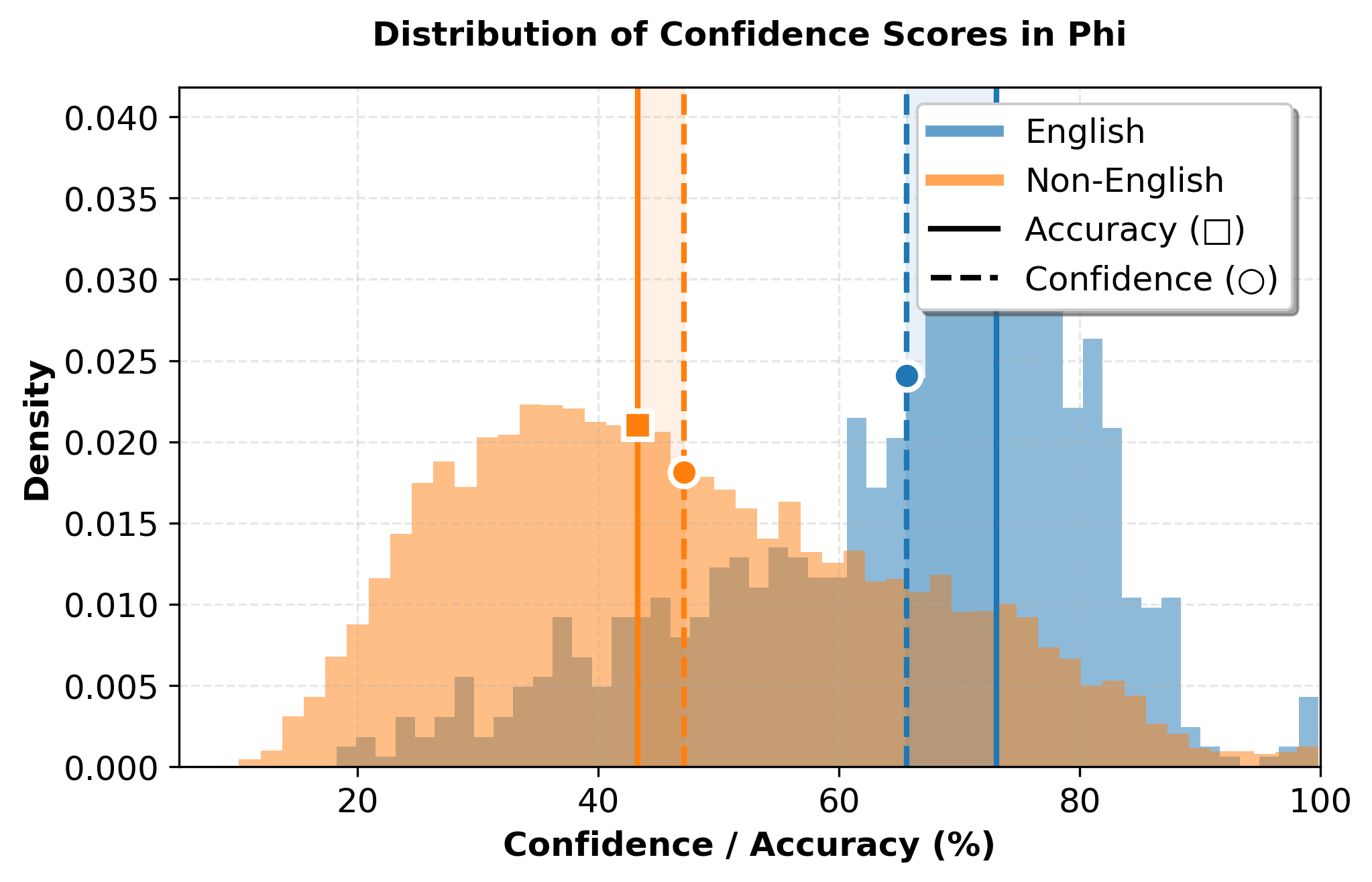}
        \caption{Phi}
        \label{fig:phi}
    \end{subfigure}

    \caption{Distribution of confidence scores (predicted probabilities) versus accuracies in English (blue) and non-English (orange) for four models: (a) Qwen 2.5, (b) DeepSeek, (c) Mistral, and (d) Phi. 
    }
    \label{fig:confidence_grid}
\end{figure}

Figure~\ref{fig:confidence_grid} illustrates the distribution of confidence scores and accuracies across English and non-English settings for Qwen2.5, DeepSeek, Mistral, and Phi. Solid vertical lines indicate mean accuracies, while dashed vertical lines indicate mean confidences. The divergence between confidence and accuracy highlights calibration behaviour: underconfidence when the dashed line falls left of the solid line, and overconfidence when it falls to the right.

Table~\ref{tab:llama_language_confidence_performance_123} reports detailed calibration and confidence statistics for LLaMA3. In addition to standard accuracy, we provide the model’s average confidence, the confidence gap (accuracy minus confidence), the proportion of correct predictions made with low confidence (``Underconf''), and the mean confidence levels assigned to correct vs.~incorrect predictions. We further include the difference between these two distributions (``Corr–Inc Gap''), which captures how well the model separates correct from incorrect responses. English shows a relatively small confidence gap (2.5\%), with strong separation between correct and incorrect predictions (23.8\% Corr–Inc Gap). In contrast, most non-English languages show lower accuracy, larger underconfidence rates, and much smaller separation between correct and incorrect predictions (average Corr–Inc Gap of 6.3\%).

\begin{table}[t]

\centering
\resizebox{\linewidth}{!}{
\rowcolors{2}{gray!10}{white}
\begin{tabular}{lccccccc}
\toprule
Language     & Acc. & Avg Conf & Conf Gap & Underconf & Corr. Conf & Inc. Conf & Corr–Inc Gap \\
             & (\%)     & (\%)     & (\%)     & (\%)      & (\%)         & (\%)           & (\%) \\
\midrule
Arabic       & 38.2 & 29.5 &  8.8 & 89.0 & 31.7 & 28.0 &  3.7 \\
Chinese      & 23.1 & 23.2 & -0.1 & 99.6 & 22.2 & 23.6 & -1.4 \\
Korean       & 42.5 & 31.2 & 11.3 & 92.0 & 33.5 & 29.5 &  4.0 \\
Japanese     & 43.0 & 24.0 & 19.0 & 92.3 & 26.8 & 21.8 &  5.0 \\
Swahili      & 32.2 & 31.5 &  0.7 & 88.8 & 34.2 & 30.3 &  3.9 \\
Italian      & 51.8 & 41.8 & 10.0 & 58.3 & 47.4 & 35.8 & 11.6 \\
Bengali      & 35.2 & 34.6 &  0.6 & 88.4 & 36.1 & 33.8 &  2.3 \\
Spanish      & 52.0 & 46.1 &  5.9 & 50.0 & 52.6 & 39.0 & 13.6 \\
Portuguese   & 50.4 & 47.1 &  3.3 & 47.0 & 53.6 & 40.6 & 13.0 \\
Indonesian   & 45.0 & 37.0 &  8.0 & 75.3 & 41.1 & 33.7 &  7.4 \\
Hindi        & 39.9 & 31.7 &  8.2 & 88.2 & 34.1 & 30.1 &  4.0 \\
German       & 44.4 & 33.8 & 10.6 & 80.4 & 37.1 & 31.2 &  5.9 \\
Yoruba       & 27.4 & 29.4 & -2.0 & 93.8 & 31.5 & 28.7 &  2.8 \\
French       & 51.3 & 41.5 &  9.8 & 59.5 & 47.4 & 35.3 & 12.1 \\
\midrule
\textbf{English}      & 61.2 & 58.8 &  2.5 & \textbf{25.7} & 68.0 & 44.2 & \textbf{23.8} \\
\midrule
\textbf{Non-English } & 41.2 & 34.5 &  6.7 & \textbf{78.8} & 37.8 & 31.5 & \textbf{ 6.3} \\
\bottomrule
\end{tabular}
}
\caption{LLaMA3 Calibration and underconfidence analysis across languages. Metrics include accuracy, average confidence, confidence gap (accuracy minus confidence), proportion of underconfident correct predictions (confidence $<$ 0.5 when correct), average confidence for correct vs.~incorrect predictions, and their difference (Corr–Inc Gap).}
\label{tab:llama_language_confidence_performance_123}
\end{table}

\subsection{Belebele Results}
\label{sec:belebele_results}
Belebele \citep{Bandarkar_2024} is a multiple-choice dataset covering 122 language variants, enabling robust evaluation of NLU across high-, medium-, and low-resource languages. The dataset is fully parallel, allowing for direct cross-linguistic comparison of model performance. In our experiments, we sample 400 examples per language. Per-language calibration metrics on Belebele are reported in Table~\ref{tab:per_language_results} for LLaMA3.

\definecolor{highlatin}{RGB}{235,245,255}     %
\definecolor{highnonlatin}{RGB}{255,235,240}  %
\definecolor{mediumlatin}{RGB}{225,255,225}   %
\definecolor{mediumnonlatin}{RGB}{245,230,250}%
\definecolor{lowlatin}{RGB}{255,250,230}      %
\definecolor{lownonlatin}{RGB}{245,235,220}   %

\begin{table*}[t]
    \centering
    \footnotesize
    \renewcommand{\arraystretch}{1.15}
\setlength{\tabcolsep}{8pt}
\rowcolors{2}{gray!10}{white} 
    \resizebox{\textwidth}{!}{
\begin{tabular}{ccccc|ccccc|ccccc}
\toprule
\multicolumn{5}{c|}{\textbf{Set 1}} & \multicolumn{5}{c|}{\textbf{Set 2}} & \multicolumn{5}{c}{\textbf{Set 3}} \\
\cmidrule(lr){1-5} \cmidrule(lr){6-10} \cmidrule(lr){11-15}
    \textbf{Lang} & \textbf{Acc} & \textbf{AUR.} & \textbf{ECE} & \textbf{Brier} &
    \textbf{Lang} & \textbf{Acc} & \textbf{AUR.} & \textbf{ECE} & \textbf{Brier} &
    \textbf{Lang} & \textbf{Acc} & \textbf{AUR.} & \textbf{ECE} & \textbf{Brier} \\
\midrule
\cellcolor{mediumnonlatin} acm & 58.2 & 78.4 & 7.1 & 19.2 & \cellcolor{mediumnonlatin} arz & 69.8 & 77.8 & 12.2 & 18.2 & \cellcolor{lowlatin} ceb & 65.8 & 75.6 & 11.1 & 19.7 \\
\cellcolor{highlatin} fin & 80.8 & 75.6 & 6.0 & 14.1 & \cellcolor{mediumnonlatin} hin & 67.0 & 72.9 & 14.0 & 20.6 & \cellcolor{highlatin} ita & 86.0 & 79.6 & 7.4 & 10.8 \\
\cellcolor{lownonlatin} khm & 4.0 & 5.5 & 66.2 & 51.4 & \cellcolor{lowlatin} lvs & 74.8 & 79.7 & 11.3 & 16.2 & \cellcolor{lownonlatin} npi & 59.8 & 74.8 & 5.1 & 20.0 \\
\cellcolor{highlatin} pol & 80.0 & 78.9 & 6.3 & 13.3 & \cellcolor{mediumlatin} slv & 81.0 & 76.6 & 8.7 & 13.8 & \cellcolor{highlatin} swe & 79.2 & 81.6 & 7.3 & 12.8 \\
\cellcolor{lowlatin} tso & 34.0 & 62.5 & 3.4 & 21.5 & \cellcolor{lowlatin} xho & 37.0 & 60.0 & 5.0 & 22.8 & \cellcolor{mediumlatin} afr & 80.8 & 81.2 & 8.3 & 12.7 \\
\cellcolor{mediumnonlatin} asm & 45.8 & 68.0 & 6.1 & 22.8 & \cellcolor{highlatin} ces & 82.2 & 80.2 & 7.8 & 11.8 & \cellcolor{highlatin} fra & 86.2 & 77.4 & 9.9 & 11.4 \\
\cellcolor{mediumlatin} hin & 57.0 & 72.1 & 4.7 & 21.1 & \cellcolor{mediumlatin} jav & 68.0 & 78.4 & 11.1 & 18.4 & \cellcolor{mediumlatin} kin & 34.8 & 63.3 & 4.7 & 21.3 \\
\cellcolor{mediumnonlatin} mal & 60.5 & 74.0 & 12.3 & 21.3 & \cellcolor{lowlatin} npi & 32.5 & 59.9 & 6.3 & 21.4 & \cellcolor{highlatin} por & 86.2 & 79.8 & 5.6 & 10.2 \\
\cellcolor{mediumlatin} sna & 35.8 & 58.6 & 6.2 & 23.0 & \cellcolor{mediumlatin} swh & 67.0 & 75.0 & 8.1 & 19.1 & \cellcolor{mediumlatin} tur & 78.2 & 78.8 & 8.5 & 14.6 \\
\cellcolor{mediumlatin} yor & 31.2 & 61.1 & 4.4 & 20.4 & \cellcolor{lowlatin} als & 73.5 & 78.0 & 7.2 & 16.2 & \cellcolor{lowlatin} azj & 66.8 & 71.8 & 14.3 & 21.4 \\
\cellcolor{mediumnonlatin} ckb & 46.0 & 71.8 & 8.3 & 22.2 & \cellcolor{lowlatin} fuv & 28.0 & 51.9 & 5.2 & 20.4 & \cellcolor{mediumlatin} hrv & 79.8 & 78.1 & 8.2 & 14.3 \\
\cellcolor{highnonlatin} jpn & 66.5 & 73.3 & 29.9 & 28.0 & \cellcolor{lownonlatin} kir & 63.0 & 73.8 & 21.9 & 24.7 & \cellcolor{mediumnonlatin} mar & 67.5 & 73.5 & 9.2 & 19.4 \\
\cellcolor{lowlatin} nso & 37.8 & 60.3 & 3.2 & 22.7 & \cellcolor{lownonlatin} snd & 17.5 & 50.5 & 30.0 & 25.3 & \cellcolor{mediumnonlatin} tam & 65.5 & 73.4 & 14.5 & 21.5 \\
\cellcolor{highnonlatin} ukr & 84.2 & 77.4 & 12.1 & 12.8 & \cellcolor{highnonlatin} zho & 76.5 & 71.2 & 30.6 & 25.2 &
\cellcolor{mediumnonlatin} amh & 34.8 & 63.2 & 5.0 & 21.5 \\
\cellcolor{lowlatin} bam & 31.2 & 60.6 & 4.8 & 20.8 & \cellcolor{highlatin} dan & 79.8 & 78.8 & 8.2 & 13.9 & \cellcolor{lowlatin} gaz & 31.8 & 53.1 & 2.4 & 21.7 \\
\cellcolor{highlatin} hun & 82.5 & 84.0 & 9.7 & 11.8 & \cellcolor{lowlatin} kac & 30.2 & 61.8 & 3.5 & 20.5 & \cellcolor{highnonlatin} kor & 77.8 & 77.9 & 14.5 & 16.3 \\
\cellcolor{lownonlatin} mkd & 77.8 & 79.1 & 7.8 & 14.4 & \cellcolor{lowlatin} nya & 32.0 & 60.7 & 3.3 & 21.1 & \cellcolor{mediumlatin} ron & 80.0 & 80.4 & 6.3 & 13.0 \\
\cellcolor{lowlatin} som & 35.2 & 59.0 & 3.9 & 22.2 & \cellcolor{mediumnonlatin} tel & 59.5 & 73.7 & 7.2 & 20.4 & \cellcolor{mediumnonlatin} urd & 59.5 & 67.3 & 19.7 & 25.8 \\
\cellcolor{highnonlatin} zho & 81.2 & 68.9 & 24.4 & 21.1 & \cellcolor{mediumnonlatin} apc & 65.0 & 78.3 & 9.7 & 18.4 & \cellcolor{mediumnonlatin} ben & 65.5 & 72.6 & 10.1 & 20.3 \\
\cellcolor{highlatin} deu & 86.8 & 72.3 & 13.4 & 12.9 & \cellcolor{lowlatin} grn & 39.8 & 65.5 & 6.6 & 22.4 & \cellcolor{lownonlatin} hye & 0.2 & 1.5 & 63.8 & 42.1 \\
\cellcolor{mediumnonlatin} kan & 58.5 & 72.1 & 5.2 & 20.9 & \cellcolor{mediumnonlatin} lao & 32.5 & 59.5 & 2.1 & 21.5 & \cellcolor{lowlatin} mlt & 69.8 & 76.4 & 9.9 & 18.4 \\
\cellcolor{mediumnonlatin} ory & 55.8 & 70.0 & 16.4 & 24.7 & \cellcolor{highnonlatin} rus & 81.2 & 81.0 & 10.4 & 13.0 & \cellcolor{lowlatin} sot & 32.8 & 57.5 & 2.6 & 21.8 \\
\cellcolor{lownonlatin} tgk & 63.8 & 70.1 & 12.5 & 22.1 & \cellcolor{mediumlatin} urd & 41.2 & 64.8 & 4.3 & 22.5 & \cellcolor{mediumlatin} zsm & 82.5 & 82.9 & 9.5 & 11.8 \\
\cellcolor{highnonlatin} arb & 79.5 & 74.2 & 11.8 & 15.5 & \cellcolor{lowlatin} ben & 35.2 & 59.9 & 4.9 & 22.4 & \cellcolor{highnonlatin} ell & 80.5 & 80.6 & 10.1 & 13.7 \\
\cellcolor{mediumnonlatin} guj & 58.0 & 68.8 & 8.2 & 22.0 & \cellcolor{mediumlatin} ibo & 40.2 & 62.0 & 3.7 & 22.6 & \cellcolor{lownonlatin} kat & 1.5 & 6.0 & 68.4 & 51.6 \\
\cellcolor{mediumlatin} lin & 34.2 & 59.3 & 3.5 & 21.9 & \cellcolor{mediumlatin} mri & 35.5 & 63.3 & 4.3 & 21.5 & \cellcolor{mediumnonlatin} pan & 58.0 & 74.4 & 11.4 & 21.6 \\
\cellcolor{lownonlatin} shn & 16.8 & 47.9 & 15.1 & 16.9 & \cellcolor{highlatin} spa & 84.0 & 79.3 & 5.8 & 11.7 & \cellcolor{mediumlatin} tgl & 75.2 & 80.1 & 7.8 & 15.3 \\
\cellcolor{lowlatin} uzn & 69.0 & 77.0 & 13.4 & 19.2 & \cellcolor{lowlatin} zul & 36.5 & 59.3 & 5.1 & 22.8 & \cellcolor{lowlatin} arb & 29.8 & 56.4 & 3.6 & 20.8 \\
\cellcolor{mediumnonlatin} bod & 29.0 & 60.2 & 6.8 & 20.3 & \cellcolor{lowlatin} sun & 65.5 & 74.3 & 13.5 & 20.9 & \cellcolor{lowlatin} hat & 55.8 & 72.9 & 4.6 & 20.9 \\
\cellcolor{mediumlatin} ilo & 54.0 & 69.8 & 9.2 & 22.6 & \cellcolor{mediumnonlatin} kaz & 63.5 & 75.6 & 19.3 & 23.0 & \cellcolor{highlatin} lit & 73.8 & 82.3 & 11.0 & 15.7 \\
\cellcolor{lownonlatin} mya & 0.8 & 7.2 & 72.5 & 55.7 & \cellcolor{mediumnonlatin} pbt & 47.5 & 67.7 & 3.1 & 22.6 & \cellcolor{lowlatin} sin & 32.2 & 59.2 & 3.8 & 21.5 \\
\cellcolor{mediumnonlatin} srp & 83.2 & 77.5 & 13.0 & 13.7 & \cellcolor{mediumnonlatin} tha & 71.8 & 75.5 & 19.8 & 21.2 & \cellcolor{mediumlatin} vie & 83.5 & 78.4 & 8.7 & 12.4 \\
\cellcolor{mediumnonlatin} ars & 62.0 & 78.5 & 9.6 & 18.8 & \cellcolor{mediumnonlatin} bul & 80.8 & 77.7 & 13.6 & 14.6 & \cellcolor{mediumlatin} est & 71.8 & 78.2 & 4.2 & 16.4 \\
\cellcolor{mediumlatin} hau & 45.2 & 67.1 & 11.7 & 23.9 & \cellcolor{mediumlatin} ind & 81.8 & 75.6 & 6.4 & 13.4 & \cellcolor{lowlatin} kea & 48.8 & 73.0 & 8.0 & 21.1 \\
\cellcolor{mediumlatin} lug & 35.5 & 57.6 & 2.5 & 22.5 & \cellcolor{highlatin} nld & 83.0 & 78.2 & 5.9 & 12.0 & \cellcolor{mediumnonlatin} pes & 79.2 & 77.8 & 8.5 & 14.4 \\
\cellcolor{mediumnonlatin} sin & 58.8 & 72.7 & 10.3 & 21.6 & \cellcolor{lowlatin} ssw & 31.8 & 61.5 & 3.0 & 20.8 & \cellcolor{mediumnonlatin} tir & 28.0 & 57.7 & 4.9 & 19.7 \\
\cellcolor{mediumlatin} war & 62.2 & 74.7 & 12.1 & 21.0 & \cellcolor{mediumnonlatin} ary & 58.5 & 72.0 & 7.7 & 21.2 & \cellcolor{highlatin} cat & 86.2 & 84.8 & 9.4 & 10.2 \\
\cellcolor{lowlatin} eus & 69.5 & 75.7 & 7.0 & 18.1 & \cellcolor{highnonlatin} heb & 77.2 & 76.9 & 17.2 & 17.8 & \cellcolor{highlatin} isl & 67.0 & 78.3 & 5.2 & 17.4 \\
\cellcolor{lownonlatin} khk & 48.0 & 70.2 & 10.0 & 22.9 & \cellcolor{lowlatin} luo & 31.8 & 55.7 & 5.7 & 21.5 & \cellcolor{highlatin} nob & 79.2 & 77.8 & 9.5 & 14.7 \\
\cellcolor{lowlatin} plt & 44.2 & 65.6 & 5.8 & 23.1 & \cellcolor{highlatin} slk & 82.0 & 76.9 & 4.7 & 13.0 &
\cellcolor{highlatin} \textbf{eng} & \textbf{87.8} &\textbf{ 87.9} & \textbf{4.0} &\textbf{ 8.1}   \\
\cellcolor{lowlatin} tsn & 31.8 & 62.7 & 4.6 & 20.8 & \cellcolor{lowlatin} wol & 33.0 & 53.2 & 5.7 & 22.3 & \textbf{Avg.} & \textbf{57.6} & \textbf{68.9 }& \textbf{10.9} &\textbf{ 19.8} \\
\bottomrule
    \end{tabular}}
    \caption{Per-language performance on the belebele test set for the \textbf{LLaMA3} model, reporting AUROC, ECE, and Brier score. Each row is color-coded by language category, based on resource availability (high, medium, low) and script type (Latin vs. non-Latin). The categories are shaded with soft pastel colors: high-resource Latin (light blue), high-resource non-Latin (light pink), medium-resource Latin (light green), medium-resource non-Latin (lavender), low-resource Latin (cream), and low-resource non-Latin (tan). English line and the Average line is \textbf{bolded}.}
    \label{tab:per_language_results}
\end{table*}

\section{Layer-Wise Calibration Analysis}
\label{sec:appendix}

\subsection{English Calibration Improves as Layer Deepens}

\label{appendix:english_gets_better}

\begin{figure}[h]
    \centering
    \includegraphics[width=\linewidth]{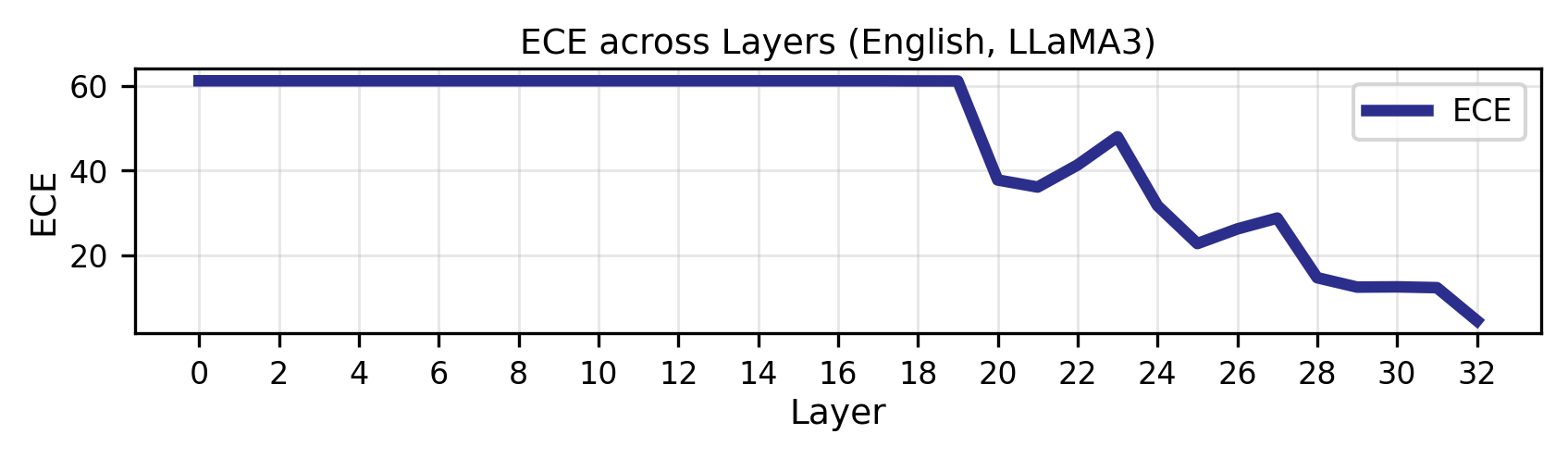}
    \caption{Layer-wise predicted confidence ECE for English in LLaMA3.}
    \label{fig:english_llama3_calibration_vs_entropy_ece}
\end{figure}

Figure~\ref{fig:english_llama3_calibration_vs_entropy_ece} shows the layer-wise Expected Calibration Error (ECE) for English in LLaMA3, illustrating how calibration improves progressively in deeper layers.

\subsection{Multilingual Calibration is Best at Late-Intermediate Layers}
\label{sec:multicalib}
\begin{figure*}[t]
    \centering
    \subcaptionbox{LLaMA3\label{fig:mmmlu_llama3_calibration_vs_entropy}}{%
        \includegraphics[width=0.32\linewidth]{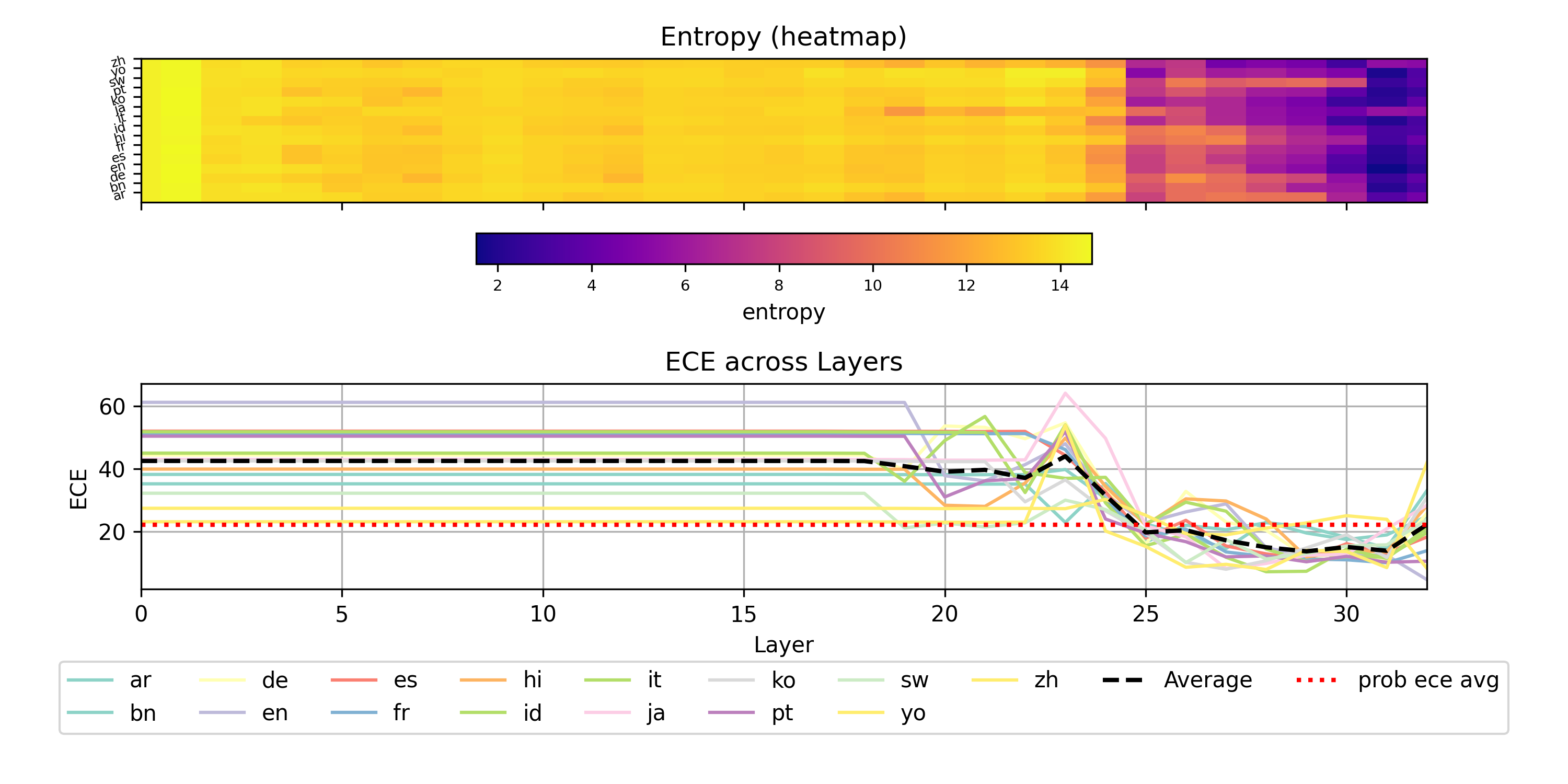}\hfill
        \includegraphics[width=0.32\linewidth]{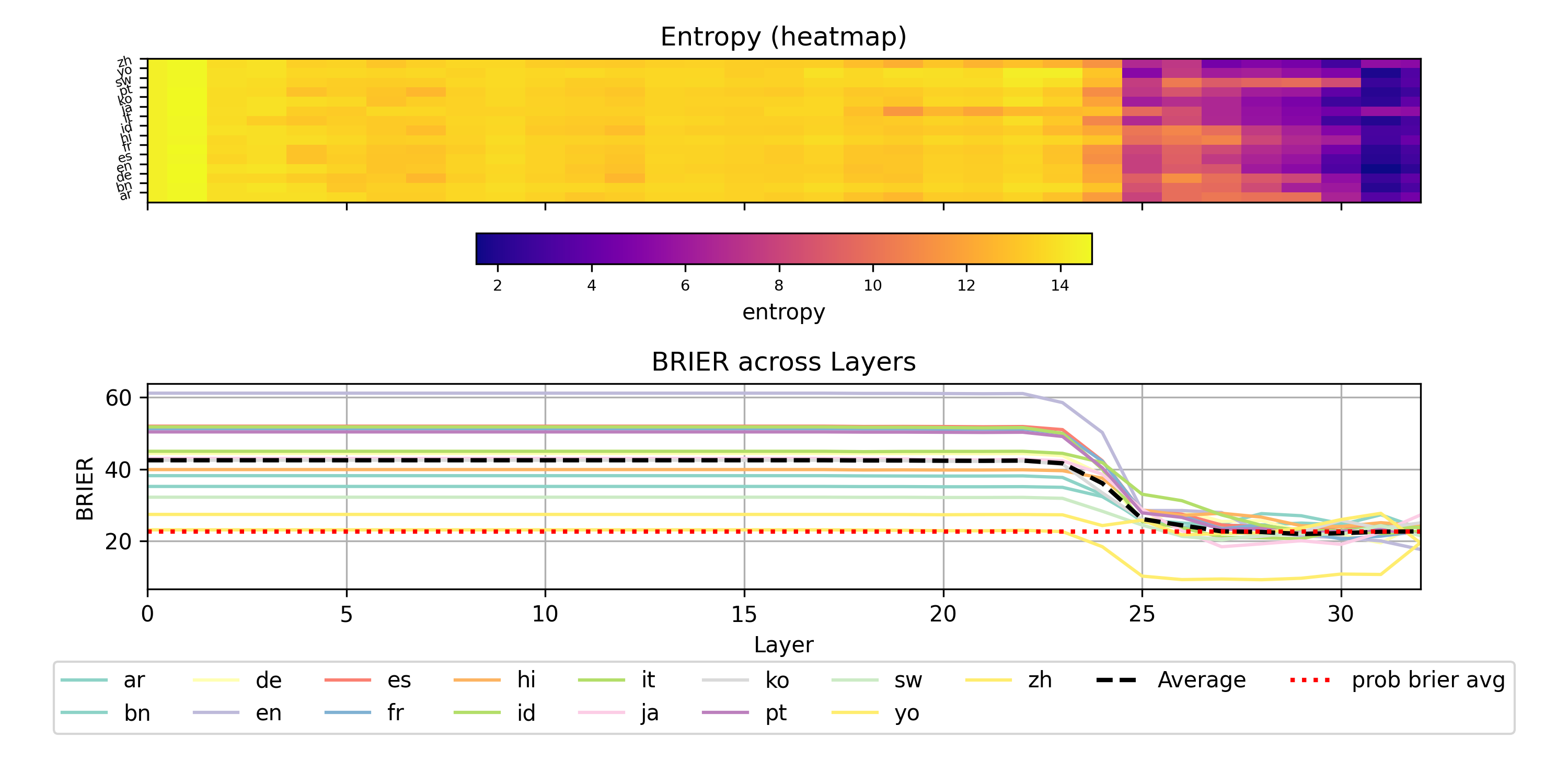}\hfill
        \includegraphics[width=0.32\linewidth]{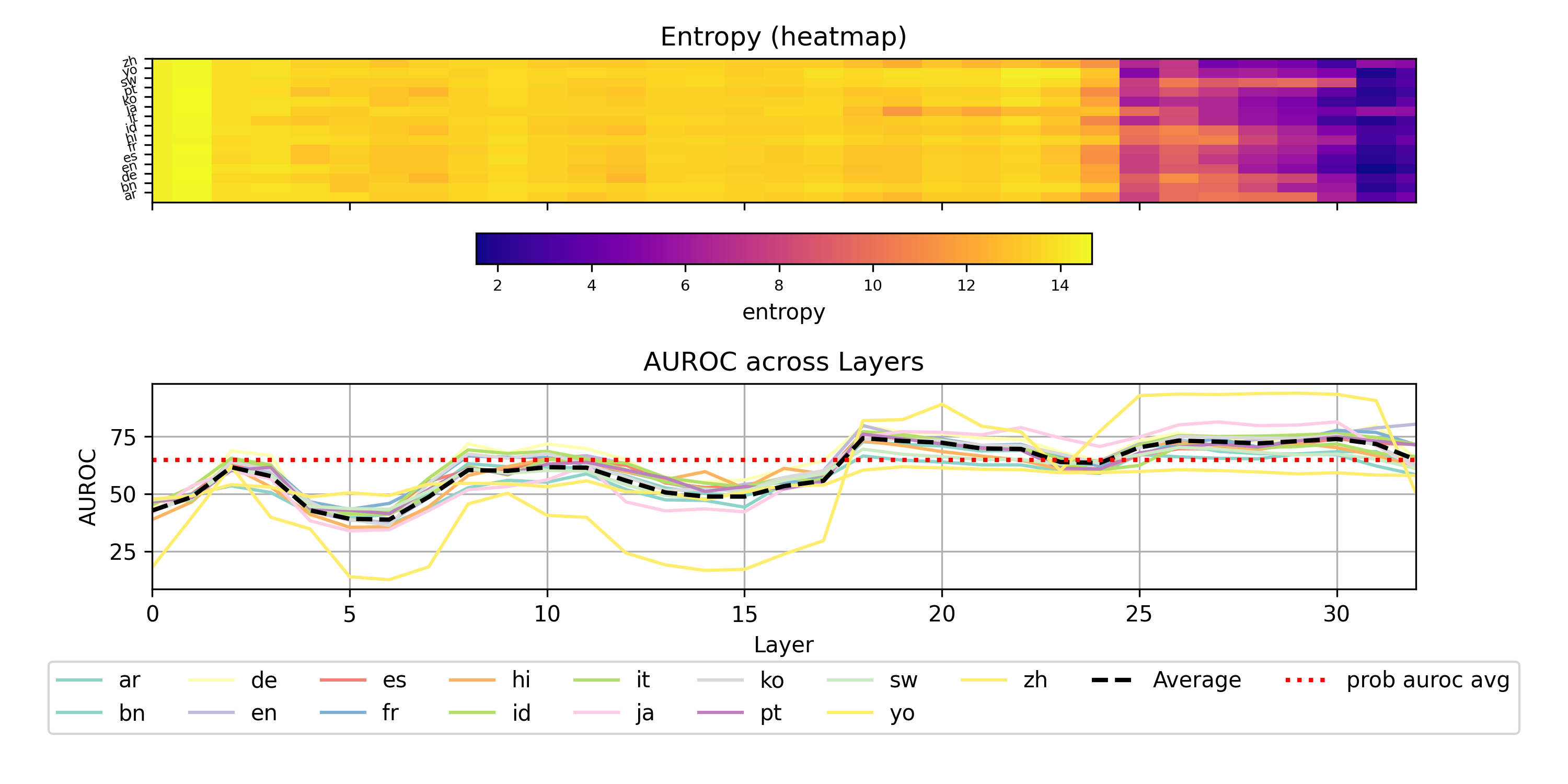}}\\[2pt]
    \subcaptionbox{Aya\label{fig:mmmlu_cohere_calibration_vs_entropy}}{%
        \includegraphics[width=0.32\linewidth]{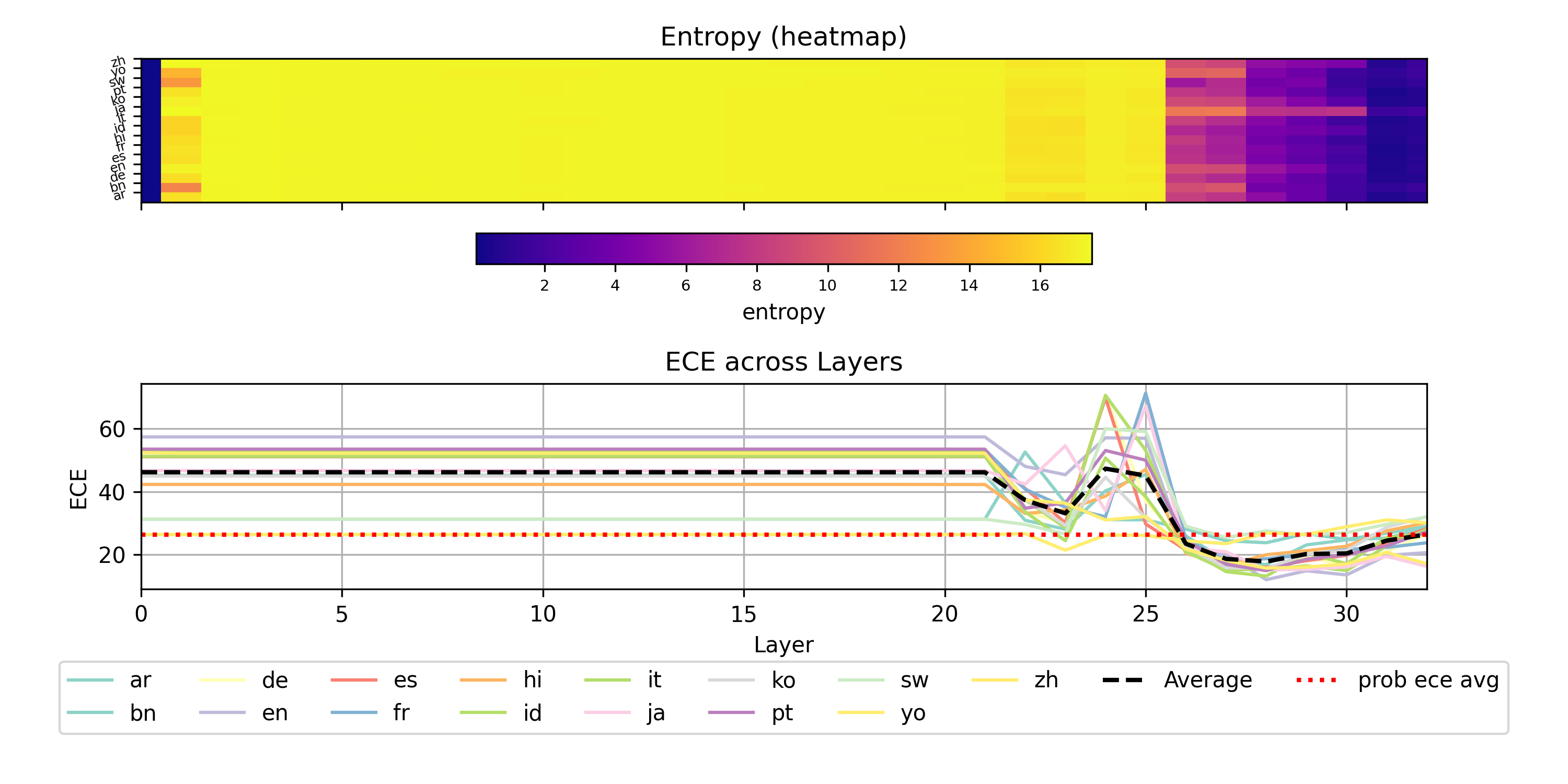}\hfill
        \includegraphics[width=0.32\linewidth]{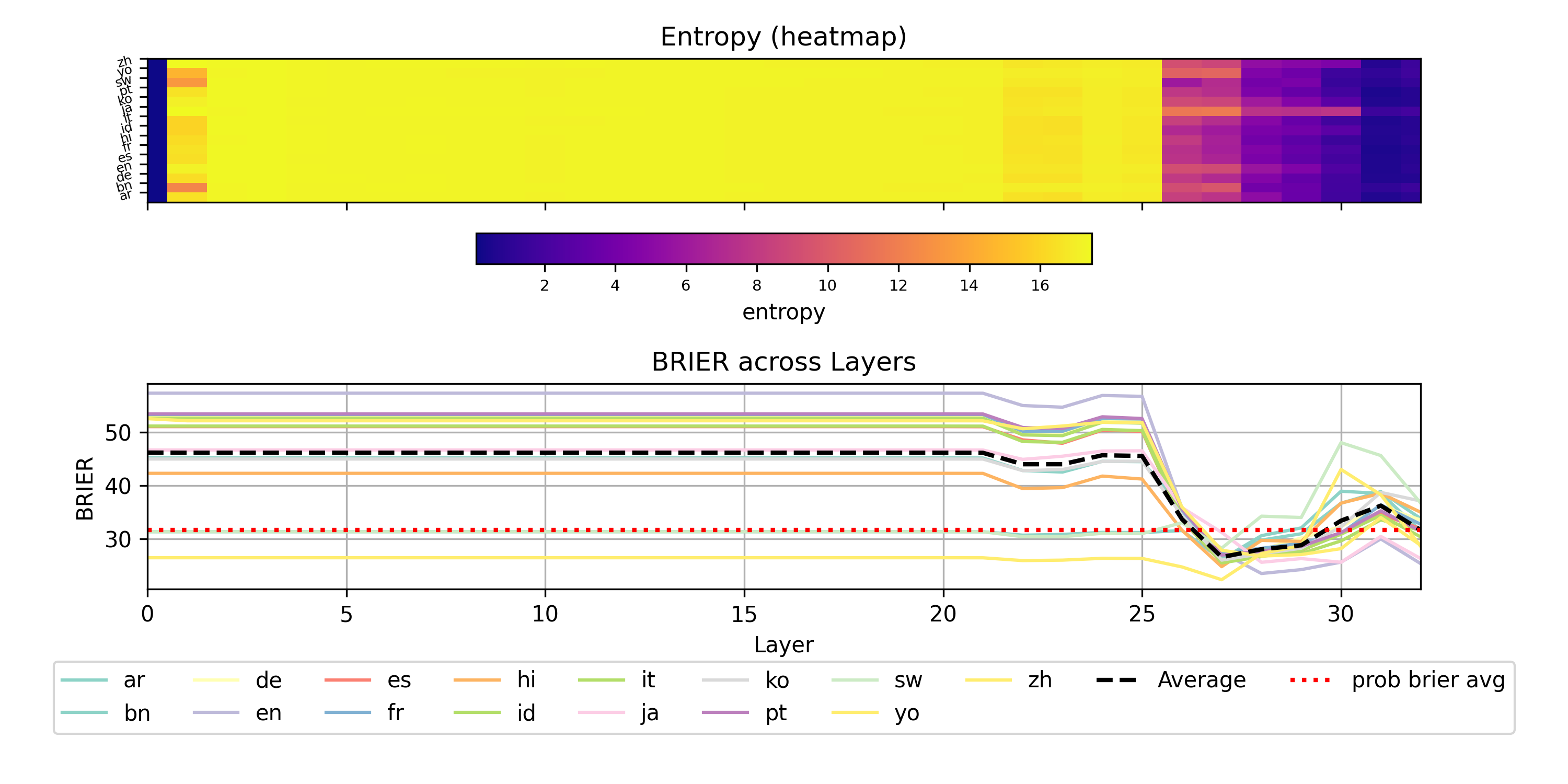}\hfill
        \includegraphics[width=0.32\linewidth]{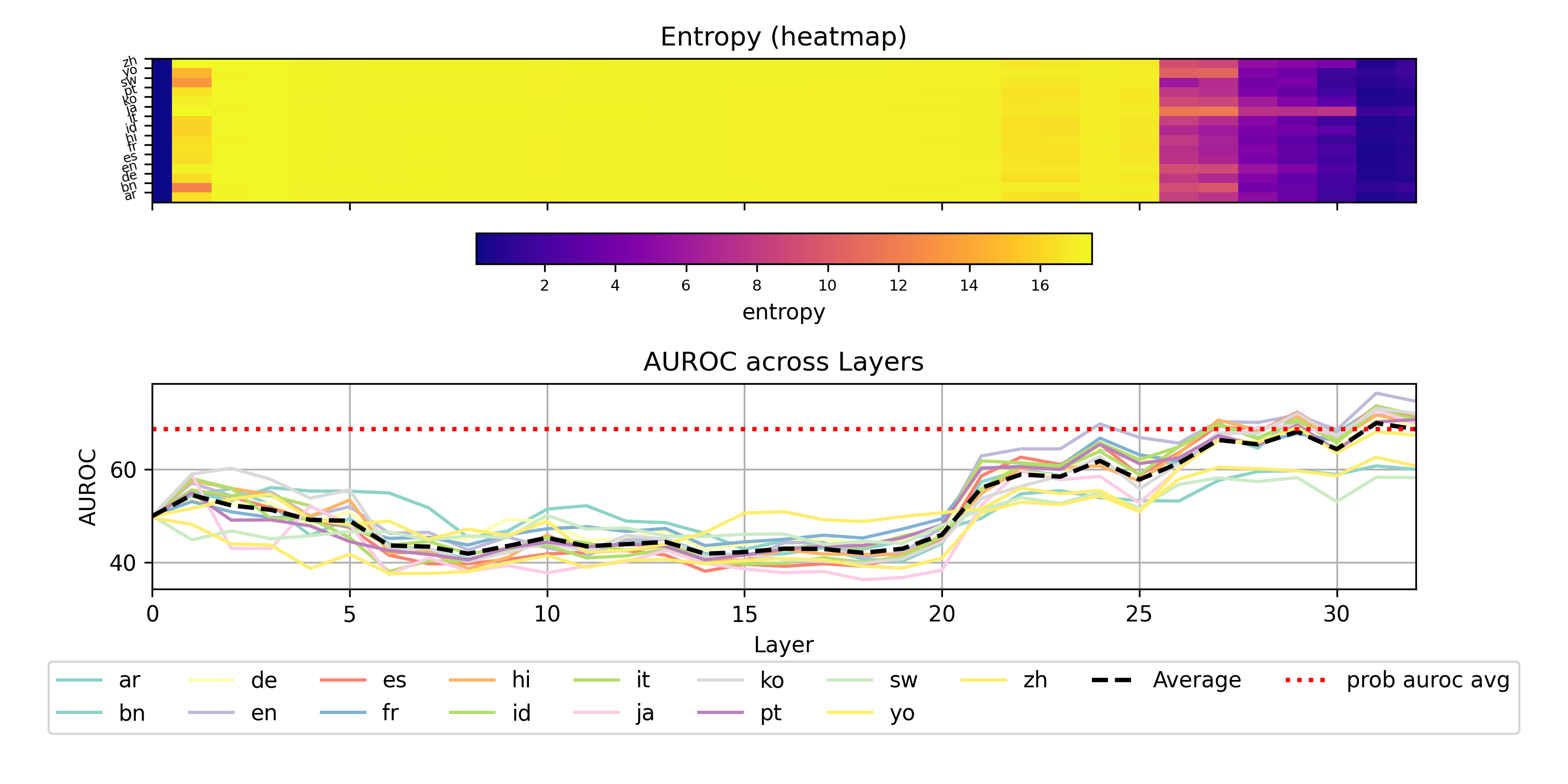}}\\[2pt]
    \subcaptionbox{Mistral\label{fig:mmmlu_mistral_calibration_vs_entropy}}{%
        \includegraphics[width=0.32\linewidth]{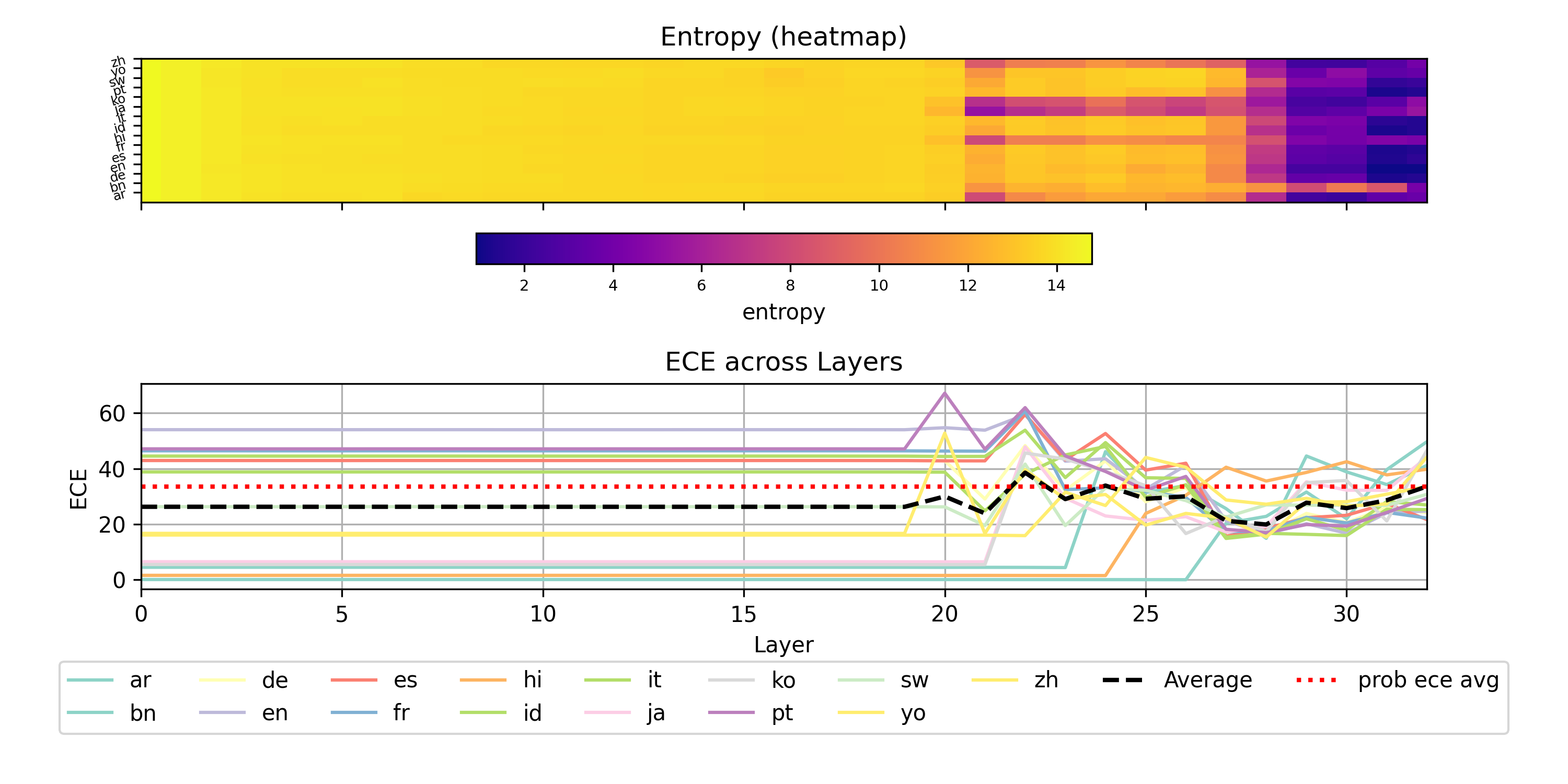}\hfill
        \includegraphics[width=0.32\linewidth]{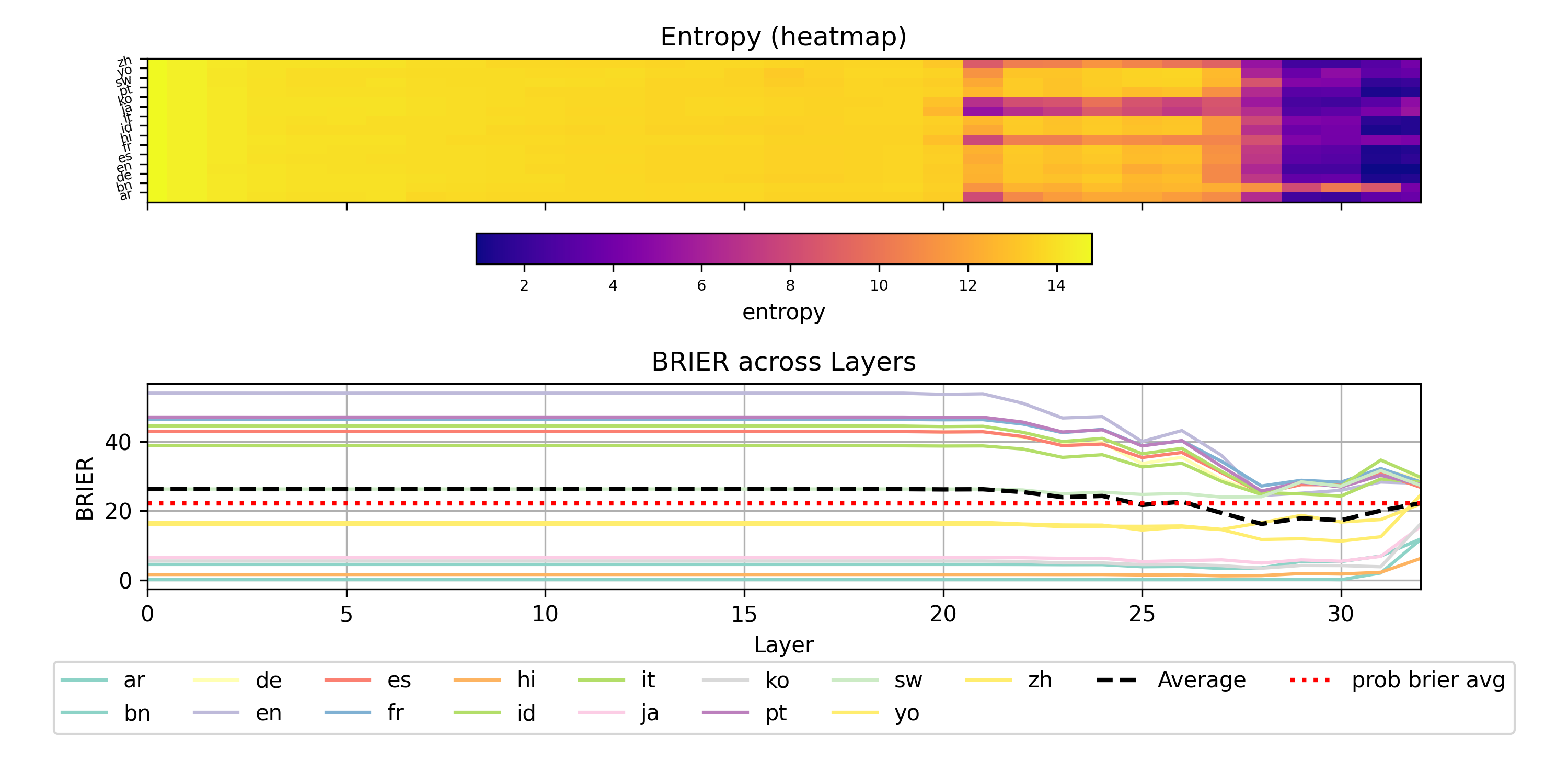}\hfill
        \includegraphics[width=0.32\linewidth]{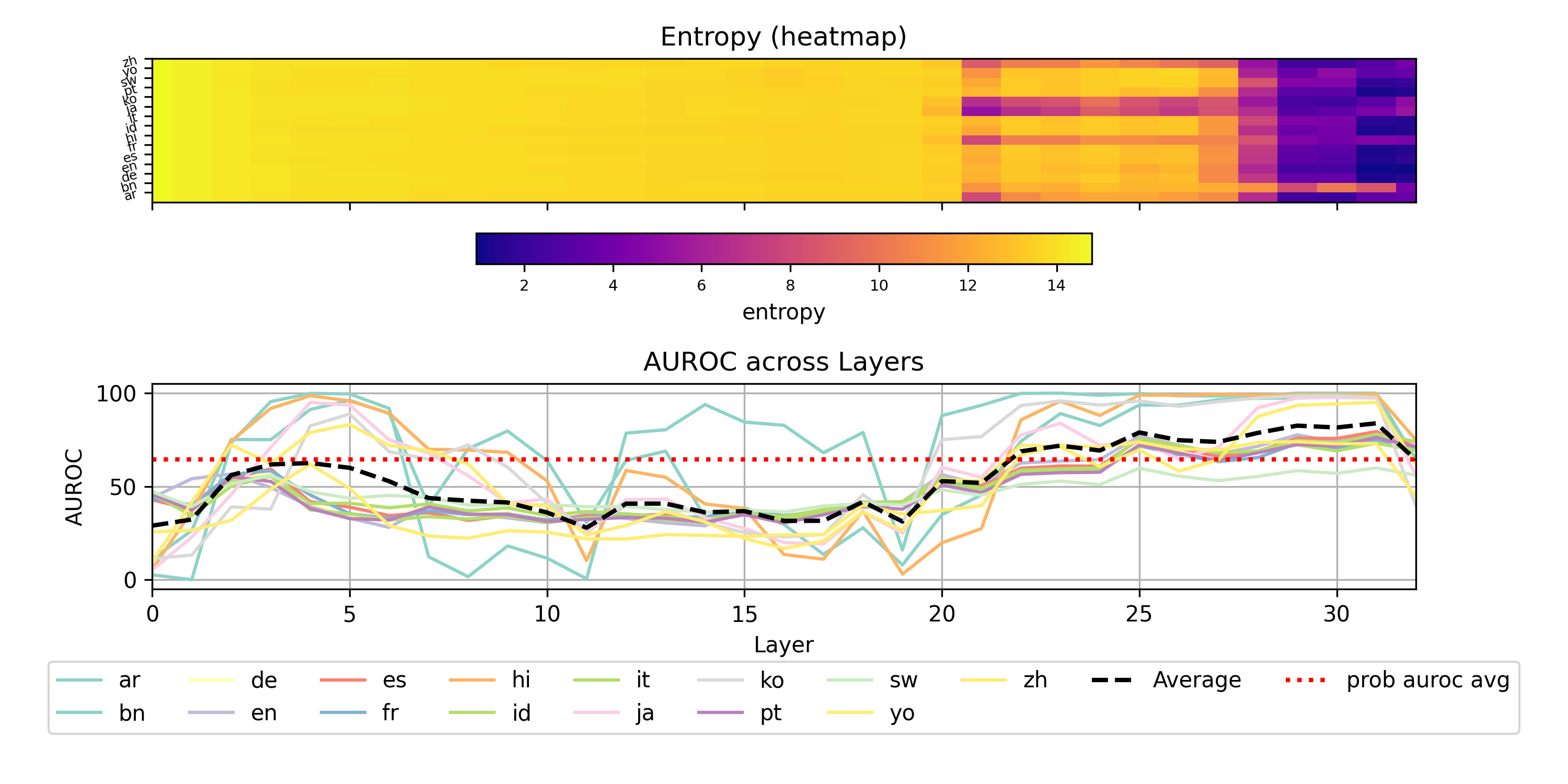}}
    \caption{Calibration metrics (ECE, Brier score, AUROC) vs.\ entropy across layers on the MMMLU dataset for LLaMA3, Aya, and Mistral.}
    \label{fig:mmmlu_calibration_vs_entropy_1}
\end{figure*}

\begin{figure*}[t]
    \centering
    \subcaptionbox{Phi\label{fig:mmmlu_phi_calibration_vs_entropy}}{%
        \includegraphics[width=0.32\linewidth]{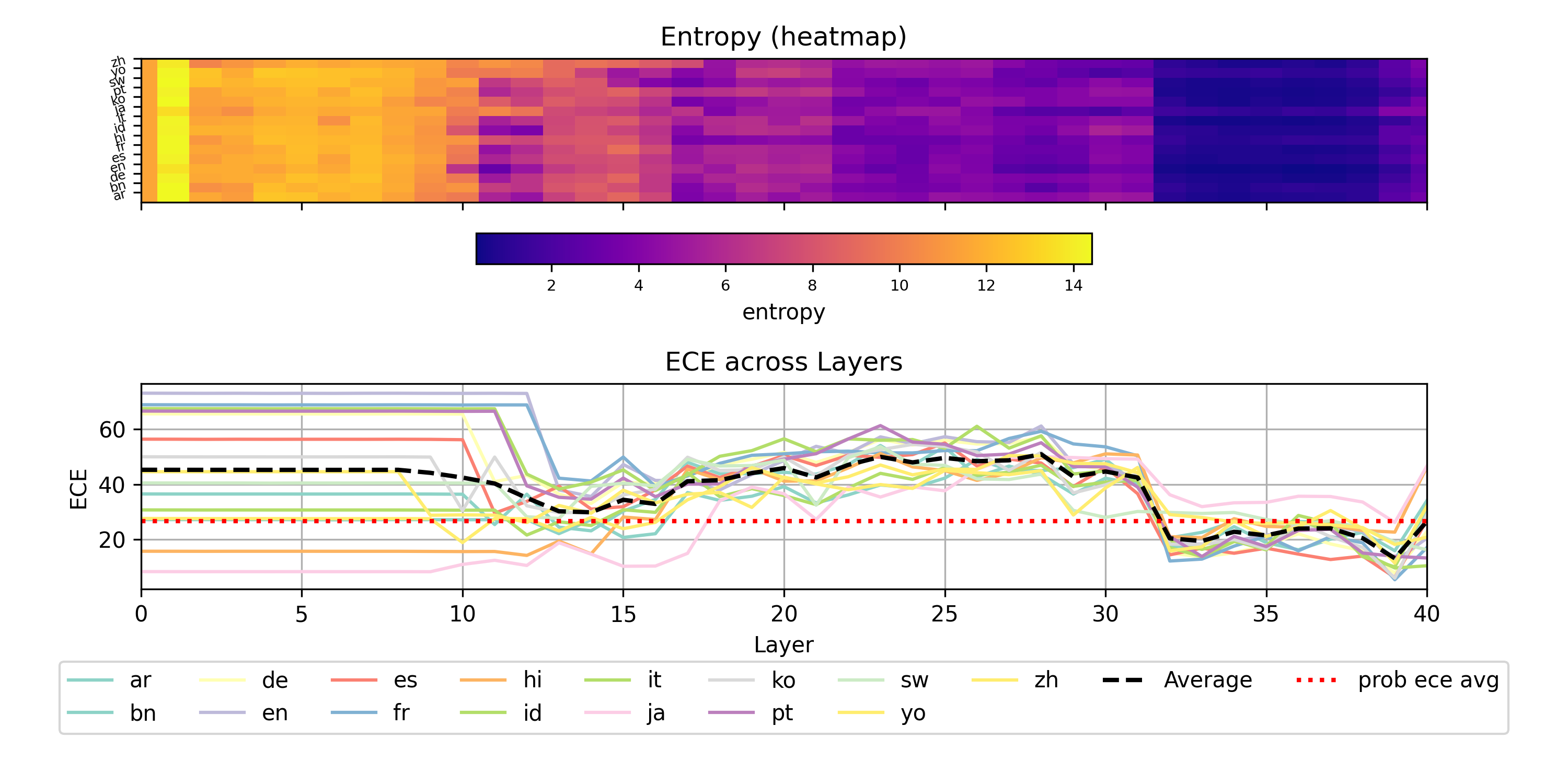}\hfill
        \includegraphics[width=0.32\linewidth]{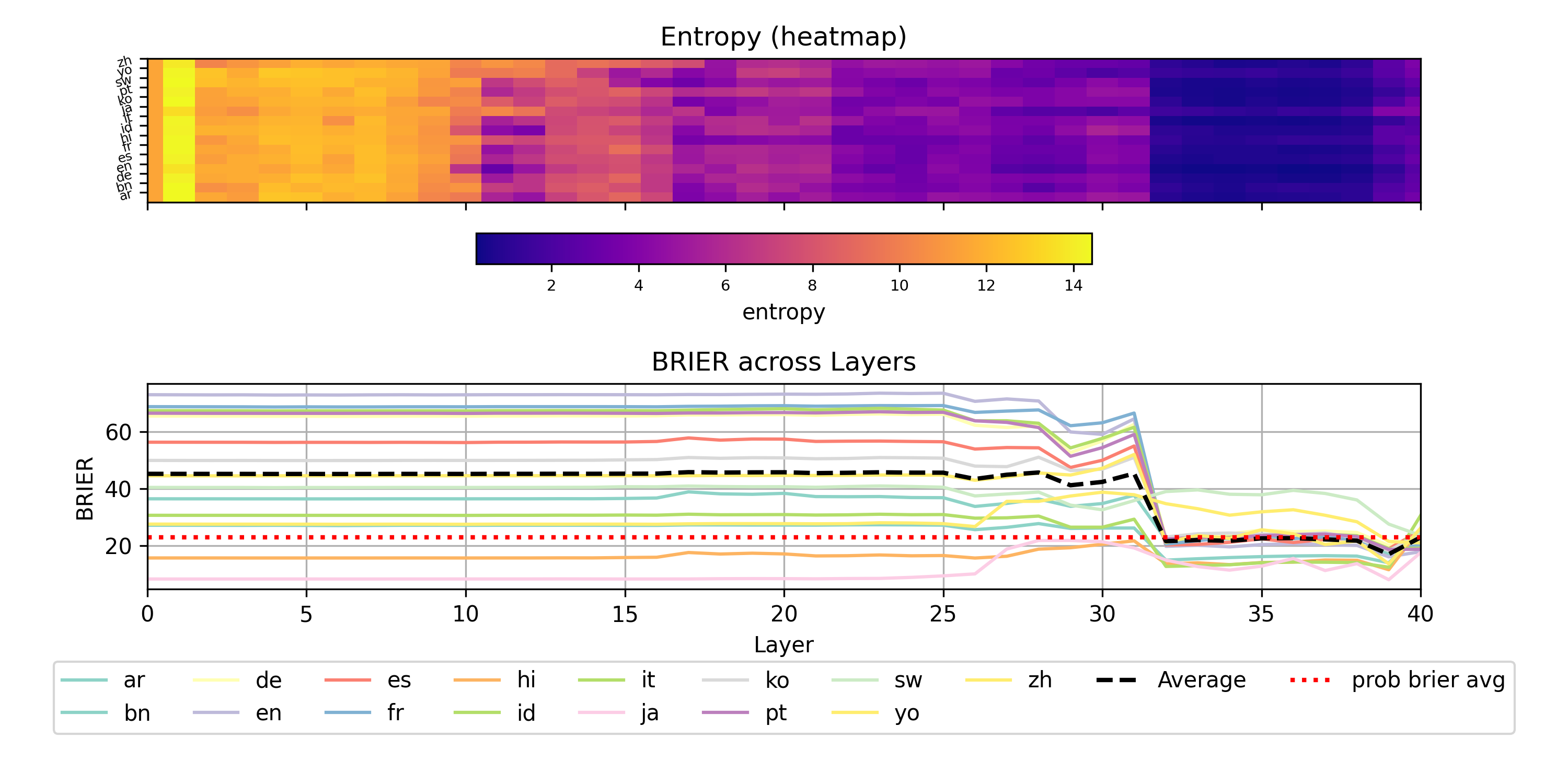}\hfill
        \includegraphics[width=0.32\linewidth]{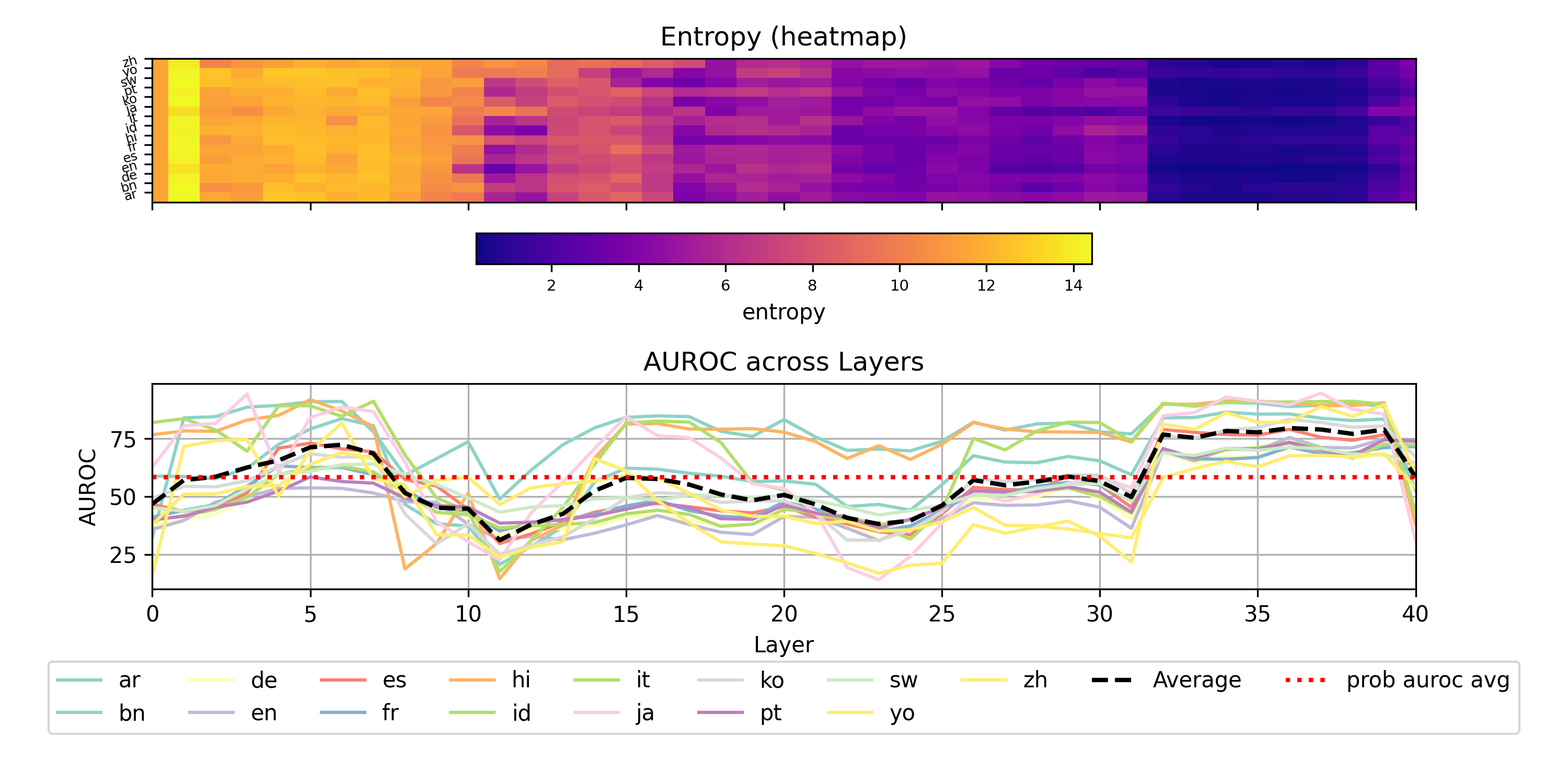}}\\[2pt]
    \subcaptionbox{Deepseek\label{fig:mmmlu_ds_calibration_vs_entropy}}{%
        \includegraphics[width=0.32\linewidth]{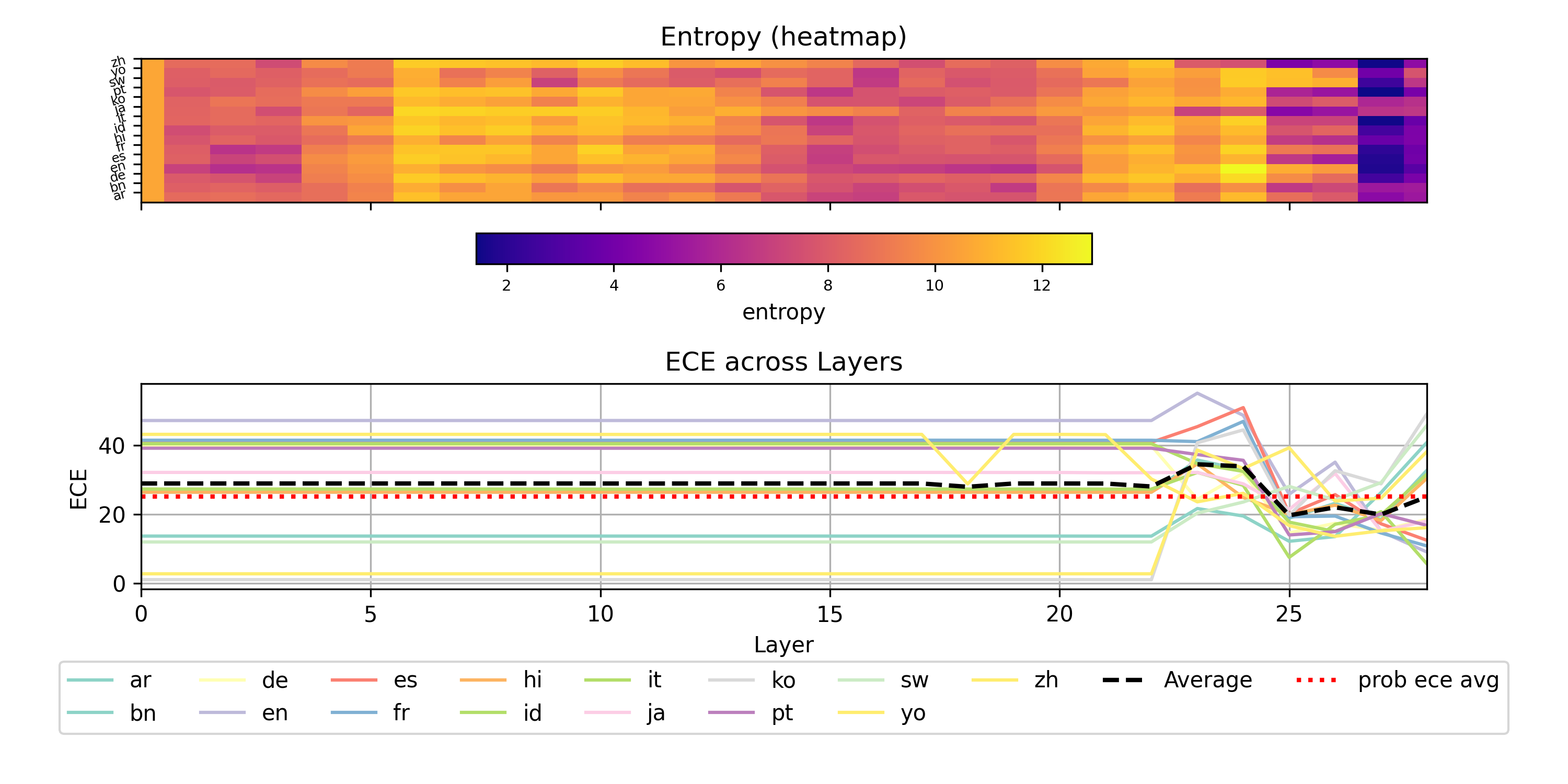}\hfill
        \includegraphics[width=0.32\linewidth]{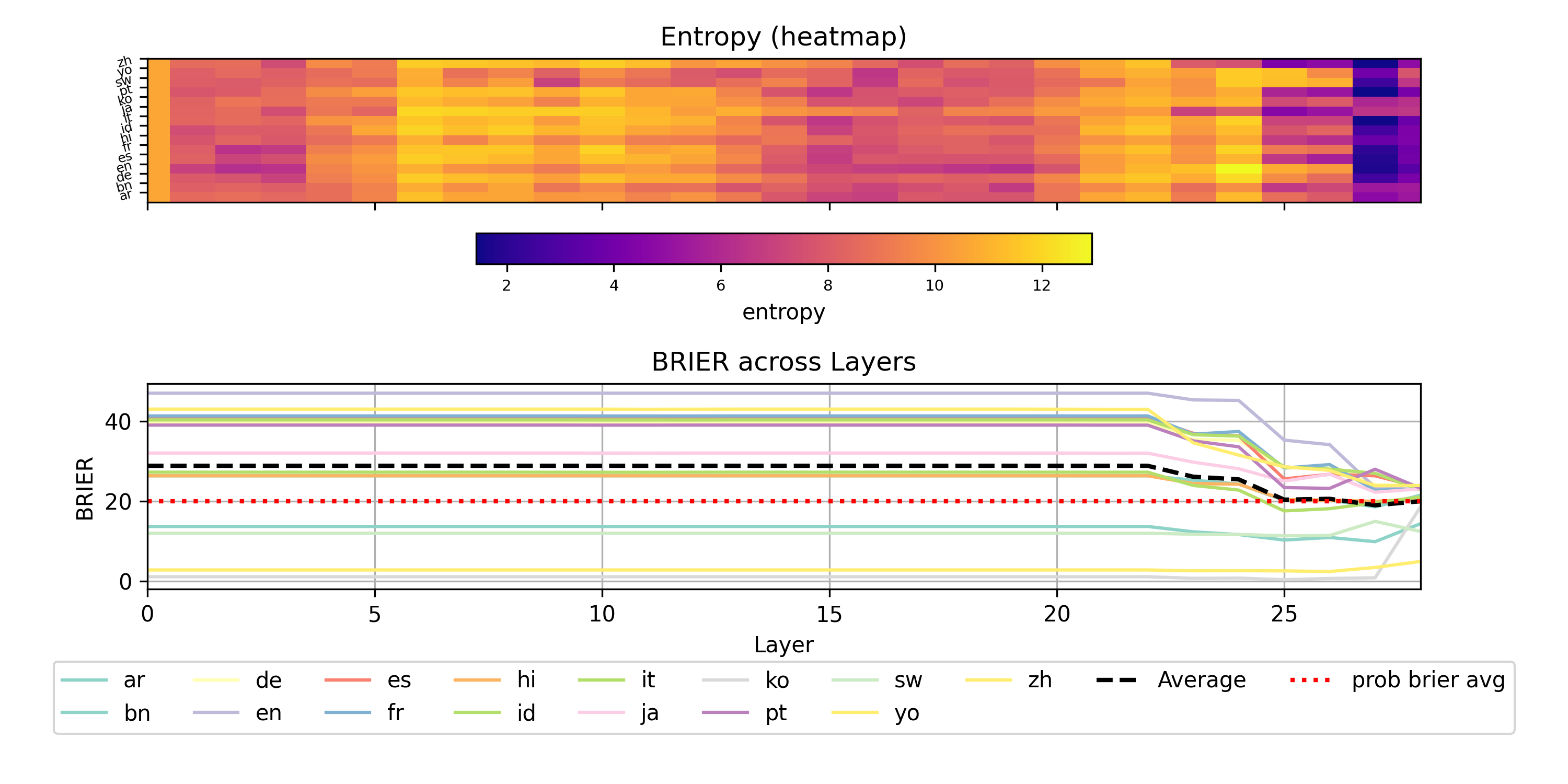}\hfill
        \includegraphics[width=0.32\linewidth]{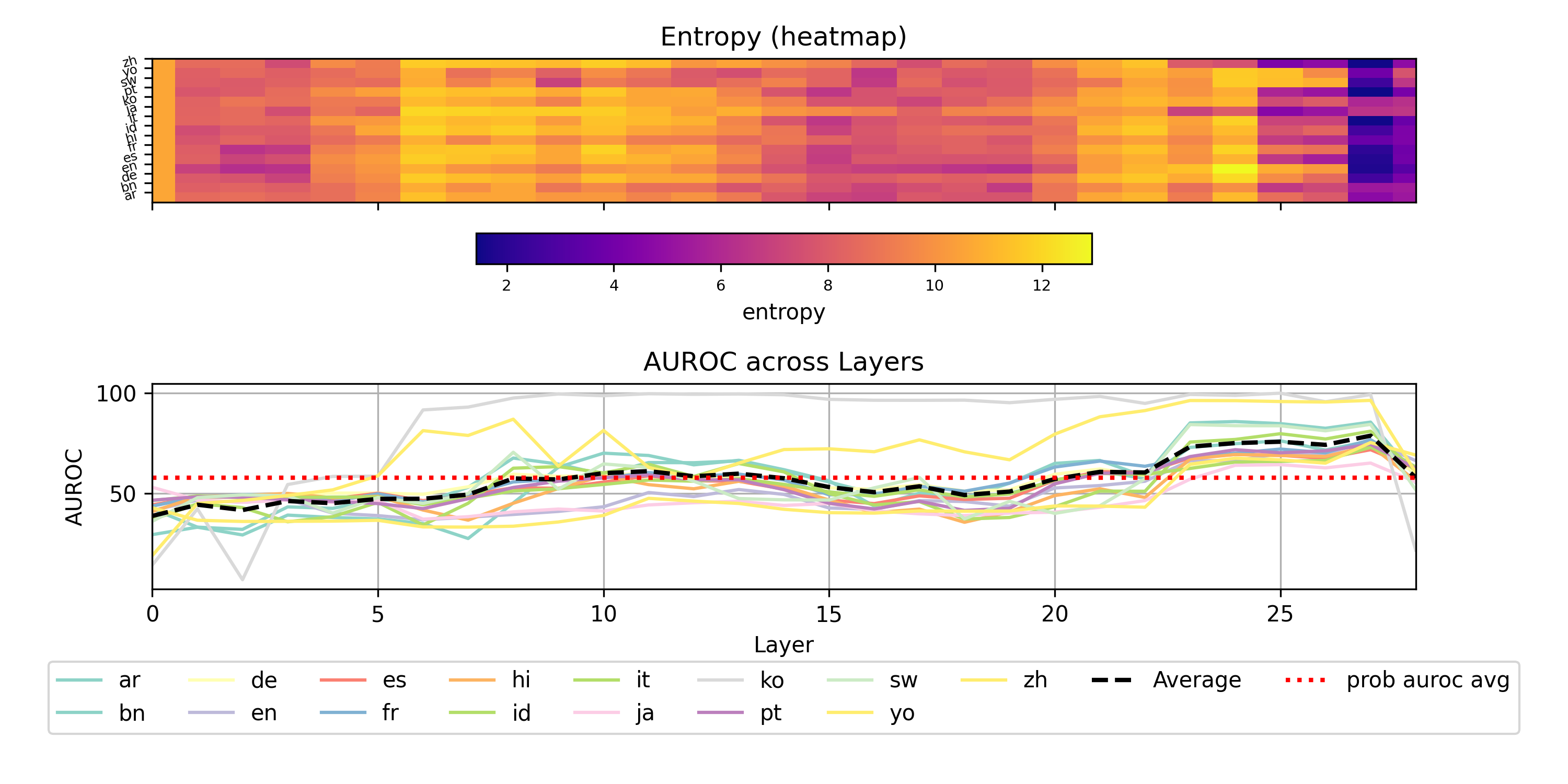}}\\[2pt]
    \subcaptionbox{Qwen 2.5\label{fig:mmmlu_qwen_calibration_vs_entropy}}{%
        \includegraphics[width=0.32\linewidth]{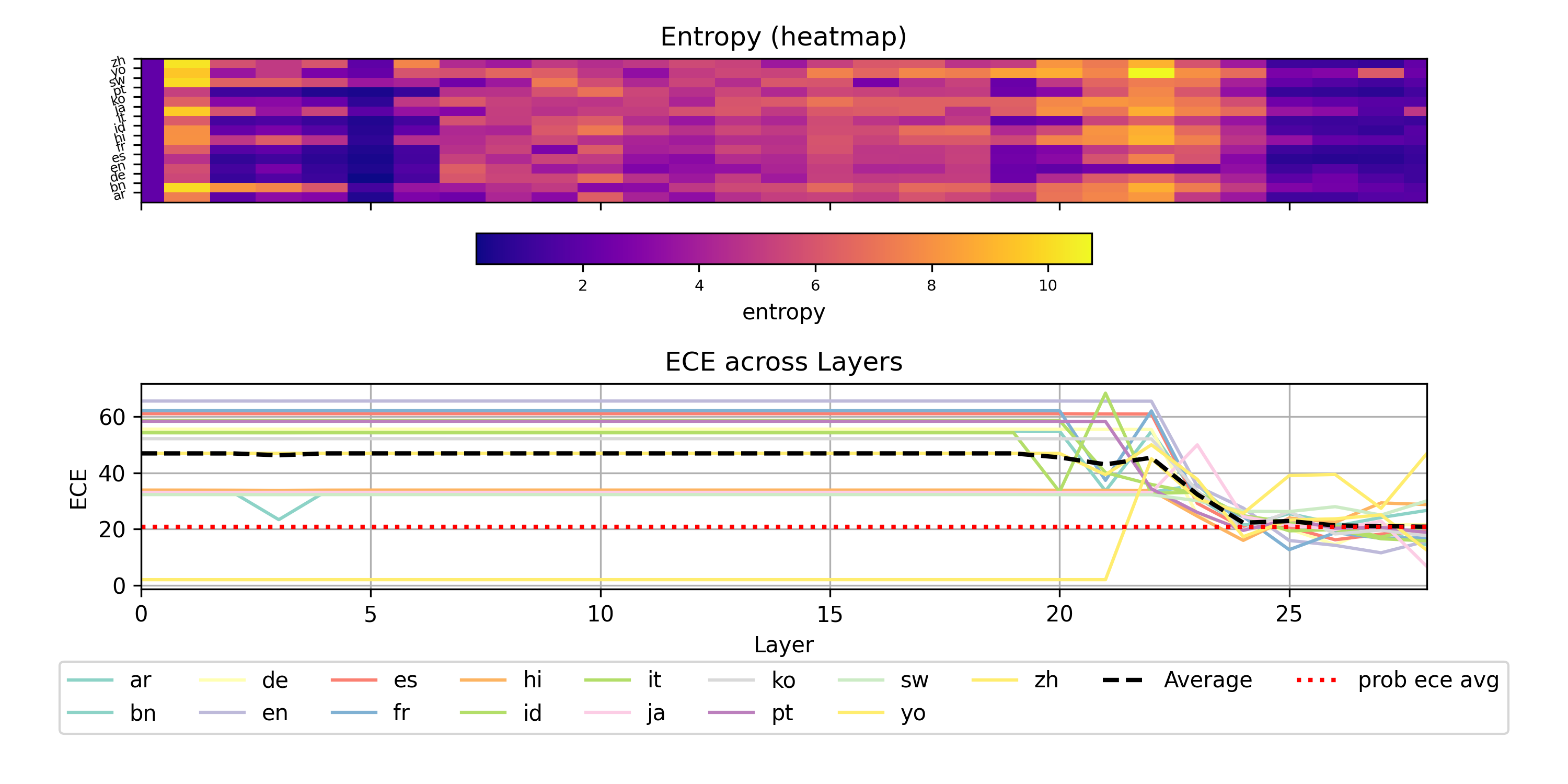}\hfill
        \includegraphics[width=0.32\linewidth]{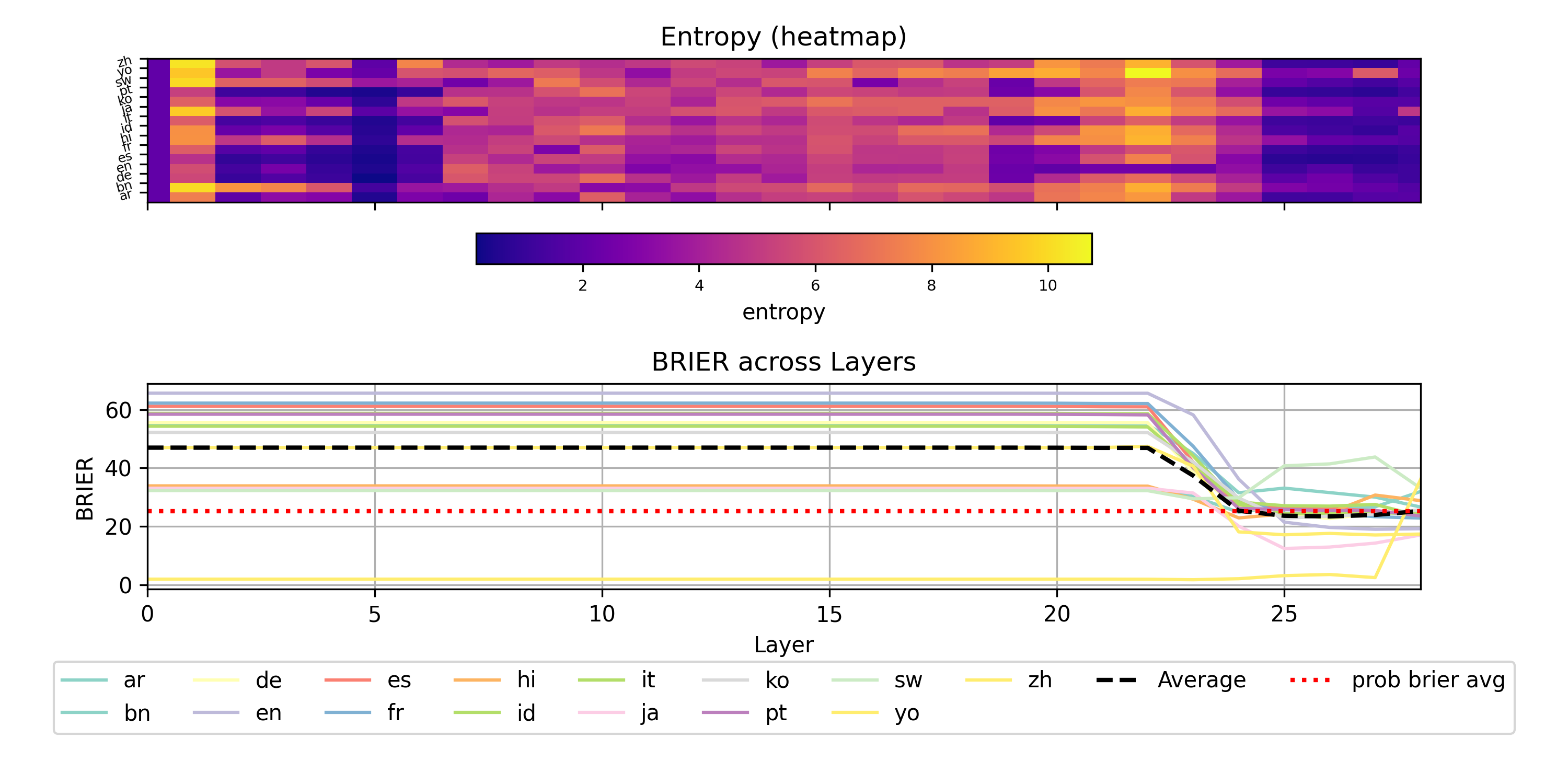}\hfill
        \includegraphics[width=0.32\linewidth]{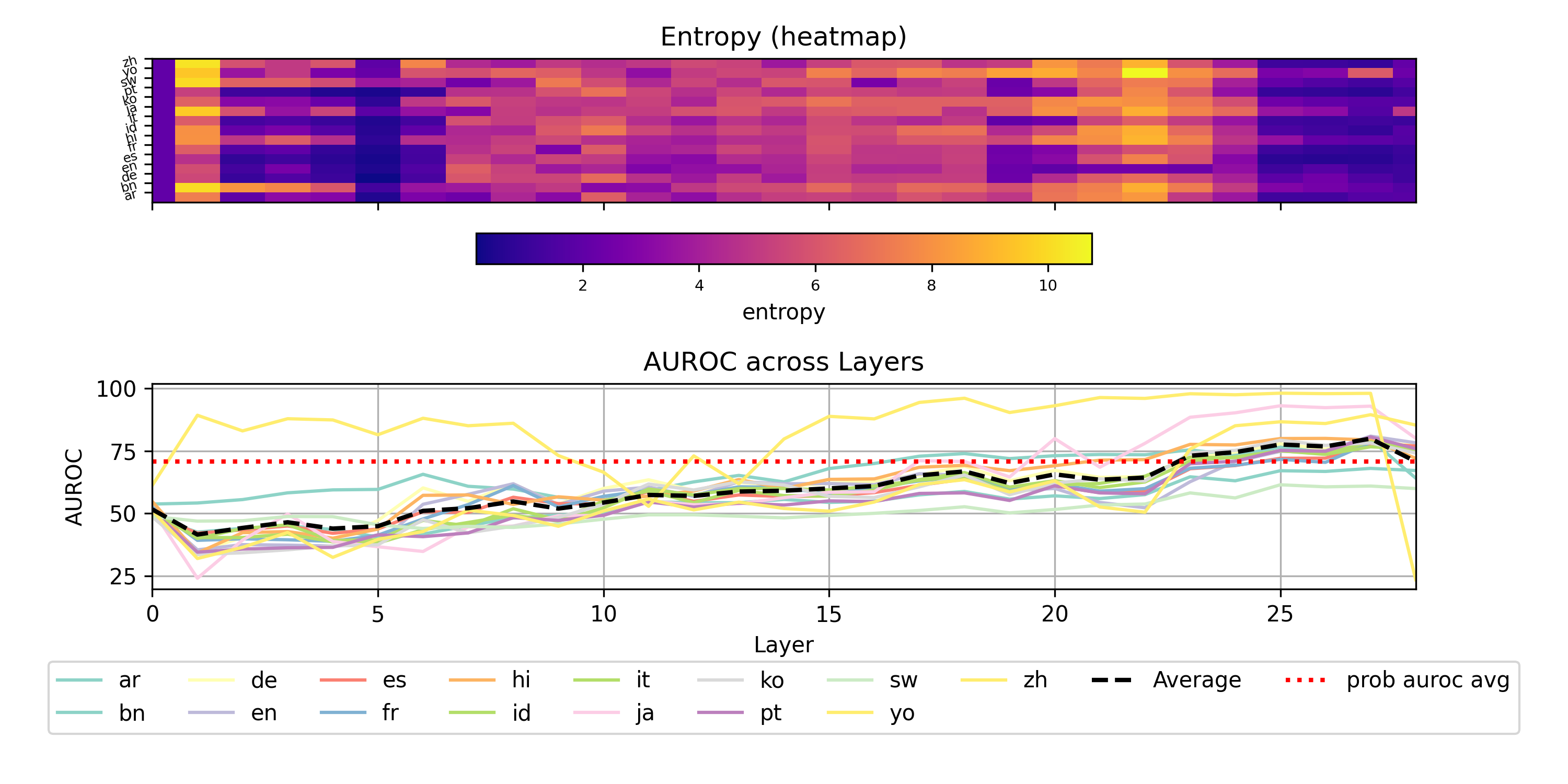}}
    \caption{Calibration metrics (ECE, Brier score, AUROC) vs.\ entropy across layers on the MMMLU dataset for Phi, Deepseek, and Qwen 2.5.}
    \label{fig:mmmlu_calibration_vs_entropy_2}
\end{figure*}

We visualise calibration performance across layers by plotting metrics against entropy on the MMMLU dataset for LLaMA3 (Figure~\ref{fig:mmmlu_llama3_calibration_vs_entropy}), Aya (Figure~\ref{fig:mmmlu_cohere_calibration_vs_entropy}), Mistral (Figure~\ref{fig:mmmlu_mistral_calibration_vs_entropy}), Phi (Figure~\ref{fig:mmmlu_phi_calibration_vs_entropy}), DeepSeek (Figure~\ref{fig:mmmlu_ds_calibration_vs_entropy}), and Qwen2.5 (Figure~\ref{fig:mmmlu_qwen_calibration_vs_entropy}).

\subsection{Reliability Diagrams}
\label{sec:reliablility}

\begin{figure*}[t]
  \centering
  \includegraphics[width=0.82\linewidth]{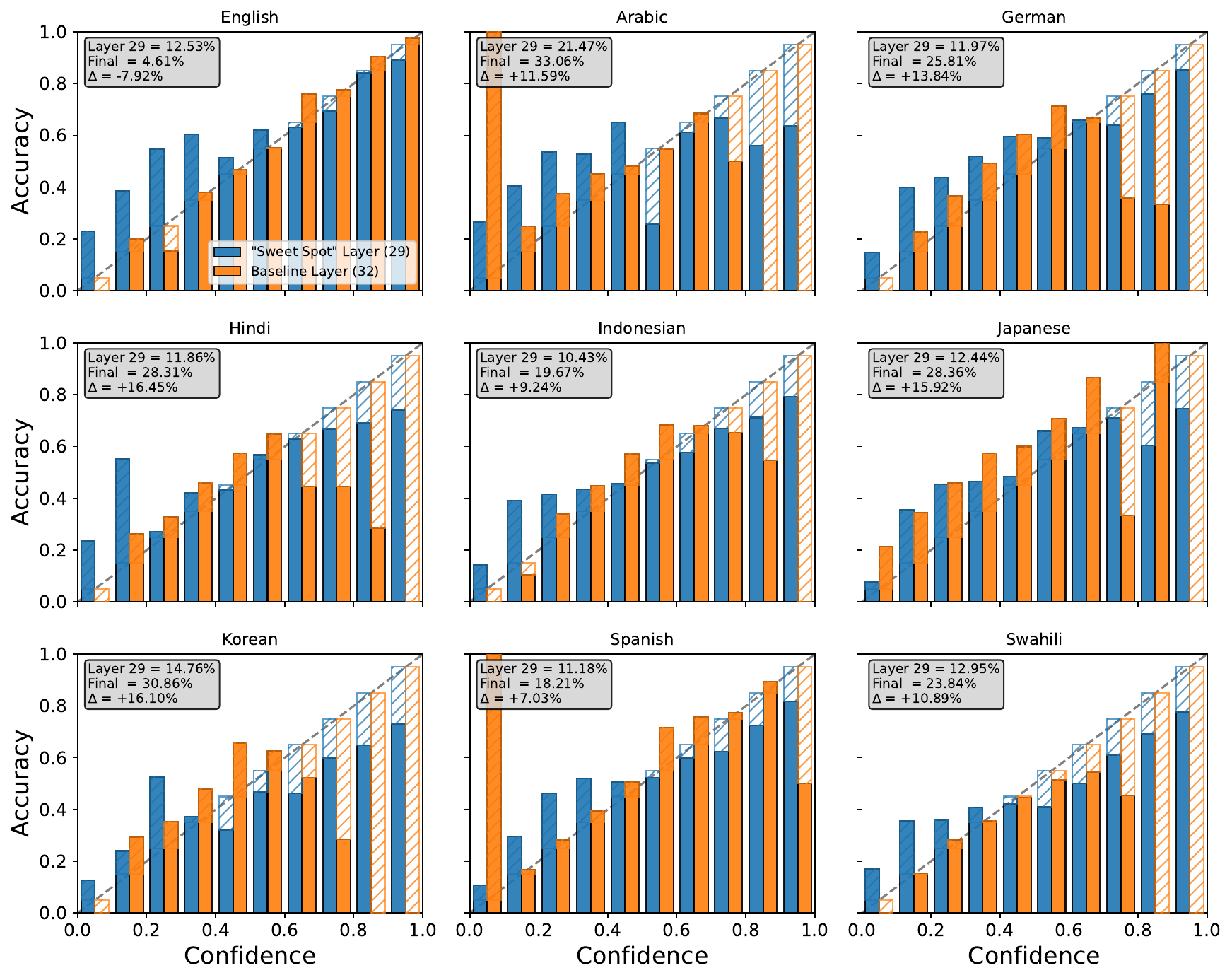}
  \caption[
    Per-language calibration reliability: Layer 29 vs.\ Final (LLaMA3)
  ]{
    \textbf{Per-language calibration reliability diagrams for LLaMA3.}
    Each panel shows a reliability histogram with evenly spaced confidence bins.
    \textcolor{blue}{\textbf{Blue}} bars correspond to the chosen intermediate layer (Layer~29),
    and \textcolor{orange}{\textbf{orange}} bars correspond to the original final layer.
    The dashed diagonal is the perfectly calibrated line ($y{=}\,x$).
    Hatched overlays indicate the absolute calibration gap within each bin.
    The inset reports ECE (\%) for both layers and
    the change $\Delta\text{ECE} = \text{ECE}_{\text{Final}} - \text{ECE}_{29}$
    (positive values denote improved calibration at Layer~29).
  }
  \label{fig:ece-reliability-per-language}
\end{figure*}

\textbf{Per-language calibration trends reveal that late-intermediate layer improves calibration for most non-English languages, with a slight trade-off for English.} 
To further explore these dynamics, Figure~\ref{fig:ece-reliability-per-language} presents per-language reliability diagrams for nine languages, comparing the selected intermediate layer (Layer~29) against the final layer of LLaMA3. Nearly all non-English languages benefit greatly from moving away from the final layer: their reliability curves align more closely with the diagonal, and ECE decreases substantially. For example, German and Hindi show reductions in ECE of more than 13\% and 16\%, respectively.
In contrast, calibration for English appears to degrade slightly at intermediate layers. Since the final layer already exhibits strong calibration for English, earlier layers offer no additional benefit. This highlights a potential bias introduced during pretraining, where the model overfits to English patterns or introduces noise during the final stages of adaptation. While the degree of improvement varies across languages, the trend remains consistent across diverse linguistic families and scripts. This suggests that the effect is not language-specific but rather a systematic property of multilingual calibration.

\section{LACE --- Supporting Details \& Robustness}
\label{sec:lace_details}
This section provides implementation details for the post-hoc calibration baselines used alongside LACE, followed by a series of robustness and generalisation analyses.

\subsection{Post-hoc Calibration Baselines}
\label{sec:baseline}
We include two widely used post-hoc calibration methods as baselines: \textbf{Temperature Scaling}~\citep{guo2017calibration} and \textbf{Isotonic Regression}~\citep{zadrozny2002transforming}. Both are trained on a held-out validation set (15k examples from MMMLU and 12k examples from Belebele, non-overlapping with the evaluation sets). The fitted calibrators are then applied to test-set predictions, and metrics (ECE, AUROC, Brier score, Accuracy) are computed using the calibrated probabilities.

\paragraph{Temperature Scaling}
Temperature scaling applies a single scalar parameter $T > 0$ to rescale logits before computing probabilities. The parameter is optimised by minimising the \textbf{negative log-likelihood (NLL)} on the validation set. In practice, we perform a coarse-to-fine grid search: first over a wide range ($T \in [0.05, 5.0]$ with 60 candidates), then locally refining around the best value with a denser grid. This procedure provides stable estimates across languages and avoids degenerate minima. The resulting optimal temperature is then used to rescale logits of all models prior to evaluation.

\paragraph{Isotonic Regression}
Isotonic regression learns a non-parametric, monotone mapping from predicted probabilities to calibrated probabilities in $[0,1]$. We use the \texttt{scikit-learn} implementation with out-of-bounds clipping and monotonicity constraints. The model is fitted on the validation set and then applied to test-set predictions.

\subsection{Cross-Lingual Layer-Sharing and Transfer}
\label{sec:crosslingual}

To better understand whether calibration improvements transfer across languages, we analyse the extent to which languages share similar ``good layer'' sets under our intermediate-layer calibration framework. If two languages select highly overlapping sets of good layers, we would expect stronger cross-lingual transfer: optimising for one language could benefit the other. Conversely, disjoint sets would suggest that calibration signals are language-specific.

\paragraph{Measuring cross-lingual similarity.}
For each language $\ell$, let $G_\ell$ denote its selected set of good layers. We compute a language--language similarity matrix using the Jaccard index:
\[
    \text{Jaccard}(\ell_i, \ell_j)
    = \frac{| G_{\ell_i} \cap G_{\ell_j} |}{| G_{\ell_i} \cup G_{\ell_j} |}.
\]
A score of $1.0$ indicates identical layer sets, while $0.0$ means no overlap. Higher similarity reflects stronger potential cross-lingual transfer.

\begin{figure}[h]
    \centering
    \begin{subfigure}{0.48\linewidth}
        \centering
        \includegraphics[width=\linewidth]{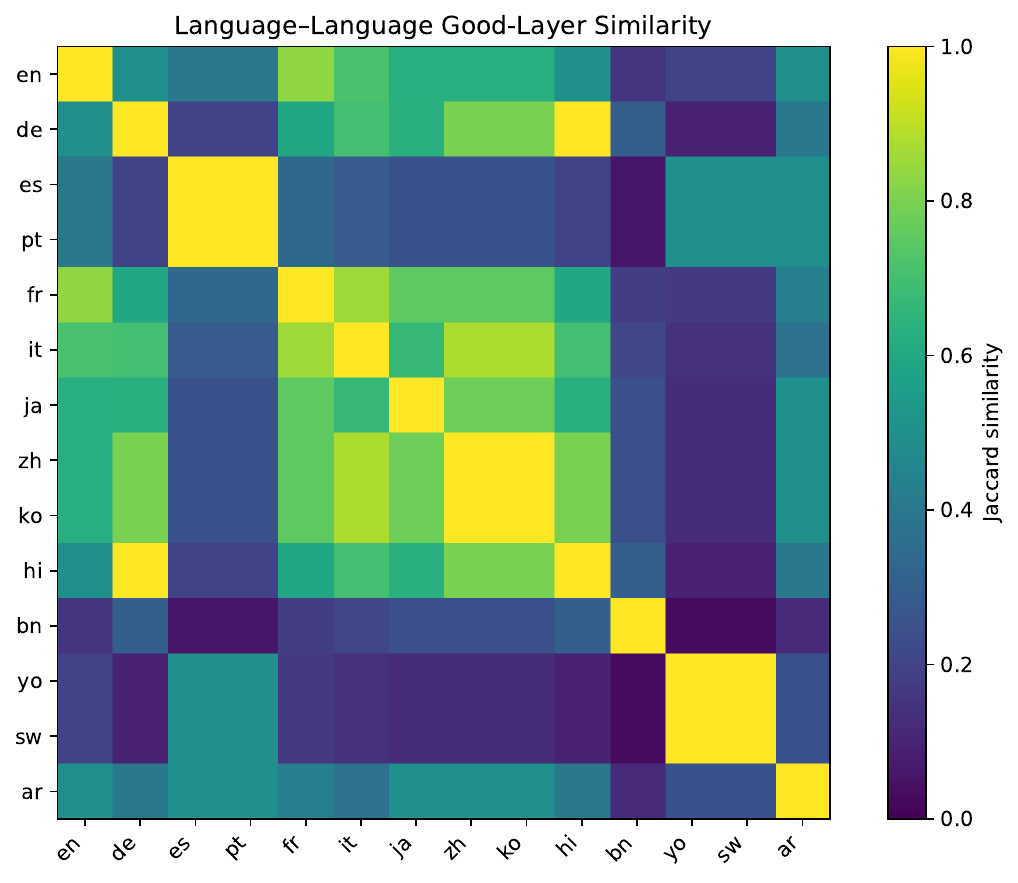}
        \caption{LLaMA-3 8B}
        \label{fig:similarity_8b}
    \end{subfigure}
    \hfill
    \begin{subfigure}{0.48\linewidth}
        \centering
        \includegraphics[width=\linewidth]{plot/language_similarity_grouped_70B.png}
        \caption{LLaMA-3 70B}
        \label{fig:similarity_70b}
    \end{subfigure}
    \caption{
        Cross-lingual Jaccard similarity between languages' good-layer sets.
        Clear correlation between Western European and CJK language groups indicate stronger organisation of calibration-relevant information in deeper layers.
    }
    \label{fig:similarity_combined}
\end{figure}

\paragraph{Findings.}
Figure~\ref{fig:similarity_combined} shows the cross-lingual similarity matrices for LLaMA-3 8B and 70B. While both models show some alignment among typologically related languages, the 70B model exhibits \emph{far clearer and more coherent clustering}, reflecting stronger structural organisation in its intermediate representations. In particular:

\begin{itemize}
    \item \textbf{Western European languages} (e.g., German, Spanish, Portuguese, French, Italian) share overlapping good-layer subsets.
    \item \textbf{CJK languages} (Japanese, Chinese, Korean) also form a consistent cluster.
    \item Languages across distant families (e.g., English vs.\ CJK; Romance vs.\ Indic or African languages) show much lower overlap.
\end{itemize}

Example good-layer sets within Romance languages in the 70B model illustrate this structured overlap:
\begin{itemize}
    \item \texttt{de}: [72, 75, 76, 77, 78, 79, 80]  
    \item \texttt{es}: [72, 73, 75, 76, 77, 78, 79, 80]  
    \item \texttt{pt}: [71, 72, 73, 75, 76, 77, 78, 79, 80]  
    \item \texttt{fr}: [77, 78, 79, 80]  
    \item \texttt{it}: [72, 77, 78, 79, 80]  
\end{itemize}

Overall, these results confirm that \textbf{similar languages tend to share partially overlapping calibration-relevant layer patterns}, while more distant languages exhibit far less similarity.

\subsection{Cross-Dataset Consistency of Good-Layer Selection}
\label{sec:crossdataset}

We assess whether ``good layers'' for calibration are consistent across different tasks by performing a cross-dataset layer-overlap analysis. For each language $\ell$ common to both MMMLU and Belebele, we extract the good-layer sets $G^{\text{MMMLU}}$ and $G^{\text{Belebele}}$ and compute their Jaccard similarity:

\[
\text{Jaccard}(m)
=
\frac{1}{|\mathcal{L}_m|}
\sum_{\ell \in \mathcal{L}_m}
\frac{
\left|
G^{\text{MMMLU}}_{m,\ell}
\cap
G^{\text{Belebele}}_{m,\ell}
\right|
}{
\left|
G^{\text{MMMLU}}_{m,\ell}
\cup
G^{\text{Belebele}}_{m,\ell}
\right|
}.
\]
This evaluation was conducted across the 12 languages common to both datasets. The results show moderate cross-dataset consistency, with an average Jaccard similarity of \textbf{0.5227} for LLaMA-3 and \textbf{0.4704} for Aya. This indicates that despite the different task formulations, the benchmarks identify partially overlapping sets of calibration-relevant layers. This suggests that certain structural properties of intermediate representations support robust calibration across tasks, though task-specific effects remain.

\paragraph{Transfer evaluation on MMMLU using Belebele-derived layers.}

To further test generalisation, we applied the good-layer sets identified on Belebele directly to MMMLU without any post-hoc calibration. The results, shown in Table~\ref{dataset-transfer}, confirm that Belebele-derived layers retain meaningful calibration performance on MMMLU, demonstrating clear cross-task transferability.

\begin{table}[h]
\centering
\small
\setlength{\tabcolsep}{5pt}
\renewcommand{\arraystretch}{1.1}
\resizebox{\linewidth}{!}{
\begin{tabular}{ll
  S[table-format=2.1]
  S[table-format=2.1]
  S[table-format=2.1]}
\toprule
\textbf{Model} & \textbf{Setting} & {\textbf{ECE} $\downarrow$} & {\textbf{AUROC} $\uparrow$} & {\textbf{Brier} $\downarrow$} \\
\midrule
\multirow{2}{*}{LLaMA-3} 
 & baseline & 22.2 & 66.1 & 23.3 \\
 & \textit{transferred}-LACE                 & \textbf{15.6} & \textbf{70.6} & \textbf{21.8} \\
\midrule
\multirow{2}{*}{Aya} 
 & baseline & 27.4 & 68.9 & 32.3 \\
 & \textit{transferred}-LACE                 & \textbf{19.8} & \textbf{69.3} & \textbf{25.2} \\
\bottomrule
\end{tabular}
}
\caption{Calibration transfer results when applying \textbf{Belebele-derived good layers} directly to \textbf{MMMLU} (without post-hoc calibration).}
\label{dataset-transfer}
\end{table}

\subsection{Robustness to Language-ID Noise}
\label{sec:langid-robustness}

To quantify how sensitive LACE is to noisy language labels, we run a controlled corruption experiment on MMMLU for two base models (Aya and LLaMA-3). 

We start from our setup in the main paper where language-specific layer sets are selected on the validation split using gold language labels. Then at evaluation time, we simulate language-ID mistakes by randomly corrupting the language tag for a proportion \(p \in \{0.0, 0.1, 0.2\}\) of test examples:
for each selected example, we replace its true language label with a \emph{different} language sampled from the remaining languages. LACE uses these (possibly incorrect) language tags to choose which language-specific layer set to average, while the underlying model logits remain unchanged.

Therefore, \(p = 0\) corresponds to perfect language identification, whereas \(p = 0.1\) and \(p = 0.2\) approximate moderate error rates in an external language-ID system; state-of-the-art LID systems achieve above 90\% accuracy on sentence-level text \citep{suarez-etal-2026-commonlid}, so \(p = 0.2\) is a conservative upper bound on realistic misidentification rates. We compute ECE, AUROC, and Brier score for each corruption level. The results are shown in Table~\ref{tab:lid_corruption}.

\begin{table}[h]
\centering
\small
\setlength{\tabcolsep}{5pt}
\renewcommand{\arraystretch}{1.1}
\resizebox{\linewidth}{!}{
\begin{tabular}{cl
  S[table-format=2.2]
  S[table-format=2.2]
  S[table-format=2.2]}
\toprule
\textbf{Corruption} & \textbf{Model} & {\textbf{ECE (\%)}} & {\textbf{AUROC (\%)}} & {\textbf{Brier (\%)}} \\
\midrule
\multirow{2}{*}{0\%}  
    & LLaMA-3 & 5.96  & 72.94 & 19.91 \\
    & Aya     & 11.42 & 68.61 & 23.16 \\
\midrule
\multirow{2}{*}{10\%} 
    & LLaMA-3 & 5.97  & 72.80 & 19.97 \\
    & Aya     & 11.60 & 68.38 & 23.31 \\
\midrule
\multirow{2}{*}{20\%} 
    & LLaMA-3 & 6.29  & 72.39 & 20.15 \\
    & Aya     & 12.07 & 68.25 & 23.39 \\
\bottomrule
\end{tabular}
}
\caption{Robustness of LACE to language-ID corruption on MMMLU.}
\label{tab:lid_corruption}
\end{table}

The impact of language-ID noise is mild. For Aya, corrupting 10\% of tags increases ECE by only \(0.18\) (about 1.5\% relative) and Brier by \(0.14\), while AUROC decreases by \(0.23\). Even at 20\% corruption, ECE rises by only \(0.65\) compared to the clean setting, and Brier changes by \(0.23\). For LLaMA-3, the effects are even smaller: ECE changes by \(+0.01\) at 10\% and \(+0.33\) at 20\%, with AUROC decreasing by less than \(0.60\) and Brier increasing by at most \(0.24\). These variations are negligible relative to the large cross-lingual calibration gaps reported in the main paper.

Intuitively, LACE remains robust because it averages over multiple ``good'' layers per language rather than relying on a single, highly language-specific layer: misrouting an example to another language's layer set still yields a reasonable ensemble, so moderate language-ID noise only weakly perturbs the final predictions.

\subsection{Sensitivity to Validation Set Size}
\label{app:validation-size-sensitivity}

A potential limitation of LACE is that it requires a labelled validation set for each language in order to identify the corresponding Good Layers. This requirement may appear burdensome for low-resource languages, which are often the settings where improved calibration is most needed. However, in our main experiments, we use only 1{,}000 labelled validation examples per language, which we believe is already a practically attainable amount of data for many languages.

To further examine the data requirement of LACE, we conduct a sensitivity analysis by reducing the validation set size from the full 1{,}000 examples to 500, 200, and 100 examples. For each reduced setting, we recompute the selected Good Layers and compare them against those obtained from the full 1{,}000-example validation set. As a measure of agreement, we report the Jaccard similarity between the two selected layer sets.

Table~\ref{tab:validation-size-sensitivity} shows that the selected Good Layers remain highly consistent when using only 500 validation examples, with an average Jaccard similarity of 0.88 relative to the 1{,}000-example baseline. When the validation size is further reduced to 200 examples, the agreement remains moderate (0.65), while at 100 examples the consistency drops to 0.51. These results suggest that LACE is robust to substantial reductions in validation data, and that in practice as few as 500 labelled examples are sufficient to recover nearly the same layer selection as the full validation set.

\begin{table}[h]
\centering
\small
\setlength{\tabcolsep}{6pt}
\renewcommand{\arraystretch}{1.1}
\begin{tabular}{l
  S[table-format=1.2]
  l}
\toprule
\textbf{Comparison} & {\textbf{Avg.\ Jaccard}} & \textbf{Interpretation} \\
\midrule
{1000 vs.\ 500} & 0.88 & High consistency \\
{1000 vs.\ 200} & 0.65 & Moderate consistency \\
{1000 vs.\ 100} & 0.51 & Lower consistency \\
\bottomrule
\end{tabular}
\caption{Stability of Good Layer selection under reduced validation set sizes, measured by Jaccard similarity against the full 1{,}000-example validation set.}
\label{tab:validation-size-sensitivity}
\end{table}

Overall, this analysis indicates that the validation data requirement of LACE is lower than it may initially seem. While performance naturally degrades as the validation set becomes very small, the method remains highly stable at 500 examples, substantially lowering the barrier to applying LACE in lower-resource settings.

\section{Experiments on the 70B Model}
\label{sec:70b}

To answer whether the calibration patterns observed in 7B--8B models persist in much larger models, we conducted a layer-wise calibration study on the \textbf{LLaMA-3.1 70B Instruct} model, using the same multilingual MMMLU setup.

\paragraph{Experimental Setup.}
We evaluated the 70B model under identical conditions as the 8B model: 15 languages and 15,000 test examples. To manage computational demands, the model was loaded using 4-bit quantization, which reduces memory usage while preserving performance. All experiments were performed on NVIDIA H100 80GB GPUs. LLaMA-3.1 70B has \textbf{80 transformer layers} (compared to 32 in the smaller models).

\begin{figure}[!ht]
\centering
\includegraphics[width=0.95\linewidth]{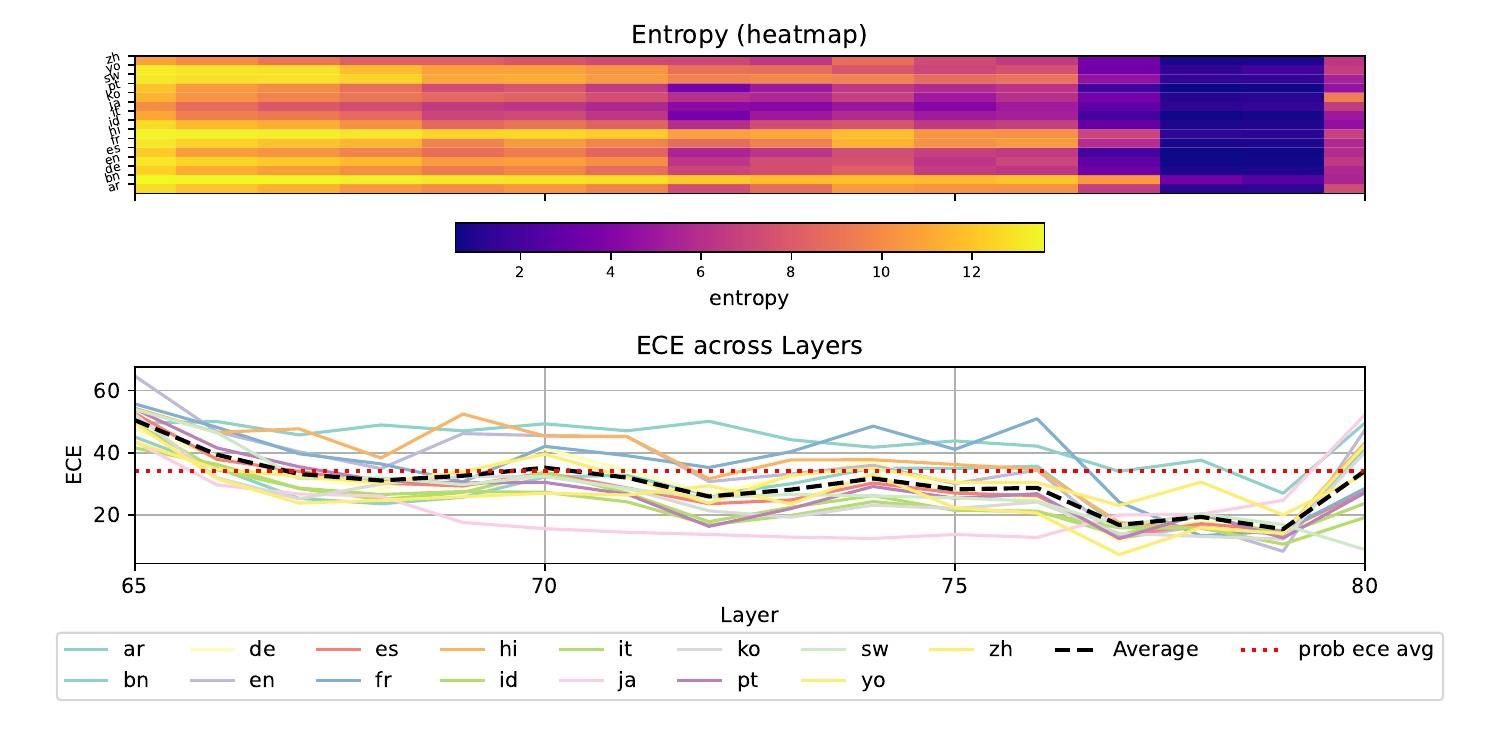}
\caption{Expected Calibration Error (ECE) across the 80 layers of LLaMA-3.1 70B. A clear calibration minimum is visible in the intermediate layers across all language groups.}
\label{fig:70b_ece}
\end{figure}

\paragraph{Layer-wise Calibration in a Large Model.}
The results, shown in Figure~\ref{fig:70b_ece}, demonstrate that the phenomenon observed in smaller models is also present in the 70B model: calibration improves substantially in the late-intermediate layers, well before the final layer. 

\paragraph{Quantifying the Improvement.}

To evaluate whether our calibration methods continue to be effective at larger scales, we also applied \textsc{Best Layer}, \textsc{Good Layer}, and \textsc{LACE} to the 70B model. The results are reported in Table~\ref{tab:70b_calibration}. Following the same format as Table~\ref{tab:llama3_cohere_rescheduled} in the main text, this table shows that the intermediate-layer signal remains strong in the 70B model: using probabilities from the best-performing intermediate layer yields substantially better calibration than relying on the final layer. Moreover, \textsc{LACE} continues to provide the largest overall improvement, achieving the best calibration among all methods. These results confirm that our proposed layer-based calibration strategies, including LACE, scale effectively to much larger models.

\begin{table}[h]
\centering
\footnotesize
\setlength{\tabcolsep}{8pt}
\renewcommand{\arraystretch}{1.14}
\resizebox{\linewidth}{!}{
\begin{tabular}{l
  S[table-format=2.2]
  S[table-format=2.2]
  S[table-format=2.2]}
\toprule
\textbf{Method} & {\textbf{ECE}} & {\textbf{Brier}} & {\textbf{AUROC}} \\
\midrule
\rowcolor{gray!15}
\textsc{Original Probability}$^\dagger$ 
    & 23.51 & 22.10 & 65.19 \\
\phantom{aaa}\footnotesize\textit{(+Temperature Scaling)}
    & 22.56 & 21.97 & 65.19 \\
\phantom{aaa}\footnotesize\textit{(+Isotonic Regression)}
    & 19.01 & 21.34 & 65.07 \\
\midrule
\rowcolor{gray!15}
\textsc{Best Layer}$^\dagger$
    & 16.31 & 16.14 & 84.57 \\
\phantom{aaa}\footnotesize\textit{(+Temperature Scaling)}
    & 13.81 & 14.94 & 84.57 \\
\phantom{aaa}\footnotesize\textit{(+Isotonic Regression)}
    & 7.62 & 13.77 & 84.40 \\
\midrule
\rowcolor{gray!15}
\textsc{Good Layer}$^\dagger$
    & 13.90 & 16.43 & 79.30 \\
\phantom{aaa}\footnotesize\textit{(+Temperature Scaling)}
    & 15.00 & 16.42 & 79.30 \\
\phantom{aaa}\footnotesize\textit{(+Isotonic Regression)}
    & 13.44 & 16.25 & 78.98 \\
\midrule
\rowcolor{gray!15}
\textsc{LACE}$^\ddagger$
    & 6.14 & 15.76 & 79.42 \\
\phantom{aaa}\footnotesize\textit{(+Temperature Scaling)}
    & 4.93 & 15.60 & 79.92 \\
\phantom{aaa}\footnotesize\textit{(+Isotonic Regression)}
    & 2.81 & 15.38 & 78.79 \\
\bottomrule
\end{tabular}
}
\caption{Calibration performance of different layer-based methods on the LLaMA-3.1 70B model using MMMLU. Intermediate-layer methods continue to yield substantial calibration improvements even at large model scale.}
\label{tab:70b_calibration}
\end{table}

\section{Generalisation Beyond MCQA: Full Details}
\label{sec:generative_full}

This appendix supplements the generative-task analysis presented in \S\ref{sec:generative}.

\subsection{Experimental Setup}
\label{sec:generative_setup}

\paragraph{Datasets.} We use two human-translated multilingual generative benchmarks.
\textbf{MKQA}~\citep{longpre-etal-2021-mkqa} is an open-ended multilingual question-answering benchmark. We sample 500 examples per language for seven languages (en, de, es, fr, ru, ja, th), spanning Latin, Cyrillic, Japanese, and Thai scripts.
\textbf{MGSM}~\citep{shi2022language} is a multilingual chain-of-thought math benchmark derived from GSM8K. We use 250 examples per language for eight languages (en, de, es, fr, ru, ja, th, bn).

\paragraph{Correctness measure.}
For MKQA, we follow the standard open-domain QA convention: a generation is scored correct if any gold answer (or alias) is contained in the normalised prediction, or vice versa. Normalisation uses Unicode-aware case folding, punctuation stripping, and whitespace normalisation (consistent with the reference implementation released by the dataset authors).
For MGSM, we extract the final numeric token from the generated chain-of-thought and require an exact match to the gold answer up to a tolerance of $10^{-4}$.

\paragraph{Confidence methods.}
We evaluate four standard generative-confidence elicitation methods, following the taxonomy of \citet{lewislim2025analysing}:
\begin{itemize}[nosep, leftmargin=*]
    \item \textbf{Perplexity}: $\exp\!\left(\tfrac{1}{T}\sum_{t=1}^{T} \log p(y_t|y_{<t},x)\right)$, the geometric mean of next-token probabilities over the generated answer of length $T$.
    \item \textbf{Margin}: the per-token softmax margin between the top-1 and top-2 candidate tokens, averaged across generated tokens. The output lies in $[0,1]$.
    \item \textbf{P(True)}~\citep{kadavath2022language}: we re-prompt the model with ``Question: $\ldots$ Proposed answer: $\ldots$ Is the proposed answer correct?'' and read the softmax probability assigned to the ``True'' token at the next position.
    \item \textbf{Verbalised}~\citep{tian-etal-2023-just}: we append ``How confident are you on a scale of 1--9?'' and decode an expected confidence $\hat c = \mathbb{E}[s | s\in\{1,\dots,9\}]$, mapped to $[0,1]$.
\end{itemize}

\subsection{Per-Language Results}

Tables~\ref{tab:mkqa_per_lang} and \ref{tab:mgsm_per_lang} report per-language accuracy and ECE for each of the four confidence methods. The full numerical layer-wise results, including AUROC and Brier scores at every layer, are released alongside the code (\texttt{layerwise\_mgsm\_llama3.csv}).

\begin{table}[h]
    \centering
    \footnotesize
    \setlength{\tabcolsep}{4pt}
    \resizebox{\columnwidth}{!}{%
    \begin{tabular}{l
        S[table-format=2.1]
        S[table-format=2.1]
        S[table-format=2.1]
        S[table-format=2.1]
        S[table-format=2.1]
    }
        \toprule
        \textbf{Lang.} & {\textbf{Acc.}} & {\textbf{Perp.}} & {\textbf{Margin}} & {\textbf{P(True)}} & {\textbf{Verb.}} \\
        \midrule
        English   & 52.0 & \bfseries 13.5 & \bfseries 10.6 & \bfseries 19.2 & \bfseries 26.9 \\
        \midrule
        German    & 43.0 & 23.4 & 23.7 & 26.3 & 31.2 \\
        Spanish   & 42.8 & 25.2 & 24.1 & 25.7 & 31.0 \\
        French    & 43.2 & 25.1 & 24.7 & 24.2 & 32.3 \\
        Russian   & 26.4 & 39.8 & 41.2 & 37.4 & 49.0 \\
        Japanese  & 18.4 & 45.3 & 47.7 & 42.4 & 55.7 \\
        Thai      & 24.0 & 44.5 & 47.6 & 34.1 & 47.2 \\
        \midrule
        \textit{Avg.\ Non-Eng.} & 33.0 & 33.9 & 34.8 & 31.7 & 41.1 \\
        \bottomrule
    \end{tabular}%
    }
    \caption{Per-language MKQA results (LLaMA3, 500 examples per language). ECE (\%, $\downarrow$) for four confidence methods. Bold marks the column-minimum.}
    \label{tab:mkqa_per_lang}
\end{table}

\begin{table}[h]
    \centering
    \footnotesize
    \setlength{\tabcolsep}{4pt}
    \resizebox{\columnwidth}{!}{%
    \begin{tabular}{l
        S[table-format=2.1]
        S[table-format=2.1]
        S[table-format=2.1]
        S[table-format=2.1]
        S[table-format=2.1]
    }
        \toprule
        \textbf{Lang.} & {\textbf{Acc.}} & {\textbf{Perp.}} & {\textbf{Margin}} & {\textbf{P(True)}} & {\textbf{Verb.}} \\
        \midrule
        English   & 65.6 & \bfseries 18.9 & \bfseries 15.9 & 11.0 & \bfseries 12.8 \\
        \midrule
        German    & 51.6 & 32.5 & 30.3 & 10.9 & 21.7 \\
        Spanish   & 56.4 & 27.8 & 26.0 & 13.0 & 16.3 \\
        French    & 44.8 & 40.1 & 38.7 &  \bfseries 9.5 & 24.4 \\
        Russian   & 47.2 & 37.9 & 36.4 & 13.5 & 19.2 \\
        Japanese  & 44.0 & 40.0 & 37.9 & 17.1 & 25.6 \\
        Thai      & 36.4 & 48.4 & 47.6 & 12.2 & 28.8 \\
        Bengali   & 14.0 & 73.3 & 70.7 & 12.2 & 30.4 \\
        \midrule
        \textit{Avg.\ Non-Eng.} & 42.1 & 42.9 & 41.1 & 12.6 & 23.8 \\
        \bottomrule
    \end{tabular}%
    }
    \caption{Per-language MGSM results (LLaMA3, 250 examples per language). ECE (\%, $\downarrow$) for four confidence methods. Bold marks the column-minimum.}
    \label{tab:mgsm_per_lang}
\end{table}

\subsection{Discussion of the P(True) Outlier on MGSM}

On MGSM, P(True) reports nearly identical ECE for every language ($\sim$10--17\%), in contrast to the wide spread observed for perplexity, margin, and verbalised. This is consistent with the documented brittleness of self-reprompting probes: \citet{kadavath2022language} and \citet{lin-etal-2023-generating} both note that the ``Is the proposed answer correct?'' template is read by the model in the language of the template (here, English), independently of the language of the original question. The resulting probabilities are dominated by the English-language meta-question rather than by the multilingual content of the answer, so the protocol effectively short-circuits the language-conditional confidence signal. We retain the method to faithfully report the benchmark and note that the other three methods, which read confidence directly from the generated tokens, all preserve the multilingual gap.

\subsection{Learned Layer Selector}
\label{sec:selector_details}

The selector's features are the example's per-layer confidences (33 values for LLaMA3); a logistic regression trained on the validation split maps this signature to a choice of layer. Experiments use the MGSM and MKQA subsets of \S\ref{sec:generative} (eight languages, 100 examples per language); the language-agnostic global-layer baseline is tuned on the same validation split in each run.

\section{Robustness of the Multilingual Calibration Gap}
\label{sec:gap_robustness}

The English/Non-English calibration gap reported in \S\ref{sec:benchmarking} is the central empirical claim of the paper. This appendix collects the robustness checks that establish the gap is preserved under statistical resampling of the evaluation instances, different ECE bin counts, and alternative MCQA scoring protocols, and examines one case in which ECE and Brier disagree.

\subsection{Bootstrap Confidence Intervals}
\label{sec:bootstrap_ci}

\noindent\begin{minipage}{\columnwidth}
    \centering
    \footnotesize
    \setlength{\tabcolsep}{3pt}
    \renewcommand{\arraystretch}{1.1}
    \rowcolors{2}{gray!10}{white}
    \resizebox{\columnwidth}{!}{%
    \begin{tabular}{l cccc}
        \toprule
        \multirow{2}{*}{\textbf{Model}}
        & \multicolumn{2}{c}{\textbf{ECE}$\downarrow$}
        & \multicolumn{2}{c}{\textbf{AUROC}$\uparrow$} \\
        \cmidrule(lr){2-3}\cmidrule(lr){4-5}
        & \textbf{EN} & \textbf{Non-EN} & \textbf{EN} & \textbf{Non-EN} \\
        \midrule
        LLaMA3   & [3.93, 8.81]   & [20.72, 24.80] & [77.67, 83.01] & [62.85, 64.71] \\
        Aya      & [17.23, 24.58] & [26.74, 29.06] & [71.57, 77.65] & [67.22, 69.11] \\
        Mistral  & [21.72, 30.28] & [33.12, 35.19] & [70.78, 76.73] & [62.09, 65.07] \\
        Qwen2.5  & [12.71, 19.09] & [21.13, 22.93] & [75.10, 81.36] & [69.17, 71.25] \\
        Phi      & [16.25, 28.71] & [26.97, 29.27] & [66.94, 75.18] & [56.38, 58.46] \\
        DeepSeek & [5.02, 16.16]  & [24.62, 27.91] & [62.79, 69.60] & [55.70, 58.58] \\
        Qwen3    & [8.19, 17.14]  & [13.17, 15.56] & [77.38, 86.99] & [71.91, 74.94] \\
        Olmo3    & [24.07, 34.66] & [26.19, 28.87] & [58.05, 70.96] & [54.47, 58.21] \\
        \bottomrule
    \end{tabular}%
    }
    \captionof{table}{95\% bootstrap confidence intervals for the per-model MMMLU entries of Table~\ref{tab:cross_family}.}
    \label{tab:cross_family_ci}
\end{minipage}

Table~\ref{tab:cross_family_ci} reports 95\% confidence intervals for the per-model entries of Table~\ref{tab:cross_family}, computed with a non-parametric bootstrap: instances are resampled with replacement within each language (2{,}000 resamples; ECE with $K{=}10$ bins), per-language metrics are recomputed, and the non-English value macro-averages the 14 non-English languages of each resample. The English intervals are wider because they rest on a single language split. The English and non-English intervals are disjoint for ECE in five of the eight models (LLaMA3, Aya, Mistral, Qwen2.5, DeepSeek) and for AUROC in every model except Olmo3.

\subsection{ECE Binning Sensitivity}
\label{sec:ECE binning}

In the main paper, following \citet{guo2017calibration}, we report ECE using \(K = 10\) uniformly spaced bins over \([0,1]\), applying the identical binning scheme across all models, layers, languages, and calibration methods.

To evaluate robustness with respect to this hyperparameter, we recomputed ECE for MMMLU predictions for both LLaMA-3 and Aya using \(K \in \{5, 10, 15, 20, 25\}\). Table~\ref{tab:ece_sensitivity} presents the complete results for both LLaMA-3 and Aya.

\begin{table}[h]
\centering
\footnotesize
\setlength{\tabcolsep}{6pt}
\renewcommand{\arraystretch}{1.1}
\resizebox{\columnwidth}{!}{%
\begin{tabular}{c
  S[table-format=2.2] S[table-format=2.2]
  S[table-format=2.2] S[table-format=2.2]}
\toprule
\multirow{2}{*}{\textbf{Bins ($K$)}} &
\multicolumn{2}{c}{\textbf{LLaMA-3 ECE (\%)}} &
\multicolumn{2}{c}{\textbf{Aya ECE (\%)}} \\
\cmidrule(lr){2-3} \cmidrule(lr){4-5}
& {English} & {Non-Eng.} & {English} & {Non-Eng.} \\
\midrule
5  & 3.21 & 17.87 & 19.90 & 25.93 \\
10 & 4.61 & 23.12 & 20.66 & 26.77 \\
15 & 5.04 & 23.27 & 20.34 & 27.45 \\
20 & 5.58 & 22.83 & 21.72 & 28.26 \\
25 & 6.07 & 23.13 & 21.34 & 28.27 \\
\bottomrule
\end{tabular}%
}
\caption{ECE binning sensitivity for LLaMA-3 and Aya on MMMLU. English consistently shows lower calibration error than non-English across all binning configurations.}
\label{tab:ece_sensitivity}
\end{table}

Although absolute values shift somewhat under different bin counts, the \emph{qualitative pattern remains unchanged}: non-English languages remain substantially less calibrated than English for every choice of \(K\), and non-English calibration is highly stable once \(K \ge 10\) (LLaMA-3 within 22.83--23.27\%, Aya within 26.77--28.27\%). Brier score and AUROC (both bin-free) exhibit the same cross-lingual trends, so our conclusions are robust to reasonable variations in the ECE computation scheme.

\subsection{ECE and Brier Disagreement under Capability Collapse}
\label{sec:ece_brier_divergence}

In Table~\ref{tab:llama3_vs_cohere_mmmlu}, LLaMA3's Chinese entry has the worst ECE among all languages (41.94) yet nearly the best Brier score among non-English languages (19.56, second only to Yoruba's 19.43). The cause is that LLaMA3's Chinese accuracy collapses to 23.1\%, below the 25\% random baseline for four-option MCQA, with confidences clustered at a similar level (average 23.2\%; Table~\ref{tab:llama_language_confidence_performance_123}). In this regime the two metrics diverge by construction: Brier rewards uniformly low confidence on mostly-wrong predictions, while ECE captures the bin-level mismatch between confidence and accuracy. The anomaly does not appear in models with functional Chinese performance (Tables~\ref{tab:qwen_mmmlu_multilingual_metrics} and~\ref{tab:deepseekq_mmmlu_multilingual_metrics}). Calibration metrics should therefore not be interpreted for languages where accuracy is at or below chance.

\subsection{Alternative MCQA Scoring Protocols}
\label{sec:protocol_robustness}

The main paper elicits MCQA confidence as the softmax probability assigned to the option-letter token at the choice position (Appendix~\ref{sec:appendix_eva}). This is a standard convention in MCQA evaluation, but not the only one. To verify that the multilingual calibration gap holds under alternative scoring protocols, we evaluate LLaMA3 under three protocols:

\begin{itemize}[nosep, leftmargin=*]
    \item \textbf{P1 -- Letter probability} (paper baseline): softmax over the four option-letter tokens (\texttt{A}, \texttt{B}, \texttt{C}, \texttt{D}) at the choice position; confidence is the max-class probability. This is the protocol used throughout the main paper, reported here at full MMMLU scale (1000 examples per language; numbers reproduced from Table~\ref{tab:llama3_vs_cohere_mmmlu}).
    \item \textbf{P2 -- Option likelihood}: each full option text is scored under the model (sum of token log-probabilities), then softmax is taken across the four options; the model's prediction is $\arg\max$, the confidence is the max softmax value. Evaluated on a balanced 14-language $\times$ 200-example MMMLU subset.
    \item \textbf{P3 -- Text generation}: greedy-decode up to 8 new tokens, regex-extract the predicted letter, and use the softmax probability of that decoded letter at the first generated position as confidence. Evaluated on the same 14-language $\times$ 200-example subset.
\end{itemize}

\begin{table}[h]
    \centering
    \footnotesize
    \setlength{\tabcolsep}{4pt}
    \resizebox{\columnwidth}{!}{%
    \begin{tabular}{l l
        S[table-format=2.2]
        S[table-format=2.2]
        S[table-format=2.2]
        S[table-format=2.2]
    }
        \toprule
        \textbf{Protocol} & \textbf{Group} & {\textbf{Acc.}} & {\textbf{ECE}$\downarrow$} & {\textbf{AUROC}$\uparrow$} & {\textbf{Brier}$\downarrow$} \\
        \midrule
        \multirow{2}{*}{P1: Letter prob.$^\dagger$}
            & EN          & \bfseries 61.20 & \bfseries 4.61  & \bfseries 80.36 & \bfseries 17.63 \\
            & Non-EN avg. & 41.47 & 23.12 & 64.06 & 22.95 \\
        \midrule
        \multirow{2}{*}{P2: Option lik.}
            & EN          & \bfseries 39.50 & \bfseries 20.10 & \bfseries 59.70 & \bfseries 29.10 \\
            & Non-EN avg. & 31.40 & 27.00 & 54.10 & 30.90 \\
        \midrule
        \multirow{2}{*}{P3: Text gen.}
            & EN          & \bfseries 40.50 & \bfseries 6.70  & \bfseries 76.10 & \bfseries 18.30 \\
            & Non-EN avg. & 27.00 & 12.20 & 49.60 & 19.00 \\
        \midrule
        \multicolumn{2}{l}{\textit{Avg. $\Delta$ (EN $-$ Non-EN)}}
            & {+13.78} & {-10.30} & {+16.13} & {-2.61} \\
        \bottomrule
    \end{tabular}%
    }
    \caption{MMMLU calibration under three MCQA scoring protocols on LLaMA3. P1 is reported at full MMMLU scale (1000 examples per language; same numbers as Table~\ref{tab:llama3_vs_cohere_mmmlu}). P2 and P3 are evaluated on a controlled 14-language $\times$ 200-example MMMLU subset. The bottom row averages the EN~$-$~Non-EN gap across protocols. $^\dagger$P1 values come from the main paper's full-scale evaluation.}
    \label{tab:protocols_summary}
\end{table}

\paragraph{Finding.} As Table~\ref{tab:protocols_summary} shows, the EN $-$ Non-EN ordering documented in the main paper is preserved on \emph{every} metric under all three scoring protocols. The averaged gap across protocols is $+13.78$ points of accuracy and $+16.13$ points of AUROC in favour of English, with ECE and Brier both reduced for English. Reading down the table, the P1 row reproduces the main-paper gap (full scale); P2 and P3 verify on a controlled subset that the gap persists even when confidence is derived from full-option likelihoods (P2) or from text-generation outputs (P3) rather than letter-token probabilities. The multilingual calibration gap therefore holds across these three scoring protocols.

\end{document}